\documentclass[11pt]{article}
\usepackage{amsmath}
\usepackage{amssymb}
\usepackage{amsthm}
\usepackage{graphicx}
\usepackage{natbib}
\usepackage[margin=1in]{geometry}
\usepackage[hidelinks]{hyperref}
\bibpunct{(}{)}{;}{a}{,}{,}

\newenvironment{keywords}%
  {\par\small\noindent\textbf{Keywords:\ }}%
  {\par\medskip}

\providecommand{\name}{}\renewcommand{\name}{\normalfont\bfseries}
\providecommand{\addr}{}\renewcommand{\addr}{\normalfont\itshape\small}
\providecommand{\email}{}\renewcommand{\email}{\normalfont\ttfamily\small\quad}
\providecommand{\AND}{}\renewcommand{\AND}{\par\medskip}

\newcommand{\jmlrheading}[7]{}
\newcommand{\ShortHeadings}[2]{}
\newcommand{\firstpageno}[1]{\setcounter{page}{#1}}
\newcommand{\editor}[1]{}

\makeatletter
\renewcommand{\@maketitle}{%
  \newpage\null\vskip 2em%
  \begin{center}%
    {\LARGE\bfseries \@title \par}%
    \vskip 1.5em%
    {\large \@author \par}%
  \end{center}%
  \vskip 1em}
\makeatother
\makeatletter
\def\@clearthm#1{\expandafter\let\csname #1\endcsname\relax
  \expandafter\let\csname end#1\endcsname\relax
  \expandafter\let\csname c@#1\endcsname\relax
  \expandafter\let\csname the#1\endcsname\relax}
\@clearthm{theorem}\@clearthm{lemma}\@clearthm{proposition}\@clearthm{remark}
\@clearthm{corollary}\@clearthm{definition}
\makeatother
\newtheorem{theorem}{Theorem}
\newtheorem{lemma}{Lemma}
\newtheorem{corollary}{Corollary}
\newtheorem{proposition}{Proposition}
\newtheorem{definition}{Definition}
\newtheorem{remark}{Remark}

\usepackage{amsmath}
\usepackage{booktabs}
\usepackage{microtype}
\usepackage[ruled,vlined,linesnumbered]{algorithm2e}
\graphicspath{{./}{../figures/}}
\newcommand{\OPT}{\mathrm{OPT}}
\newcommand{\LP}{\mathrm{LP}}
\newcommand{\Pdim}{\mathrm{Pdim}}
\newcommand{\supp}{\mathrm{supp}}
\newcommand{\G}{\mathcal{G}}
\newcommand{\F}{\mathcal{F}}
\newcommand{\Lc}{\mathcal{L}}
\newcommand{\Dc}{\mathcal{D}}
\providecommand{\R}{}\renewcommand{\R}{\mathbb{R}}
\newcommand{\Cl}{\mathrm{Cert}}
\newcommand{\Sup}{\textnormal{\textsc{Sup}}}

\jmlrheading{}{2026}{}{}{}{}{Haifeng Li and Mo Hai}
\ShortHeadings{CASP: Learning-Augmented Offline Approximation}{Li and Hai}
\firstpageno{1}
\editor{}

\title{CASP: Learning-Augmented Offline Approximation with Verifiable
Certificates and Bounded-Loss PAC Guarantees}

\author{\name Haifeng Li \email mydlhf@cufe.edu.cn\\
      \addr School of Information, Central University of Finance and Economics\\
      Beijing 102206, China
      \AND
      \name Mo Hai (corresponding author) \email haimo@cufe.edu.cn\\
      \addr School of Information, Central University of Finance and Economics\\
      Beijing 102206, China}

\begin{document}
\maketitle

\begin{abstract}
Machine-learned predictions can speed up offline NP-hard optimization, but asking a predictor
what to do amounts to asking it to solve the problem, and committing an unchecked prediction
forfeits every worst-case guarantee. CASP (Certificate-Augmented Solution Pruning) instead asks
which parts of the search space may be ignored, and accepts each answer only after a sound
polynomial-time verifier has checked it, so correctness never depends on prediction quality. We
develop the learning theory of this design. The verifier makes the induced loss class uniformly
bounded, so certificate parameters are learnable from $\tilde O(\varepsilon^{-2}\log K)$ samples
($K$ the maximum instance size),
whereas the unverified commitment class admits no distribution-free rate and, under cost
spread $R$, none below $\Omega(R/\varepsilon^2)$. Filtering noisy
predictions by verifiable confidence dominates the standard min-combiner, with a margin we compute
in closed form, and the prediction stays useful even given the LP, because it breaks ties on
degenerate optimal faces, where every symmetric LP policy, meaning one whose commitments depend
on the instance only through the verifiable confidence values, provably stalls. Experiments on
five problems test the theory's quantitative predictions. With trained predictors, unverified pruning loses up to
$26\%$ of the optimum under distribution shift, while the verified deployment of the same
predictions loses nothing.
\end{abstract}


\section{Introduction}

\subsection{Motivation}
Approximation algorithms face a bottleneck that has resisted progress for over a decade. For a
large class of NP-hard problems the gap between the best inapproximability bound and the best
known ratio has not moved: the $\ln n$ factor for Set Cover is tight under
$\mathrm{P}\neq\mathrm{NP}$ \citep{feige,dinursteurer}, and Vertex Cover cannot be approximated
below $2$ under the Unique Games Conjecture \citep{khotregev}. Learning-augmented algorithms offer
a way around worst-case analysis: consult a machine-learned prediction at run time, beat the
classical bound when the prediction is accurate, and fall back to it when it is not
\citep{lykouris,purohit,mitz,dinitz}. The successes, however, are almost entirely about
\emph{online} problems such as caching and scheduling, where the prediction concerns a future the
algorithm cannot see. Only recently have a few works injected predictions into offline
approximation \citep{antoniadis,cohenaddad,braverman}, and they share one template: predict part of
the optimal solution and prove that the ratio degrades gracefully with prediction error.

We believe progress has been slow because the online template hides a circularity when
transplanted offline. Asking the predictor \emph{what to do} on an NP-hard instance is
self-defeating: a predictor strong enough to output a near-optimal solution has implicitly solved
the problem, while a weaker predictor produces advice that no polynomial procedure can make
trustworthy. Offline approximation is caught between a predictor too strong to exist and one too
weak to use. The ``ratio degrades with error'' template softens the second horn but leaves the
first untouched: the predictor still points at the solution. This paper asks whether predictions
can be used so that the predictor answers a question strictly easier than the original problem,
so that correctness survives arbitrarily wrong predictions, and so that good predictions can make
exact solving polynomial.

\subsection{Our approach}
Our answer inverts the information flow. Instead of asking the predictor \emph{what to do}, a
positive signal, we ask \emph{what may be safely ignored}, a negative signal, and we require the
answer as a polynomial-time-verifiable \emph{certificate}. A certificate asserts that a region of
the search space can be removed; a verifier checks the assertion; only certificates that pass are
used. This dissolves the circularity in three steps. First, deciding that a region need not be
examined is typically far easier than finding the optimum, so the predictor answers a genuinely
simpler question. Second, removing a region may keep some optimum intact, which we call
OPT-preserving, or may inflate the optimum by at most a factor $\rho$, which we call
$\rho$-approximation-safe; this graded notion states exactly when the framework is exact and when
it is approximate. Third, every certificate must pass the verifier, so a wrong prediction is at
worst rejected and never causes an incorrect pruning; whenever
$\rho\,\alpha_{\mathrm{red}}\le\alpha$, in particular under OPT-preserving certificates with an
exact reduced solve, the framework is never worse than its classical fallback
(Theorem~\ref{thm:robust}), in contrast to positive-signal methods whose guarantees lean on the
prediction being good.

\subsection{Contributions}
Beyond formalizing CASP, our contributions concern \emph{learning}, and they share one thesis,
that \emph{checking, not the sign of the signal, is the resource}.
\begin{enumerate}
\item \textbf{The framework.} A certificate system $(\Lc,V,P)$ with two graded safety classes and
five core results: pruning monotonicity, an OPT-preservation identity, composability, a
robustness upper bound, and an exact-solvability condition (Section~\ref{sec:framework}). Table~\ref{tab:verif}
records, for every certificate in the paper, whether the verifier checks the safety level itself
or only the underlying assertion.
\item \textbf{A quantitative theory of confidence filtering.} Filtering noisy predictions by a
verifiable confidence signal dominates the standard min-combiner, and on an explicit Vertex Cover
family we compute the asymptotic margin exactly:
$\min\{\eta[(C-\eta)+\beta(1-\eta)^2],\,\beta\}/(1+2\beta)$, zero at zero noise, increasing at
small noise, and capped at the fallback gap (Theorem~\ref{thm:margin}). The prediction is not
redundant given the LP. On degenerate optimal faces every symmetric LP-commit policy
(Definition~\ref{def:symlp}) provably stalls, and the filter wins by a constant margin at every
noise level because the prediction supplies the tie-break that no function of the LP confidences
can (Theorem~\ref{thm:lponly}). Proposition~\ref{prop:degen} delimits when filtering cannot help, and
a two-threshold filter recovers the high-noise regime at no extra sample cost
(Proposition~\ref{prop:cfplus}). Both quantitative results are equality instances of general
theorems for decomposable families (Theorems~\ref{thm:genmargin} and~\ref{thm:gendeg}), so the
explicit families carry only the arithmetic.
\item \textbf{Bounded loss, and its exact boundary.} The verifier makes the loss class uniformly
bounded, so certificate parameters are learnable with $\tilde O(\varepsilon^{-2}\log K)$ samples,
for single- and multi-parameter classes (Section~\ref{sec:pac}); the structurally identical
bare-commitment class has the same pseudo-dimension yet admits no distribution-free rate
(Theorem~\ref{thm:sepB}, Lemma~\ref{lem:heavytail}), and under bounded cost spread $R$ its sample
complexity is $\Omega(R/\varepsilon^2)$ while CASP's stays $R$-independent
(Theorem~\ref{thm:spreadlb}). Certified optimality behaves the same way.
CASP emits checkable proofs of exactness on a recognizable class (Theorem~\ref{thm:sepA}), the
bare interface emits none, and a fair min-combiner adversary matches both guarantees
(Section~\ref{sec:collapse}). These contrasts are properties of the prediction interface rather
than complexity separations.
\item \textbf{Instantiations.} Five structurally diverse problems, with classical reductions
recast as verifiable certificates: Set Cover, Vertex Cover with triggerable exact solvability on a
named class and a planted distribution, Facility Location with a facility-integral exactness
trigger, 0/1 Knapsack via reduced-cost fixing, and Steiner Tree via classical reduction tests
(Sections~\ref{sec:sc}--\ref{sec:steiner}). None of the individual reductions
is new; the uniform certificate semantics laid over them is.
\item \textbf{Experiments.} Fourteen experiment groups test the theory's quantitative
predictions, from Monte-Carlo
recomputation of the closed forms (E14) to trained predictors under distribution shift, where
unverified pruning loses up to $26\%$ of the optimum and the verified deployment of the same
predictions loses nothing (E13).
\end{enumerate}

\section{Related Work}
\paragraph{Algorithms with predictions.}
The field began with online problems: competitive caching with learned advice \citep{lykouris},
ski rental and scheduling \citep{purohit}, the survey of \citet{mitz}, prediction portfolios
\citep{dinitz}, a primal--dual treatment \citep{bamas}, and tight consistency--robustness trade-offs
\citep{weizhang}. There the prediction concerns an unseen future, so consuming it uncritically is
unavoidable. Offline the situation is different, and the recent offline line is the closest to
this paper. \citet{antoniadis} give prediction-augmented algorithms for
minimum-weight feasible-subset problems in which the predictor supplies solution membership and
the ratio degrades smoothly with prediction error; \citet{cohenaddad} study an
$\varepsilon$-accurate prediction model; \citet{braverman} break approximation
barriers under edge predictions; \citet{permutation} treat permutation problems. All of these commit a (partial) predicted solution and bound the damage as a
function of prediction error. CASP differs on two axes. Correctness and the worst-case ratio are
decoupled from prediction quality entirely, because nothing unverified is ever committed
(Theorems~\ref{thm:robust} and~\ref{thm:noisy}); and the certificate interface supports guarantees
that the commitment interface does not express, although a fair adversary with a fallback can
match them, a boundary we prove rather than blur (Section~\ref{sec:collapse}). The algorithm of \citet{antoniadis} serves as our positive-signal baseline in E10/E10$'$.

\paragraph{Learning for combinatorial solvers.}
Closest in mechanism is machine-learned problem reduction, which trains a model to predict which
variables can be deleted: graph reduction for maximum-weight clique \citep{sunernst}, search-space
classification for subset problems \citep{lauridutta}, and learned duals that warm-start exact
algorithms \citep{dinitzduals}. These are negative signals in our sense, committed without a
verifier, so an erroneous prediction silently destroys optimality; E13 quantifies the difference
by deploying the same trained predictions with and without our verifier under distribution shift.
In continuous optimization, \emph{safe screening rules} \citep{elghaoui,ndiaye} delete variables
of sparse learning programs through duality-based, per-variable, individually verifiable tests,
which is the negative-signal-plus-soundness discipline of CASP with analytic rather than learned
proposals.
Data-driven algorithm configuration \citep{gupta,branch,balcan21,balcan20,cheng} supplies the
pseudo-dimension machinery behind Section~\ref{sec:pac}, but tuned parameters carry no per-instance
guarantee; E5 compares against a configuration-style baseline without a verifier. Learning inside
MILP solvers, to branch, to cut, to dive on predicted partial assignments \citep{nair20}, or to
presolve \citep{l2p,cappart} (see \citealp{bengio21} for a survey), and warm starts with
predictions \citep{sakaue}, accelerate search without shrinking the instance, whereas verified pruning shrinks the instance
itself.
The volume of \citet{beyond} frames the broader beyond-worst-case agenda to which this work
belongs.

\paragraph{Certifying algorithms and classical reductions.}
A certifying algorithm returns, with each output, an easily checked witness of correctness
\citep{certifying}. CASP applies that discipline to predictions rather than to a fixed computation:
an ML output is used only after a sound verifier accepts it, and on the class $\mathcal{K}_{\log}$ the
pipeline additionally emits a checkable proof of exact optimality (Theorem~\ref{thm:sepA}). The
certificates themselves are classical. We reuse greedy and frequency bounds for Set Cover
\citep{chvatal,hochbaum,feige,dinursteurer}, LP rounding and filtering for Facility Location
\citep{jv,lin1992approximations,sta,li} together with the long-documented LP tightness of its benchmarks
\citep{erlenkotter}, Nemhauser--Trotter persistency and the $2k$ kernel for Vertex Cover
\citep{nt,cygan,khotregev}, reduced-cost fixing for Knapsack
\citep{ingargiola,martellototh,pisinger05}, Steiner reduction tests
\citep{duinvolgenant,polzin,scipjack}, and the presolve reductions of modern MILP solvers
\citep{achterberg}. None of these reductions is ours. The contribution is the uniform semantics
laid over them: a verifier, graded safety, and learnable parameters, under which predictions can
accelerate and errors can be rejected.

\section{The CASP Framework}\label{sec:framework}
Consider a minimization problem $\Pi$. An instance $I$ is given by a ground set of building blocks
$\G(I)$, a universe of requirements $U(I)$, a cost $c:\G(I)\to\R_{\ge0}$, and a polynomial-time-decidable feasibility predicate;
a solution is a subset of $\G(I)$, feasible iff it satisfies the predicate. Let $\OPT(I)$ be the minimum feasible cost
($+\infty$ if infeasible). An algorithm is an $\alpha$-approximation ($\alpha\ge1$) if it always
returns a feasible $S$ with $c(S)\le\alpha\cdot\OPT(I)$.

\subsection{Certificate systems and graded safety}
\begin{definition}[Certificate system]\label{def:cert}
A certificate system for $\Pi$ is a triple $\mathcal{C}=(\Lc,V,P)$: an assertion language $\Lc$ of
decidable Boolean predicates on instance structure; a verifier $V(I,\phi,w)\in\{0,1\}$ computable
in polynomial time ($w$ a witness); and a pruning operator $P(I,\phi)$ producing $I_\phi$ with
$\G(I_\phi)\subseteq\G(I)$ and $U(I_\phi)\subseteq U(I)$.
\end{definition}

\begin{definition}[Soundness]\label{def:sound}
$V$ is \emph{sound} for $\Lc$ if $V(I,\phi,w)=1\Rightarrow\phi$ is true on $I$.
\end{definition}
Soundness is the keystone. Any assertion that passes verification is true, so a wrong witness
yields ``not verified'' rather than an incorrect pruning.

\begin{definition}[Two pruning safety classes]\label{def:safe}
Let $\phi$ be true on $I$, $I_\phi=P(I,\phi)$.
(1) $\phi$ is \emph{OPT-preserving} if some optimum $S^\star$ of $I$ survives the pruning: for
block pruning, $S^\star\subseteq\G(I_\phi)$; for a fix--reduce pruning with fixed set $F$
(Definition~\ref{def:legal}), $F\subseteq S^\star$, $S^\star\setminus F\subseteq\G(I_\phi)$, and
$S^\star\setminus F$ is feasible in $I_\phi$.
(2) $\phi$ is \emph{$\rho$-approximation-safe} if $I_\phi$ is feasible and
$\OPT(I_\phi)+c_{\mathrm{fix}}\le\rho\cdot\OPT(I)$, where $c_{\mathrm{fix}}$ is the fixed cost of
the legal pruning (Definition~\ref{def:legal}; $c_{\mathrm{fix}}=0$ for block pruning).
\end{definition}
OPT-preserving implies $1$-approximation-safe. Exact-solving results need OPT-preservation; the
robustness bound tolerates $\rho$-safety. The explicit boundary lets every later result state
when the framework is exact and when it is approximate.

\subsection{Core theorems}
The results below settle, in order, the questions any certificate framework must answer; each is
used later in the paper. \emph{When is pruning lossless?} Lemma~\ref{lem:mono} and
Theorem~\ref{thm:optpres} show that pruning never lowers the optimum and loses nothing when an
optimum survives. \emph{Do
certificates compose?} Theorem~\ref{thm:compose} shows their safety factors multiply, so many small
prunings chain without losing the guarantee. \emph{What if the prediction is wrong?}
Theorem~\ref{thm:robust} bounds the worst case regardless of prediction quality, the property that
decouples correctness from the predictor and, later, makes the loss learnable
(Section~\ref{sec:pac}). \emph{When does pruning change the problem's complexity?}
Theorem~\ref{thm:exact} gives a poly-checkable condition under which an NP-hard instance becomes
polynomially solvable, the basis of the certified-optimality results of Section~\ref{sec:sep}.

\begin{definition}[Legal pruning]\label{def:legal}
$P(I,\phi)$ is \emph{legal} if it is either (a) \emph{block pruning}: only blocks are deleted,
$U(I_\phi)=U(I)$, $\G(I_\phi)\subseteq\G(I)$; or (b) \emph{fix--reduce}: some elements are deleted
while some blocks are \emph{fixed} into the solution, every deleted element is satisfied by a fixed
block, and the fixed cost $c_{\mathrm{fix}}$ is accounted separately so that any feasible $S$ of
$I_\phi$ has $S\cup F$ feasible in $I$ with cost $c(S)+c_{\mathrm{fix}}$.
\end{definition}
All certificates in this paper are legal: Set Cover (SC) and Facility Location (FL) use block
pruning ($U$ unchanged); Vertex Cover (VC) uses fix--reduce (fix $P_1$, delete its covered edges,
$c_{\mathrm{fix}}=c(P_1)$); Steiner Tree uses block pruning on edges; Knapsack exclusion is block
pruning, and inclusion is the maximization mirror of fix--reduce (Remark~\ref{rem:max}).

\begin{lemma}[Pruning monotonicity]\label{lem:mono}
For any true $\phi$ and legal $P$, $\OPT(I_\phi)+c_{\mathrm{fix}}\ge\OPT(I)$; in particular block
pruning ($c_{\mathrm{fix}}=0$) gives $\OPT(I_\phi)\ge\OPT(I)$.
\end{lemma}

\begin{theorem}[OPT-preserving value identity]\label{thm:optpres}
If $V$ is sound and a verified $\phi$ is OPT-preserving, then $\OPT(I_\phi)=\OPT(I)$ for block
pruning; for a fix--reduce $\phi$ the corresponding identity is $\OPT(I_\phi)+c_{\mathrm{fix}}=\OPT(I)$
(equality case of Lemma~\ref{lem:mono}), the form used for Vertex Cover (Theorem~\ref{thm:nt}).
\end{theorem}

\begin{theorem}[Composability]\label{thm:compose}
If $\phi_1$ is $\rho_1$-safe on $I$ and $\phi_2$ is $\rho_2$-safe on $I_{\phi_1}$, then the
composite is $\rho_1\rho_2$-safe. For OPT-preservation, \emph{sequential} composition suffices
under block pruning: if $\phi_1$ is OPT-preserving on $I$ and $\phi_2$ is OPT-preserving on
$I_{\phi_1}$, the composite is OPT-preserving on $I$. If instead both certificates are verified
on $I$ and applied \emph{simultaneously}, OPT-preservation of the composite requires that both
preserve the \emph{same} optimum $S^\star$.
\end{theorem}

\begin{remark}\label{rem:seq}
The distinction matters: two certificates that preserve \emph{different} optima can, applied
simultaneously, delete every optimum (each may prune the other's witness).
Algorithm~\ref{alg:casp} therefore applies certificates sequentially, re-verifying each on the
current reduced instance. Certificates that preserve every optimum, such as Knapsack
exclusion/inclusion (Theorem~\ref{thm:knap}) and Steiner R1, may be batched freely, and the
$\rho_1\rho_2$-safe conclusion of the first clause holds under either discipline.
\end{remark}

\begin{algorithm}[t]
\LinesNumbered
\DontPrintSemicolon
\caption{CASP --- Certificate-Augmented Solution Pruning}\label{alg:casp}
\KwIn{instance $I$; predictor $\mathcal{M}$; certificate system $\mathcal{C}=(\Lc,V,P)$;
      $\alpha$-approximation fallback $\mathcal{A}$; time limit $T$}
\KwOut{a feasible solution $S$ of $I$}
$\Phi \leftarrow \mathcal{M}(I)$\tcp*{predict candidate certificates $\{(\phi_j,w_j)\}$}
$\Phi_{\mathrm{valid}} \leftarrow \{(\phi,w)\in\Phi : V(I,\phi,w)=1\}$\tcp*{\textbf{Layer 1}: verifier keeps only sound certificates}
\If{$\Phi_{\mathrm{valid}} \neq \varnothing$}{
  $I_\Phi \leftarrow P(I,\Phi_{\mathrm{valid}})$\tcp*{prune sequentially, re-verifying on the current instance (Remark~\ref{rem:seq})}
  \If{$I_\Phi$ is feasible}{
    $S \leftarrow \mathrm{Solve}(I_\Phi,\,T)$\;
    \lIf{$S$ is found within $T$}{\Return $S$}
  }
}
\Return $\mathcal{A}(I)$\tcp*{\textbf{Layer 2}: classical fallback}
\end{algorithm}

Algorithm~\ref{alg:casp} implements the inversion with two safety layers. \emph{Layer~1}
(the verifier) discards every certificate whose witness does not check out, so an erroneous
prediction can only be \emph{ignored}, never acted upon; certificates are applied sequentially,
each re-verified on the current reduced instance, unless they preserve every optimum, in which
case batching is safe (Remark~\ref{rem:seq}). \emph{Layer~2} (the feasibility recheck and
the classical fallback $\mathcal{A}$) takes over whenever pruning would leave the instance infeasible
or the reduced solve exceeds the time limit. Hence the output is always feasible and, by Theorem~\ref{thm:robust}, within
$\max(\rho\,\alpha_{\mathrm{red}},\alpha)$ of optimal, independently of prediction quality;
whenever $\rho\,\alpha_{\mathrm{red}}\le\alpha$ it is never worse than $\mathcal{A}$ alone.
Prediction quality affects only how much is pruned, hence the speedup, not correctness.

\begin{theorem}[Robustness upper bound]\label{thm:robust}
If $\mathcal{A}$ is an $\alpha$-approximation, $\mathrm{Solve}$ gives an
$\alpha_{\mathrm{red}}$-approximation ($\alpha_{\mathrm{red}}\ge1$) on the reduced instance, $V$ is
sound, and the adopted pruning, that is the composite of all applied certificates with safety
factors multiplying under composition (Theorem~\ref{thm:compose}), is $\rho$-safe
(Definition~\ref{def:safe}, with $c_{\mathrm{fix}}$ accounted for fix--reduce prunings), then
Algorithm~\ref{alg:casp} returns a
feasible $S$ (including any fixed blocks) with
\begin{equation}\label{eq:robust}
  c(S)\le\max(\rho\,\alpha_{\mathrm{red}},\,\alpha)\cdot\OPT(I).
\end{equation}
In particular if
$\rho\,\alpha_{\mathrm{red}}\le\alpha$ then $c(S)\le\alpha\,\OPT(I)$: CASP is never worse than the
fallback.
\end{theorem}

\begin{remark}[Role of this bound in the rest of the paper]
Bound~\eqref{eq:robust} supports most of what follows. It renders the loss family
\emph{uniformly bounded}, hence PAC-learnable with distribution-free samples (Theorem~\ref{thm:pac},
Section~\ref{sec:pac}); it \emph{decouples correctness from prediction quality}, which we confirm
against adversarial certificate injection in E5; and the confidence-filter of
Theorem~\ref{thm:cfilter} inherits it directly, as its fallback branch is exactly this
$\alpha$-approximation. Most later safety and learnability statements reduce to~\eqref{eq:robust}.
\end{remark}

\begin{theorem}[Emergent polynomial solvability]\label{thm:exact}
If $V$ is sound, if the adopted certificates are \emph{jointly} OPT-preserving, meaning that
some optimum of $I$ survives the composite pruning, as holds under sequential composition of
OPT-preserving certificates (Theorem~\ref{thm:compose}) and, a fortiori, when each certificate
preserves every optimum, and if the reduced instance satisfies
$|\G(I_\Phi)|=O(\log|\G(I)|)$, then CASP returns
an \emph{exact} optimum of $I$ in polynomial time.
\end{theorem}

\begin{remark}
Theorem~\ref{thm:exact} is what distinguishes CASP from solver presolve \citep{achterberg}: when
OPT-preserving certificates compress the effective degrees of freedom to logarithmic size, an
NP-hard instance \emph{emerges} polynomially solvable. Presolve lacks a predictor, lacks
$\rho$-grading, and provides no such hardness-collapse guarantee. Presolve does implement several of the same
underlying reductions, dual fixing for example, so speedup comparisons must control for the
solver's presolve settings; see the protocol note opening Section~\ref{sec:exp}.
\end{remark}

\subsection{Noise and overhead}\label{sec:robust}
Two practical questions complete the framework. First, what happens to correctness when the
predictor is adversarially wrong? Second, when does pruning pay off in wall-clock terms?
Theorem~\ref{thm:noisy} answers the first (nothing: the guarantee is prediction-independent), and
Proposition~\ref{prop:overhead} states the condition for the second.

\begin{theorem}[Noisy-certificate degradation]\label{thm:noisy}
Suppose each predicted certificate is \emph{wrong}, i.e.\ its assertion is false on $I$, with
probability $\eta$. Then for every $\eta\in[0,1]$ CASP stays feasible and
$\mathbb{E}[c(S)]\le\max(\rho\alpha_{\mathrm{red}},\alpha)\OPT(I)$: soundness
(Definition~\ref{def:sound}) keeps $\Phi_{\mathrm{valid}}$ true-only and the feasibility recheck
of Algorithm~\ref{alg:casp} is a second layer, so $\eta$ affects only the number of valid
certificates, hence pruning rate and speedup, not correctness.
\end{theorem}

\begin{proposition}[Net-speedup condition]\label{prop:overhead}
Let $T_{\mathcal{M}}$ be the time of one predictor call, $T_V$ that of one verification,
$k=|\Phi|$ the number of predicted certificates, and let $T_{\mathrm{red}}$ and
$T_{\mathrm{full}}$ be the exact-solve times of the reduced and of the original instance. When a
verified pruning is adopted and the reduced solve completes, CASP runs in
\[
  T=T_{\mathcal{M}}+kT_V+T_{\mathrm{red}}
\]
against the baseline $T_{\mathrm{full}}$, with $T_{\mathcal{M}},T_V$ polynomial; the net speedup
is positive iff $T_{\mathcal{M}}+kT_V<T_{\mathrm{full}}-T_{\mathrm{red}}$, and qualitative when
pruning turns an exponentially-blowing instance into a solvable one. When no certificate
validates, the overhead $T_{\mathcal{M}}+kT_V$ is paid on top of the fallback; if $\mathrm{Solve}$
hits the time limit, the fallback's running time is additionally incurred.
\end{proposition}

\section{Instantiations}\label{sec:inst}
We instantiate CASP on five structurally diverse problems. Set Cover and Vertex Cover are
covering LPs; Facility Location is a two-variable LP with a facility-integral exactness
certificate; 0/1 Knapsack is a single-constraint packing maximization; and the Steiner Tree
certificate has neither a covering nor a packing LP at its core. Together they confirm that the
framework does not live on a single covering-LP skeleton. Each subsection follows the same template: exhibit a
polynomial-time \emph{verifiable} certificate, show it is legal (Definition~\ref{def:legal}), so
that Lemma~\ref{lem:mono} applies, and either OPT-preserving or $\rho$-safe, and then read off the
resulting exactness or approximation guarantee through Theorems~\ref{thm:optpres}
and~\ref{thm:robust}. What differs across problems is only the certificate; the safety and
robustness machinery is shared. Table~\ref{tab:verif} distinguishes, per certificate, what the
verifier actually checks: for most certificates the \emph{safety level itself} is verified, while
for the two complementary-slackness certificates (Set Cover and Facility Location) only the
assertion is verified and OPT-preservation rests on the unverifiable assumption \Sup{}
(Remark~\ref{rem:sup}).

\begin{table}[t]\centering\footnotesize\setlength{\tabcolsep}{4pt}
\caption{What the verifier checks, per certificate. ``Assertion'' is the logical statement $\phi$;
``safety level'' is OPT-preservation or $\rho$-safety. The conditional entries rest on an
assumption (\Sup{}), not a verified property.}\label{tab:verif}
\begin{tabular}{llll}\toprule
Certificate & Assertion verified & Safety level & Safety verified?\\\midrule
SC LP-threshold ($\tau\le1/f$) & yes (LP opt.\ $+$ $\tau$) & $f$-safe & yes\\
SC complementary slackness & yes ($\phi_{CS}$) & OPT-pres.\ under \Sup{} & \textbf{no} (conditional)\\
VC NT persistency & yes (half-integral LP) & OPT-preserving & yes\\
FL LP-threshold & yes & $\Delta$-safe & yes\\
FL complementary slackness & yes ($\phi_{\mathrm{close}},\phi_{\mathrm{cut}}$) & OPT-pres.\ under \Sup{} & \textbf{no} (conditional)\\
FL facility-integral & yes ($y^\star$ integral) & OPT-preserving (exact) & yes\\
Knapsack reduced-bound & yes (bound $+$ incumbent) & OPT-preserving & yes\\
Steiner R1/R2 & yes (degree / Dijkstra) & OPT-preserving & yes\\
\bottomrule\end{tabular}
\end{table}

\subsection{Set Cover}\label{sec:sc}
\emph{The certificate is the LP value itself:} a set the LP barely uses can be dropped, and the
maximum frequency $f$ sets exactly how aggressively. This yields an $f$-safe pruning
(Theorem~\ref{thm:lpsafe}) and, under complementary slackness, an OPT-preserving one
(Theorem~\ref{thm:cscs}).
A Set Cover instance is $(U,\mathcal{S},c)$ with $\mathcal{S}\subseteq2^U$; minimize the cost of a
subfamily covering $U$. Let $f=\max_{e\in U}|\{S\in\mathcal{S}:e\in S\}|$. The LP relaxation
$\min\sum_S c_S x_S$ s.t.\ $\sum_{S\ni e}x_S\ge1\,(\forall e)$, $x\ge0$ has optimum $x^\star$ and
value $\LP(I)\le\OPT(I)$.

\begin{lemma}[Feasibility preservation]\label{lem:scfeas}
If $\tau\le1/f$ and we delete all $S$ with $x^\star_S<\tau$, the result $I_\tau$ stays feasible.
\end{lemma}

\begin{theorem}[LP-threshold certificate is $f$-safe]\label{thm:lpsafe}
For $\tau\le1/f$,
\[
  \OPT(I_\tau)\le f\cdot\OPT(I);
\]
with Theorem~\ref{thm:robust}
($\rho=f,\alpha_{\mathrm{red}}=1$), CASP's ratio is $\le\max(f,\alpha)$.
\end{theorem}

\begin{corollary}[Exact safety boundary]\label{cor:tradeoff}
For $\tau>1/f$, sets of the LP rounding cover $\{S:x^\star_S\ge1/f\}$ may be deleted, the
$f$-bound can fail, and $I_\tau$ may become
infeasible (consider an element covered by $f$ sets, each with $x^\star_S=1/f$; any $\tau>1/f$
deletes them all). Hence $\tau\le1/f$ is the exact $f$-safe boundary; beyond it, any quality gain
is empirical with no guarantee. Accordingly, the deployed verifier checks $\tau\le1/f$ as part of
the certificate (Table~\ref{tab:verif}): for $\tau>1/f$ the certificate is rejected and CASP falls
back, so the threshold family $\F_C$ of Sections~\ref{sec:pac} and~\ref{sec:sep} is uniformly
bounded over \emph{all} $\tau\in[0,1]$, not merely over the safe range.
\end{corollary}

\begin{remark}[A counting floor, vacuous in practice]\label{rem:rate}
Markov counting ($\sum_S x^\star_S\ge\tau\cdot\#\{S:x^\star_S\ge\tau\}$) bounds the survivors by
$\LP(I)/(\tau c_{\min})$ for $c_{\min}=\min_S c_S>0$, hence a pruning-rate floor
$1-\LP(I)/(\tau c_{\min}|\mathcal{S}|)$. We record it for completeness but do not claim it as a
contribution: on all our benchmarks the floor is vacuous (non-trivial on $0/600$ instances, and
still vacuous in a $|\mathcal{S}|=5000$ stress test; E2), so E2 reports pruning rates
\emph{descriptively}, and a non-trivial, instance-adaptive floor is an open problem.
\end{remark}

\begin{theorem}[Complementary slackness, conditionally OPT-preserving]\label{thm:cscs}
Let $(x^\star,y^\star)$ be primal--dual optimal. The assertion $\phi_{CS}(S)$, namely
$\sum_{e\in S}y^\star_e<c_S$, implies $x^\star_S=0$. If some integer optimum $S^\star$ satisfies
$\supp(S^\star)\subseteq\supp(x^\star)$ (condition \Sup{}; e.g.\ a unique integral LP optimum
or an integral LP), then deleting all sets satisfying $\phi_{CS}$ is OPT-preserving.
\end{theorem}

\begin{remark}[\Sup{} is an assumption, not a verified property]\label{rem:sup}
The verifier checks $\phi_{CS}$, and hence $x^\star_S=0$, in polynomial time, but condition
\Sup{} references an integer optimum and is \emph{not} polynomially verifiable in general.
This certificate's OPT-preservation is therefore \emph{conditional}, in contrast to the
unconditionally verifiable safety of the NT, facility-integral, reduced-bound, and Steiner
certificates; Table~\ref{tab:verif} records the distinction for every certificate in the paper.
Experimentally, E6's OPT-preserving/approximation split for Set Cover is determined \emph{post hoc}
against exact optima, and a deployment lacking ground truth can claim only $f$-safety for this
certificate.
\end{remark}

\subsection{Vertex Cover}\label{sec:vc}
\emph{The certificate is Nemhauser--Trotter (NT) persistency, recast as a verifiable reduction:}
it is
OPT-preserving (Theorem~\ref{thm:nt}) and, when the half-integral core is logarithmic, upgrades to
\emph{emergent exact solvability}, solving the NP-hard instance optimally in polynomial time
(Theorem~\ref{thm:Cdet}), the sharpest instance of Theorem~\ref{thm:exact}.
For $G=(W,E,c)$ minimize the weight of a vertex set covering all edges. The LP $\min\sum_v c_v x_v$
s.t.\ $x_u+x_v\ge1\,(\forall uv\in E)$, $x\ge0$ is half-integral: all vertices of the polytope lie
in $\{0,\frac12,1\}$ (Nemhauser--Trotter \citep{nt}).

\begin{theorem}[NT persistency is OPT-preserving]\label{thm:nt}
Given a half-integral optimum $x^\star$ with $P_0=\{x^\star=0\}$, $P_1=\{x^\star=1\}$,
$P_{1/2}=\{x^\star=\frac12\}$, there is a minimum cover $S^\star$ with $P_1\subseteq S^\star$ and
$S^\star\cap P_0=\varnothing$; deleting $P_0$ and fixing $P_1$ leaves exactly the core $G[P_{1/2}]$,
and $\OPT(G)=c(P_1)+\OPT(G[P_{1/2}])$.
\end{theorem}

\subsubsection{Triggerable emergent exact solvability}
We turn ``emergent solvability'' into a provable, triggerable exact algorithm. Call $G$
\emph{$g$-NT-decomposable} if $|P_{1/2}|\le g(|W|)$.

\begin{theorem}[Named-class exactness]\label{thm:Cdet}
If $G$ is $(c_0\log n)$-NT-decomposable ($n=|W|$, logarithms base $2$), minimum weighted vertex
cover is solved exactly in $n^{c_0}\,\mathrm{poly}(n)$ time, and the trigger is one LP solve. Since $|P_{1/2}|\le2\LP(G)$
for unit costs (in general $|P_{1/2}|\le2\LP(G)/c_{\min}$),
every unit-cost graph with LP value $O(\log n)$, and a fortiori with vertex-cover number
$\tau(G)=O(\log n)$, is
solved exactly in polynomial time, recovering the LP $2k$-kernel as a CASP certificate.
\end{theorem}

\subsection{Facility Location}\label{sec:fl}
\emph{Two certificates:} an LP-threshold certificate that is $\Delta$-safe
(Theorem~\ref{thm:flsafe}), and a \emph{facility-integral} certificate; whenever the LP opens
facilities integrally, a polynomially checkable event, nearest assignment is \emph{exactly}
optimal (Theorem~\ref{thm:flint}). The integrality observation itself is folklore, and the frequent
LP-tightness of standard FL benchmarks has been documented since Erlenkotter's dual ascent
\citep{erlenkotter}; the new element is its packaging as a per-instance \emph{verifiable exactness trigger} with an
emitted optimality proof, which upgrades the weakest branch of the framework to certified-exact.
Metric (uncapacitated) Facility Location has a set of facilities (open cost $f_i$ for facility
$i$), a set of clients, and distances $d_{ij}$ satisfying the triangle inequality; open a subset
and assign clients to minimize open plus connection cost. The LP uses $y_i\in[0,1]$ and
$x_{ij}\in[0,1]$:
$\min\sum_i f_iy_i+\sum_{ij}d_{ij}x_{ij}$ s.t.\ $\sum_i x_{ij}\ge1$, $x_{ij}\le y_i$, $x,y\ge0$.
Let $\delta_j=|\{i:x^\star_{ij}>0\}|$ (the LP support of client $j$) and $\Delta=\max_j\delta_j$.

\begin{theorem}[FL LP-threshold certificate is $\Delta$-safe]\label{thm:flsafe}
For $\tau\le1/\Delta$, keeping facilities with $y^\star_i\ge\tau$ (and their edges) keeps $I_\tau$
feasible and $\OPT(I_\tau)\le\Delta\cdot\OPT(I)$; the bound uses no triangle inequality.
\end{theorem}

The dual has $u_j$ for coverage and $w_{ij}$ for $x_{ij}\le y_i$; we avoid the customary
$\alpha_j,\beta_{ij}$, which are reserved for the approximation factor and the triangle-family
parameter. Complementary slackness
yields a two-channel certificate $\phi_{\mathrm{close}}(i):\sum_j w^\star_{ij}<f_i\Rightarrow
y^\star_i=0$ and $\phi_{\mathrm{cut}}(i,j):u^\star_j-w^\star_{ij}<d_{ij}\Rightarrow
x^\star_{ij}=0$, OPT-preserving under condition \Sup{} (as in Theorem~\ref{thm:cscs}). We now
\emph{strengthen} the weakest branch with a checkable condition that yields \emph{exact} solutions.

\begin{theorem}[Facility-integral certificate: FL becomes exact]\label{thm:flint}
Let $(x^\star,y^\star)$ be LP-optimal with $y^\star$ \emph{facility-integral}
($y^\star_i\in\{0,1\}$, a poly-checkable property), $F^\star=\{i:y^\star_i=1\}$. Then assigning each
client to its nearest open facility in $F^\star$ is an \emph{integer optimum}; pruning every closed
facility and its edges is OPT-preserving, and CASP returns an exact optimum in polynomial time.
\end{theorem}

This upgrades FL from ``$\Delta$-safe, OPT-preserving only if unique'' to ``\emph{exact}
whenever the LP is facility-integral,'' triggering the same emergent exactness as NT for VC. In
practice the trigger fires; on real ORLIB-cap data the certificate holds on all $37/37$ instances
with zero optimality mismatch (E1). The high firing rate is expected, since LP tightness on these
benchmarks is well documented \citep{erlenkotter}; the empirical contribution is the per-instance
checkable proof of exactness that accompanies each solve.

\subsection{0/1 Knapsack}\label{sec:knap}
To show the framework is not confined to one covering-LP skeleton we add a problem with the
opposite structure: 0/1 Knapsack is a \emph{packing}, \emph{maximization} problem with a
\emph{single} constraint. \emph{The certificate is classical reduced-cost variable fixing
\citep{ingargiola,martellototh}, the engine of modern core algorithms \citep{pisinger05}, recast
in our language:} a Dantzig bound that forces an item in or out certifies that no optimum flips it,
an inclusion/exclusion OPT-preserving pruning (Theorem~\ref{thm:knap}). Items $i$ have value $v_i$, weight $w_i$; capacity $W$; maximize
$\sum v_i x_i$ s.t.\ $\sum w_i x_i\le W$, $x\in\{0,1\}^n$. Since \citet{antoniadis} also study
Knapsack, the two prediction interfaces can be contrasted directly on the same problem.
Remark~\ref{rem:max} records the maximization mirror of the framework used throughout this
subsection; in particular a certificate is \emph{OPT-preserving} if some optimum survives pruning. The negative signal is an \emph{exclusion-by-reduced-bound}
certificate, the CASP reframing of classical knapsack variable fixing. The predictor supplies
only a strong feasible \emph{incumbent} $z_{\mathrm{low}}$ (value of an exhibited packing, hence
$z_{\mathrm{low}}\le\OPT$).

\begin{remark}[Maximization mirror]\label{rem:max}
Section~\ref{sec:framework} is phrased for minimization. The mirror used here reads: $\OPT$ is
the maximum feasible value; an $\alpha$-approximation returns $c(S)\ge\OPT/\alpha$; legality
replaces cost accounting by value accounting (fixing item $i$ reduces the capacity by $w_i$ and
credits $v_i$, so any feasible $S$ of $I_\phi$ has $S\cup\{i\}$ feasible in $I$ with value
$v(S)+v_i$); monotonicity becomes $\OPT(I_\phi)+v_{\mathrm{fix}}\le\OPT(I)$; OPT-preservation is
unchanged (some optimum survives); and Theorems~\ref{thm:optpres}--\ref{thm:exact} transfer with
the inequalities reversed, by the same proofs.
\end{remark}

\begin{theorem}[Reduced-bound exclusion/inclusion certificate]\label{thm:knap}
Let $U_i$ be the LP (Dantzig) upper bound subject to forcing $x_i=1$, and $U_i^0$ the bound
forcing $x_i=0$, each computable in $O(n)$ by linear-time weighted-median selection. Let $z_{\mathrm{low}}$ be the value of an exhibited
feasible packing. If $\phi_{\mathrm{excl}}(i):U_i<z_{\mathrm{low}}$ holds, no optimum contains $i$,
so excluding $i$ is OPT-preserving; if $\phi_{\mathrm{incl}}(i):U_i^0<z_{\mathrm{low}}$ holds, every
optimum contains $i$. Both are verified in $O(n)$.
\end{theorem}

The surviving items form the knapsack \emph{core} around the break item; a good incumbent shrinks
it, and a verified $O(\log n)$ core brute-forces ($2^{O(\log n)}=\mathrm{poly}$) to an \emph{exact}
optimum with a checkable proof. This is the emergent-exactness mechanism of
Theorem~\ref{thm:exact} (via Remark~\ref{rem:max}), now on a packing problem with a single
constraint and a reduced-cost
certificate, structurally disjoint from SC/VC/FL. Against \citet{antoniadis} (predicting membership, degrading with error), CASP
predicts only a \emph{bound} used to \emph{exclude}; correctness is fully decoupled from prediction
quality (a bad $z_{\mathrm{low}}$ simply fails verification).

\subsection{Steiner Tree}\label{sec:steiner}
\emph{The certificate composes two classical reduction tests, recast as verifiable assertions.}
Steiner Tree asks for a minimum-weight subtree connecting a terminal set (all edge weights
strictly positive); its structure is
graph connectivity rather than a covering or packing LP. The certificate combines two
polynomial-time-verifiable, OPT-preserving reductions from the Steiner preprocessing literature
\citep{duinvolgenant,polzin}, engineered at scale in exact solvers \citep{scipjack}. \textbf{(R1)}
Any non-terminal of degree at most one lies in no optimal Steiner tree, since a leaf of an optimal
tree is a terminal; the test is applied iteratively. \textbf{(R2)} An edge $(u,v)$ of weight $w$
is excludable whenever the shortest $u$--$v$ path in $G\setminus\{(u,v)\}$ costs at most $w$,
because any tree using the edge can be rerouted through the dominating path at no extra cost. The
verifier runs one Dijkstra per tested edge. Both tests are applied \emph{iteratively}, each
re-verified on the current reduced graph, since two individually excludable edges may reroute
through each other and batching R2 deletions is then unsound; under this sequential
discipline the composition is OPT-preserving (Theorem~\ref{thm:compose}, Remark~\ref{rem:seq}).

\section{Learnability of Pruning Parameters}\label{sec:pac}
Pruning parameters, the threshold $\tau$ of Section~\ref{sec:inst} and later the filter
thresholds of Section~\ref{sec:cfilter}, should not be hand-set per instance but \emph{learned}
from a sample, which raises the question of how many samples suffice. The verifier makes this
question well-posed. By capping the loss at a problem constant (Theorem~\ref{thm:robust}) it
renders the loss family uniformly bounded, so standard uniform-convergence tools apply. Let instances follow unknown $\Dc$, policy
$h_\theta:I\mapsto\Phi$, loss $\ell_\theta(I)=c(\mathrm{CASP}_\theta(I))/\OPT(I)\in[1,B]$ with
$B=\max(\rho\alpha_{\mathrm{red}},\alpha)$ from Theorem~\ref{thm:robust}. Throughout the learning
results, $K:=\max_I|\G(I)|$ denotes the maximum instance size.

\begin{proposition}[Single-parameter pseudo-dimension]\label{prop:pdim1}
For the single-threshold class, $\ell_\tau(I)$ is piecewise-constant in $\tau$ with $\le|\G(I)|$
breakpoints, so $\Pdim(\F)\le O(\log K)$.
\end{proposition}

\begin{theorem}[PAC generalization]\label{thm:pac}
If $\Pdim(\F)=d<\infty$ then for
\begin{equation}\label{eq:pacN}
  N\ge\frac{cB^2}{\varepsilon^2}\Big(d\log\tfrac{B}{\varepsilon}+\log\tfrac1\delta\Big),
\end{equation}
with probability $\ge1-\delta$ every $\theta$ satisfies
$\big|\mathbb{E}_\Dc\ell_\theta-\tfrac1N\sum_i\ell_\theta(I_i)\big|\le\varepsilon$; hence ERM's $\hat\theta$
is within $2\varepsilon$ of the in-class optimum.
\end{theorem}

\begin{theorem}[Multi-parameter / combinatorial generalization]\label{thm:multipac}
Let policies be parameterized by $\theta\in\R^p$ such that, for every instance, the pruning
decisions change across at most $\Lambda=\mathrm{poly}(K)$ polynomial boundaries of degree $\le\Delta$
in $\theta$. Then $\ell_\theta(I)$ is piecewise-constant on $\le(8e\Delta\Lambda/p)^p$ cells, and
$\Pdim(\F)=O(p\log(\Delta\Lambda))=O(p\log K)$. With the safety-net bound $\ell_\theta\le B$
preserved for all $\theta$, the sample complexity is
\[
  N=O\!\Big(\varepsilon^{-2}B^2\big(p\log K\log\tfrac1\varepsilon+\log\tfrac1\delta\big)\Big),
\]
polynomial in all parameters.
\end{theorem}

Theorem~\ref{thm:multipac} extends to rich certificate families the boundedness contrast that
Section~\ref{sec:sep} develops: the bare multi-parameter commitment class has the same
$O(p\log K)$ pseudo-dimension yet an unbounded range (Lemma~\ref{lem:heavytail} applies
verbatim), so the range gap of Theorem~\ref{thm:sepB} persists and is closed only by
reinstating a fallback (Theorem~\ref{thm:collapse}). E4 tests both regimes: threshold recovery on the benign
distribution and gap magnitude on a heavy-tailed one (Section~\ref{sec:exp}).

\section{Verifiable Confidence Filtering}\label{sec:cfilter}
How should an algorithm consume a noisy \emph{positive} prediction $\hat S$, a candidate
solution, without inheriting its errors? The standard safe answer commits the entire
prediction and hedges with a classical fallback.

\begin{definition}[Min-combiner positive scheme]\label{def:comb}
Let $\mathrm{fb}$ be a fixed polynomial-time $\alpha$-approximation for $\Pi$. The \emph{combiner}
$A^+_{\mathrm{comb}}$, on instance $I$ and prediction $\hat S$, computes a feasible
$S_{\mathrm{pred}}$ from $\hat S$ (commit-and-complete), computes $S_{\mathrm{fb}}=\mathrm{fb}(I)$,
and outputs whichever is cheaper.
\end{definition}

The min-combiner is safe, since its loss never exceeds $\alpha$ (Theorem~\ref{thm:collapse}),
but coarse; it trusts all of $\hat S$ or none of it. This section shows that a strictly finer
operation, enabled by a \emph{verifiable} per-element confidence signal, provably dominates it,
with a margin we compute in closed form.

Let a predictor output $\hat S\subseteq\G(I)$ together with, for each $i\in\hat S$, a
\emph{confidence} $\sigma_i\in[0,1]$ that is a sound, polynomial-time-verifiable functional of the
instance (in our instantiations $\sigma_i=x^\star_i$, the LP value the verifier already computes).
For $\theta\in[0,1]$ define the \emph{confidence filter} (CF)
\[
  A^{\mathrm{cf}}_\theta(I,\hat S)=\min\!\big(\;\mathrm{cost}[\text{commit }\{i\in\hat S:\sigma_i\ge\theta\}\text{, then complete}]\;,\;c(\mathrm{fb}(I))\big),
\]
i.e.\ commit only the predictions whose verifiable confidence clears $\theta$, greedily complete to
feasibility, and fall back if that is worse; here $\mathrm{fb}$ is the fixed $\alpha$-approximate
fallback of Definition~\ref{def:comb}, whose realized cost satisfies $c(\mathrm{fb}(I))\le\alpha\,\OPT(I)$
and, unlike $\alpha\,\OPT(I)$, is computable. Let $\ell^{\mathrm{cf}}_\theta(I)=A^{\mathrm{cf}}_\theta(I,\hat S)/\OPT(I)$.

\begin{theorem}[Confidence-filter domination]\label{thm:cfilter}
Let $\ell_{\mathrm{mc}}$ be the loss of the min-combiner of Definition~\ref{def:comb} (commit-all,
then min with fallback). Then:
\emph{(i) Containment, consistency, boundedness.} $\ell^{\mathrm{cf}}_0\equiv\ell_{\mathrm{mc}}$; $\ell^{\mathrm{cf}}_\theta\le\alpha$
for every $\theta$; and under a perfect prediction $\ell^{\mathrm{cf}}_\theta(I)=1$ for every
$\theta\le\min_{i\in S^\star}\sigma_i$.
\emph{(ii) Weak domination.} For every distribution $\Dc$,
$\min_\theta\mathbb{E}_\Dc[\ell^{\mathrm{cf}}_\theta]\le\mathbb{E}_\Dc[\ell_{\mathrm{mc}}]$.
\emph{(iii) Strict domination with a computable margin.} Part (ii) is strict on explicit families:
Theorem~\ref{thm:margin} computes the asymptotic margin exactly on a Vertex Cover family, and
Theorem~\ref{thm:lponly} shows the filter also strictly beats every symmetric prediction-free
LP-commit policy (Definition~\ref{def:symlp}); Proposition~\ref{prop:degen} delineates when
strictness is impossible.
\emph{(iv) Learnability.} $\ell^{\mathrm{cf}}_\theta(I)$ is piecewise-constant in $\theta$ with $\le|\hat S|\le K$
breakpoints, so $\Pdim=O(\log K)$ and $\theta^\star$ is PAC-learnable with
$\tilde O(\varepsilon^{-2}\log K)$ samples.
\end{theorem}

\begin{remark}[Relation to the collapse analysis]
Section~\ref{sec:sep} will show that the \emph{sign} of the signal buys nothing once the positive
side may min-combine (Corollary~\ref{cor:noeff}); certified optimality and bounded loss are
supplied by the checking-and-fallback layer, wherever it is installed. Theorem~\ref{thm:cfilter}
is the guarantee that survives that analysis. Within the positive-signal world, a verifiable
confidence signal, the same soundness device CASP is built on, lets a finer policy beat the
min-combiner by checking each prediction's confidence before trusting it, and no fair adversary
removes the margin. The next two subsections make it quantitative, and Section~\ref{sec:cfexp}
measures it on two problems and three noise models. The practical rule is to never commit a noisy
prediction wholesale but to filter it through the verifiable LP value first. On Vertex Cover this
single change brings the cost ratio from the min-combiner's $1.32$ down to $1.03$ at noise level
$0.5$ (Figure~\ref{fig:cf}).
\end{remark}

\subsection{An exact margin}\label{sec:margin}
Theorem~\ref{thm:cfilter}(ii) alone is close to definitional, since the filter family contains
the min-combiner at $\theta=0$; the content lies in the size of the margin and its scaling with
noise. We compute both exactly on an explicit family.

\begin{definition}[Flip noise]\label{def:flip}
Given an instance with a designated optimum $S^\star$, the prediction $\hat S\sim F(\eta)$ contains
each $i\in S^\star$ independently with probability $1-\eta$ and each $i\notin S^\star$ independently
with probability $\eta$.
\end{definition}

\begin{definition}[The family $D_{\beta,C}(n)$]\label{def:Dfam}
For $C>1$ and $\beta>0$, the Vertex Cover instance $D_{\beta,C}(n)$ consists of $n$ disjoint edges
$\{a_i,b_i\}$ with $c(a_i)=1$, $c(b_i)=C$, and $g=\lfloor\beta n\rfloor$ disjoint unit-weight
triangles. $S^\star$ takes every $a_i$ and two designated vertices per triangle;
$\OPT=n+2g$. The VC LP has a unique optimum: $x^\star_{a_i}=1$, $x^\star_{b_i}=0$ on edges and
$x^\star\equiv\frac12$ on triangles, so the verifiable confidence is $\sigma_{a_i}=1$,
$\sigma_{b_i}=0$, $\sigma_v=\frac12$ on triangle vertices. The fallback is LP rounding
($\{v:x^\star_v\ge\frac12\}$, cost $n+3g$); completion is greedy with arbitrary tie-breaking.
\end{definition}

\begin{theorem}[Exact asymptotic margin]\label{thm:margin}
On $D_{\beta,C}(n)$ under flip noise $F(\eta)$, $\eta\in[0,1)$:
\emph{(i)} for every $\theta\in(\frac12,1]$ the confidence filter is \emph{surely} optimal:
$\ell^{\mathrm{cf}}_\theta=1$ deterministically;
\emph{(ii)} the min-combiner satisfies
\[
  \lim_{n\to\infty}\mathbb{E}[\ell_{\mathrm{mc}}]
  \;=\;1+\frac{\min\!\big\{\eta\big[(C-\eta)+\beta(1-\eta)^2\big],\;\beta\big\}}{1+2\beta};
\]
\emph{(iii)} hence the domination margin converges to
\[
  \mathrm{margin}(\eta)\;=\;\frac{\min\!\big\{\eta\big[(C-\eta)+\beta(1-\eta)^2\big],\;\beta\big\}}{1+2\beta},
\]
which is $0$ at $\eta=0$ (consistency), strictly increasing for small $\eta$, and capped at the
fallback gap $\beta/(1+2\beta)$; the cap is attained at some $\eta<1$ iff the unsaturated branch
$\eta[(C-\eta)+\beta(1-\eta)^2]$ reaches $\beta$ on $[0,1)$, for instance whenever
$C>1+\beta$, while for $C\le1+\beta$ the margin may track the unsaturated branch throughout and need not be
monotone in $\eta$.
\end{theorem}

The margin is zero at zero noise, grows with noise, and is capped at the fallback gap. E12
measures this profile (Figure~\ref{fig:cf}), and E14 recomputes the closed form point by
point (Section~\ref{sec:e14}). At $\beta=1$ and $C=2$ the closed form gives $0.2917$ at
$\eta=0.5$, matching the observed $0.29$ up to finite-$n$ and distributional differences of the
benchmark family; the cap $1/3$ is approached only as $\eta\to1$ at these parameters.

\subsection{Prediction is not redundant given the LP}\label{sec:lponly}
The filter's confidence $\sigma$ is computed by the verifier from the LP, which raises a fair
objection. Why not drop the prediction and act on the LP alone? To make $\sigma$ well defined under
degeneracy we canonicalize. The default canonicalization is per-variable:
$\sigma_i:=\max\{x_i:x\ \text{LP-optimal}\}$, computable in polynomial time by one auxiliary LP per
variable; the alternative $\bar\sigma$ reads the analytic center of the optimal face (the limit
point of central-path interior-point methods, exactly computable on the symmetric families below),
and Remark~\ref{rem:canon} shows the conclusions hold under both. We quantify over the following
class.

\begin{definition}[Symmetric LP-commit policies]\label{def:symlp}
Fix a canonicalization $\sigma$ of the LP confidences. A \emph{symmetric (prediction-free)
LP-commit policy} is a deterministic procedure that commits the set $\{i:\sigma_i\in A\}$ for a
fixed $A\subseteq[0,1]$, then greedily completes to feasibility, optionally taking the min with a fallback that is itself
symmetric: every stage of the pipeline, completion tie-breaking and fallback included, must
be invariant under instance automorphisms that preserve $\sigma$. (We use greedy, which has no
ties on the families below; a vertex-returning LP rounding is \emph{not} symmetric, since which
vertex of a degenerate optimal face the solver returns is not a function of $\sigma$.) Variables
with equal confidence are therefore committed or filtered together. The \emph{LP-commit policy}
of the text is $A=\{1\}$; the $\mathrm{LP}\theta$ arm of E12$'$ is $A=[\theta',1]$.
\end{definition}

The next family makes every symmetric
LP-commit policy provably suboptimal while the filter, using the prediction \emph{only to break
the tie on a degenerate optimal face}, wins at every noise level.

\begin{definition}[The twin-gadget family $H_{m,n}(\varepsilon)$]\label{def:Hfam}
A Set Cover instance with (a) $n$ \emph{pairs}: element $u_i$ coverable by $A_i=\{u_i\}$ at cost $1$
or $B_i=\{u_i\}$ at cost $C>1$; and (b) $m$ \emph{gadgets}: elements $\{p_j,q_j,r_j\}$ coverable by
\emph{twin} sets $T_j,T'_j=\{p_j,q_j,r_j\}$, both of cost $1+\varepsilon$, or by singletons
$s^1_j=\{p_j\}$ (cost $1$), $s^2_j=\{q_j\}$ (cost $\frac12$), $s^3_j=\{r_j\}$ (cost $\frac13$), with
$\varepsilon\in(0,\frac38)$. $S^\star$ takes all $A_i$ and one designated twin $T_j$ per gadget;
$\OPT=n+m(1+\varepsilon)$. The optimal LP face on a gadget is the degenerate segment
$\{x_{T_j}+x_{T'_j}=1,\,x_s=0\}$; its analytic center gives $\bar\sigma_{T_j}=\bar\sigma_{T'_j}=\frac12$
and $\bar\sigma_{s^k_j}=0$, while pairs give $\bar\sigma_{A_i}=1$, $\bar\sigma_{B_i}=0$.
\end{definition}

\begin{theorem}[Prediction breaks LP degeneracy]\label{thm:lponly}
On $H_{m,n}(\varepsilon)$ under flip noise $F(\eta)$, for every $\eta\in[0,1)$ and under either
canonicalization: every symmetric LP-commit policy (Definition~\ref{def:symlp}) pays at least
$n+\frac{11}{6}m$ \emph{surely}, since on each gadget it commits both twins or neither, paying
$2(1+\varepsilon)>\frac{11}6$ or falling into the harmonic-singleton greedy trap at
$\frac{11}6$, while the confidence filter at $\theta=\frac12$ pays
$n+m\big[(1+\varepsilon)+\tfrac{11}{6}\eta(1-\eta)\big]$ in expectation on its commit branch
(the min with the realized fallback can only decrease this). Hence
\[
  \mathbb{E}[\mathrm{cost(LP\mbox{-}commit)}]-\mathbb{E}[\mathrm{cost}(A^{\mathrm{cf}}_{1/2})]
  \;\ge\;m\Big[\tfrac{11}{6}\big(1-\eta(1-\eta)\big)-(1+\varepsilon)\Big]
  \;\ge\;m\big(\tfrac38-\varepsilon\big)\;>\;0 ,
\]
with equality in the first step for the branch-only variant against the policy $A=\{1\}$ under the
analytic-center canonicalization,
i.e.\ the filter strictly beats every symmetric LP-commit policy at \emph{every} noise level,
by a per-gadget margin at least $\frac38-\varepsilon$. The prediction's role is exactly to break
the tie on the degenerate optimal LP face, which no symmetric policy can: any function of the
confidences treats $T_j$ and $T'_j$ identically.
\end{theorem}

\begin{remark}[Symmetry is the right boundary]\label{rem:symnec}
Quantifying over symmetric policies is necessary. A policy that breaks ties by
an arbitrary fixed instance order, say committing the lexicographically first twin of each
gadget, attains $\OPT$ on $H_{m,n}(\varepsilon)$ surely, because the twins there are interchangeable; but
it is not a function of the verifiable confidences, and its tie-break is exactly the information
a prediction supplies. Randomization is excluded for the same reason: a policy that flips a
private fair coin per gadget commits a single twin and attains $\OPT$ in expectation, but a coin,
like a fixed order, is a tie-breaking resource that no function of $\sigma$ supplies, and
Definition~\ref{def:symlp} quantifies over deterministic maps. The same holds for a combiner whose fallback rounds a solver-returned
\emph{vertex} of the degenerate face: it too attains $\OPT$ here, and it too is excluded by
Definition~\ref{def:symlp} for the same reason, because the solver's vertex choice is a
tie-break that no function of $\sigma$ can express. Theorem~\ref{thm:lponly} therefore isolates a
\emph{tie-breaking resource}:
no policy expressible through the LP confidences alone can commit a single twin, while a (possibly
learned) prediction can; E13(b) realizes this with a generator tag invisible to the LP. The
same reading applies to hypothesis (B1) of Theorem~\ref{thm:gendeg}.
\end{remark}

\begin{remark}[Robustness to the canonicalization]\label{rem:canon}
Under the max-canonicalization $\sigma_i=\max\{x_i:x\ \text{optimal}\}$, both twins get $\sigma=1$,
so the natural policy $A=\{1\}$ commits \emph{both}, paying $2(1+\varepsilon)$ per gadget, and the
filter's advantage against it becomes
$(1+\varepsilon)-\frac{11}{6}\eta(1-\eta)\ge1+\varepsilon-\frac{11}{24}>0$ for every
$\eta$; committing neither ($1\notin A$) lands in the $\frac{11}6$ trap, as the theorem's case
analysis covers. The separation is therefore not an artifact of how ties are canonicalized; E14
(Section~\ref{sec:e14}) validates both canonicalizations against the closed forms, and
Section~\ref{sec:general} shows the construction is one instance of a general
decomposable-family principle.
\end{remark}

\begin{proposition}[Degeneracy limit of confidence filtering]\label{prop:degen}
If $\sigma_i=s$ is constant across $\hat S$ (an \emph{LP-opaque} instance, e.g.\ any VC instance
whose NT partition is all-$\frac12$), then $\{A^{\mathrm{cf}}_\theta\}_\theta$ contains exactly two
policies, the min-combiner for $\theta\le s$ and prediction-free completion combined with the
fallback for $\theta>s$, so per-element filtering is impossible and strict domination can only come
from choosing between those two. Filtering earns its margin precisely on instances with
\emph{dispersed} confidence values.
\end{proposition}

Proposition~\ref{prop:degen} delimits the guarantee and explains the experimental design of
E12; the margin lives on instances with mixed LP values. It also motivates predictor-supplied (rather than
verifier-computed) confidence, such as calibrated ML confidence checked by the verifier, as the
natural extension when the LP is uninformative.

\paragraph{Repairing the single-threshold interface: the two-threshold filter.}
The single-threshold family has a second, subtler limitation, symmetric to
Proposition~\ref{prop:degen}. It contains no policy that ignores the prediction yet still commits
high-confidence variables; at $\theta$ beyond every $\sigma_i$ it commits nothing and degenerates
to the fallback. When the prediction is noisier than the LP is informative, a
prediction-free LP-commit policy can therefore beat every $A^{\mathrm{cf}}_\theta$ (E12$'$,
Section~\ref{sec:e12p}, measures this crossover). The repair, which we call the two-threshold filter CF$^+$, gives the two
signals separate gates:
\[
  A^{\mathrm{cf+}}_{\theta_1,\theta_2}(I,\hat S)
  =\min\!\big(\mathrm{cost}\big[\text{commit }(\hat S\cap\{\sigma\ge\theta_1\})
  \cup\{i:\sigma_i\ge\theta_2\}\text{, then complete}\big],\,c(\mathrm{fb}(I))\big).
\]

\begin{proposition}[Two-threshold filter]\label{prop:cfplus}
\emph{(i) Containment.} The family $\{A^{\mathrm{cf+}}_{\theta_1,\theta_2}\}$ contains the
min-combiner ($\theta_1{=}0$, $\theta_2$ above every $\sigma_i$), every confidence filter
$A^{\mathrm{cf}}_\theta$ ($\theta_2$ above every $\sigma_i$), every threshold LP-commit policy
$\mathrm{LP}\theta$ ($\theta_1$ above every $\sigma_i$), and the fallback; hence
$\min_{\theta_1,\theta_2}\mathbb{E}_\Dc[\ell^{\mathrm{cf}}_{\theta_1,\theta_2}]$ weakly dominates the best of all
four, on every distribution.
\emph{(ii) Learnability.} $\ell^{\mathrm{cf}}_{\theta_1,\theta_2}(I)$ is piecewise-constant on at most
$(|\hat S|{+}1)(K{+}1)$ rectangular cells, so $\Pdim=O(\log K)$, the $p{=}2$ case of
Theorem~\ref{thm:multipac}, and $(\theta_1^\star,\theta_2^\star)$ is PAC-learnable with
$\tilde O(\varepsilon^{-2}\log K)$ samples. The safety bound $\ell\le\alpha$ of
Theorem~\ref{thm:cfilter}(i) is inherited unchanged.
\end{proposition}

\subsection{Decomposable families}\label{sec:general}
The families $D_{\beta,C}$ and $H_{m,n}$ were designed so that the margins close in closed form,
and one may reasonably ask how much of Theorems~\ref{thm:margin} and~\ref{thm:lponly} is an
artifact of that design. Only the arithmetic is. Both are equality instances of two
general statements about \emph{component-decomposable} families, whose hypotheses are checkable
per component type.

\begin{theorem}[Margin on decomposable families]\label{thm:genmargin}
Let $I_n$ consist of independent copies of finitely many component types $c$ on disjoint ground
sets, $n_c$ copies of type $c$ with $n_c/n\to w_c$, per-component costs bounded by a constant, and
noise independent across elements. Suppose for each type $c$:
\emph{(A1)} there is $\theta^\star$ such that on every noise realization the filter's
commit-and-complete cost on $c$ equals $\OPT_c$;
\emph{(A2)} commit-all-then-complete has expected cost $\OPT_c+\varphi_c(\eta)$ on $c$; and
\emph{(A3)} the fallback costs $\OPT_c+\delta_c$ on $c$, deterministically.
Then $\ell^{\mathrm{cf}}_{\theta^\star}=1$ surely, and
\[
  \lim_{n\to\infty}\ \mathbb{E}[\ell_{\mathrm{mc}}]-\mathbb{E}[\ell^{\mathrm{cf}}_{\theta^\star}]
  \;=\;\frac{\min\{\sum_c w_c\,\varphi_c(\eta),\ \sum_c w_c\,\delta_c\}}{\sum_c w_c\,\OPT_c}.
\]
Theorem~\ref{thm:margin} is the instance with types $\{\mathrm{pair},\mathrm{triangle}\}$,
$\varphi_{\mathrm{pair}}=\eta(C-\eta)$, $\varphi_{\mathrm{tri}}=\eta(1-\eta)^2$,
$\delta_{\mathrm{pair}}=0$, $\delta_{\mathrm{tri}}=1$.
\end{theorem}

\begin{theorem}[Degeneracy advantage on decomposable families]\label{thm:gendeg}
In the same setting add \emph{core} types $g$ and suppose:
\emph{(B0)} on non-core components the filter and every symmetric LP-commit policy
(Definition~\ref{def:symlp}) incur equal cost surely;
\emph{(B1)} every symmetric LP-commit policy pays at least $\OPT_g(1+\delta_g)$ on $g$, with
$\delta_g>0$;
\emph{(B2)} with probability $q_g(\eta)$ the filtered prediction commits a cost-$\OPT_g$ feasible
cover of $g$, and the expected cost of surplus commits on $g$ is at most $\rho_g(\eta)$; and
\emph{(B3)} when no cover is committed the filter's completion on $g$ costs at most
$\OPT_g(1+\delta_g)$.
Then
\[
  \mathbb{E}[\mathrm{cost(LP\mbox{-}commit)}]-\mathbb{E}[\mathrm{cost(filter\ branch)}]
  \;\ge\;\sum_g n_g\big[q_g(\eta)\,\delta_g\,\OPT_g-\rho_g(\eta)\big],
\]
strictly positive whenever $q_g\delta_g\OPT_g>\rho_g$ for some $g$. Theorem~\ref{thm:lponly} is the
instance with a single core type, $q=1-\eta(1-\eta)$, $\delta_g\OPT_g=\tfrac{11}6-(1+\varepsilon)$,
$\rho_g=(1+\varepsilon)\eta(1-\eta)$, and there the bound holds with \emph{equality}.
\end{theorem}

The content of the specific families is therefore only that their components \emph{verifiably
satisfy} (A1)--(A3) and (B0)--(B3) with closed-form $\varphi,\delta,q,\rho$; any other
decomposable family with these properties inherits the same conclusions, and E14 validates the two
canonical instances over the full parameter grid.

\section{Interface Contrasts and Their Collapse}\label{sec:sep}
This section delimits what the certificate interface does and does not buy. Two guarantees
separate CASP from the bare membership-vector commitment interface of Definition~\ref{def:pos},
namely certified optimality (Contrast A) and uniformly bounded loss (Contrast B). Both disappear
against a fair adversary that min-combines with the classical fallback of
Definition~\ref{def:comb} and may run CASP's own pipeline (Section~\ref{sec:collapse}). Neither
is therefore a complexity separation between paradigms. Each isolates what the
checking-and-fallback layer supplies, wherever that layer is installed, and the filtering margin
of Section~\ref{sec:cfilter} is what no fair adversary removes. We spell the contrasts out
nonetheless, both because they calibrate the learning theory through the range factor $B$
of~\eqref{eq:pacN} and because the boundary itself supports the thesis that checking rather than
the sign of the signal is the resource.

\subsection{A common model}
Fix a minimization problem $\Pi$ whose optimality-verification language
$\mathrm{OptVer}=\{(I,k):\OPT(I)\ge k\}$ is coNP-complete (true for Vertex Cover, Set Cover, and
Facility Location; for the \emph{maximization} problem 0/1 Knapsack the corresponding language is
$\{(I,k):\OPT(I)\le k\}$, and all statements below transfer mutatis mutandis). Both paradigms below
receive $I$ plus oracle advice and run in polynomial time.

\begin{definition}[Bare commitment interface / $P$-algorithm]\label{def:pos}
A $P$-algorithm $A^+$ runs in polynomial time; its nontrivial advice is a candidate solution
$\hat S\subseteq\G(I)$ (a ``what to do'' membership vector); it outputs a feasible
$S=A^+(I,\hat S)$ and nothing else; the interface carries no auxiliary proof object. Advice
quality is the Hamming error $h=|\hat S\,\triangle\,S^\star|$ to a nearest optimum. This
membership-type interface covers the models of \citet{antoniadis} and \citet{cohenaddad};
edge-type predictions \citep{braverman} are analogous commitments. The definition restricts the \emph{interface}, not
the computation: a $P$-algorithm may run any polynomial computation, including CASP's own
pipeline, a freedom Section~\ref{sec:collapse} exercises.
\end{definition}

\begin{definition}[Poly-time optimality-proof system]\label{def:proofsys}
A polynomial-time predicate $\Cl(I,S,\pi)$ with $|\pi|=\mathrm{poly}(|I|)$ is an
\emph{optimality-proof system} if (soundness) $\Cl(I,S,\pi)=1$ implies $S$ feasible and
$c(S)=\OPT(I)$. An algorithm \emph{certifies optimality on $I$} if it outputs $(S,\pi)$ with
$\Cl(I,S,\pi)=1$; its \emph{certified-exact rate} on $\Dc$ is
$r(\Dc)=\Pr_{I\sim\Dc}[\text{it certifies optimality on }I]$.
\end{definition}
Certified optimality is the strongest a-posteriori guarantee, since a short and independently
checkable proof of optimality accompanies the solution. It is the quantity on which the two
interfaces diverge.

\begin{lemma}[coNP barrier]\label{lem:barrier}
If $\mathrm{OptVer}$ for $\Pi$ is coNP-hard, no optimality-proof system certifies optimality on
\emph{every} instance unless $\mathrm{coNP}\subseteq\mathrm{NP}$.
\end{lemma}

\subsection{Contrast A: certified optimality}
We instantiate on Vertex Cover; the construction transfers to any problem with a legal
OPT-preserving certificate and a poly-recognizable small-core trigger (Facility Location via
Theorem~\ref{thm:flint}). For a weighted graph $G=(W,E,c)$, let $x^\star\in\{0,\frac12,1\}^W$ be a
Nemhauser--Trotter half-integral LP optimum with partition $P_0,P_{1/2},P_1$ (Section~\ref{sec:vc}).
Let $\mathcal{K}_{\log}=\{G:|P_{1/2}|\le c_0\log|W|\}$ for a fixed constant $c_0$, with $P_{1/2}$ computed
from the LP optimum returned by a fixed polynomial-time solver (persistency holds for every
half-integral optimum, so membership is solver-relative but the certificate is sound for any
choice).

\begin{theorem}[Self-certification and universal barrier]\label{thm:sepA}
\emph{(i) Self-certification.} There is a polynomial-time algorithm, CASP with the NT
certificate followed by core enumeration, that on every $G\in \mathcal{K}_{\log}$ outputs a minimum
vertex cover $S$ \emph{together with} $\pi$ such that $\Cl(G,S,\pi)=1$ for a fixed sound
optimality-proof system $\Cl$; membership $G\in \mathcal{K}_{\log}$ is decided in polynomial time by one LP
solve.
\emph{(ii) Barrier, binding both paradigms.} Unless $\mathrm{coNP}\subseteq\mathrm{NP}$, no
polynomial-time algorithm, CASP included, and no $P$-algorithm of Definition~\ref{def:pos} even
given a perfect prediction $\hat S=S^\star$, outputs, on \emph{every} VC instance, a feasible
cover together with a sound poly-time-checkable optimality proof (Lemma~\ref{lem:barrier}).
Certified optimality is therefore available only on restricted classes; CASP attains it on all of
$\mathcal{K}_{\log}$ via (i), while the bare interface emits no certificate on any instance, by
construction of the interface rather than for complexity reasons, and the same interface equipped
with CASP's checker certifies all of $\mathcal{K}_{\log}$ (Corollary~\ref{cor:noeff}).
\end{theorem}

A concrete planted family realizes the self-certification regime distributionally and generates
the instances of E7.

\begin{definition}[Double-pendant planted model $G(n,k,H)$]
Pick $k$ \emph{centers}, each with two private degree-1 leaves; add an arbitrary graph among the
centers; add an independent set of $n-3k-s$ \emph{outer} vertices joined only to centers; implant a
\emph{hard kernel} $H$ on $s$ fresh vertices, a vertex-disjoint induced subgraph each of whose LP
values is forced to $\frac12$ (e.g.\ disjoint triangles), attached to centers only.
\end{definition}

\begin{proposition}[Planted emergence and interface certification gap]\label{prop:planted}
In $G(n,k,H)$ with $|H|=s$, and assuming the canonical LP optimum below is the one returned (it is
unique): (i) every center is NT-persistent, every leaf and outer vertex lies in
$P_0$, and the NT core is exactly $H$, so $|P_{1/2}|=s$; (ii) for $s=O(\log n)$ CASP solves VC
exactly in polynomial time \emph{and emits a certified-optimal proof}
(Theorem~\ref{thm:sepA}(i)), while the whole-graph LP optimum is fractional
($P_{1/2}=V(H)\neq\varnothing$; for $k=O(s)$ the integrality gap is in addition bounded away from
$1$, the $H$-part contributing $\OPT/\LP=4/3$ on disjoint triangles); (iii) the bare commit-and-complete interface of Definition~\ref{def:pos}
emits no optimality certificate on any of these instances, since its output carries no proof
object, whereas by Corollary~\ref{cor:noeff} the same interface equipped with CASP's checker certifies
all of them (they lie in $\mathcal{K}_{\log}$).
\end{proposition}

Proposition~\ref{prop:planted} is a distributional illustration of the checking layer rather
than a paradigm separation. On inputs whose whole-graph LP optimum is fractional, certified
exactness with probability one becomes available exactly when a checking layer is attached to the
pipeline, whichever paradigm hosts it.

\subsection{Contrast B: bounded loss}
We give a quantitative contrast on the learning axis, in the ERM setting of
Section~\ref{sec:pac}: instances are i.i.d.\ from $\Dc$; a policy maps an instance to a
prediction; we learn its real parameter by ERM under the cost-ratio loss
$\ell(I)=c(\mathrm{ALG}(I))/\OPT(I)\ge1$. The standard uniform-convergence bound
(Theorem~\ref{thm:pac}) for a loss family $\F$ with range $[1,1+B]$ and pseudo-dimension $d$ is
\begin{equation}\label{eq:uc}
N(\varepsilon,\delta)=O\!\Big(\tfrac{B^2}{\varepsilon^2}\big(d\log\tfrac{B}{\varepsilon}+
\log\tfrac1\delta\big)\Big),
\end{equation}
and the dependence on the range $B$ is necessary, because for unbounded-range families no
distribution-free uniform-convergence bound exists. Let $\F_C=\{\ell_\tau:\tau\in[0,1]\}$ be CASP's
single-threshold class (Section~\ref{sec:sc}); let
$\F^+=\{\ell^+_t:t\in\R\}$ be the structurally identical \emph{positive-signal commitment}
class, given by a learned scorer $s:\G(I)\to\R$ and the policy that commits every block with
score at least $t$ and then greedily completes to feasibility. The defining difference is the
absence of a verifier.

\begin{theorem}[Boundedness contrast]\label{thm:sepB}
\emph{(1)} For Set Cover with maximum frequency $f$ and fallback factor $\alpha$,
$\ell_\tau(I)\le B_C:=\max(f,\alpha)$ for all $\tau$ and all $I$, and $\Pdim(\F_C)=O(\log K)$;
hence
\begin{equation}\label{eq:Nc}
  N_C=O\!\Big(\varepsilon^{-2}\max(f,\alpha)^2\big(\log K\log\tfrac1\varepsilon+\log\tfrac1\delta\big)\Big)
  =\tilde O(\varepsilon^{-2}\log K).
\end{equation}
\emph{(2)} $\Pdim(\F^+)=O(\log K)$ as well, yet there is a Set Cover family and a scorer (not
necessarily consistent) on which $\ell^+_t$ is \emph{unbounded}: for every $M$ some instance has
$\sup_t\ell^+_t(I)\ge M$. Thus $\F^+$ has no finite range and, by Lemma~\ref{lem:heavytail} below,
admits \emph{no} distribution-free uniform-convergence guarantee. Restricting to cost spread
$c_{\max}/c_{\min}\le R$ caps the range at $\Theta(R)$, making the uniform-convergence
bound~\eqref{eq:uc} a factor $O(R^2)$ larger than~\eqref{eq:Nc}; Theorem~\ref{thm:spreadlb}
shows an $\Omega(R/\varepsilon^2)$ lower bound, so the spread dependence itself is unavoidable.
\end{theorem}

\begin{lemma}[Unbounded families admit no distribution-free rate]\label{lem:heavytail}
Let $\F$ be a loss family with $\ell_t\ge0$ for which there is a benign instance $I_0$ with
$B_0:=\sup_t\ell_t(I_0)<\infty$ and, for every $M$, an instance $I_M$ with
$\sup_t\ell_t(I_M)\ge M$. Then for every sample size $N$ there is a two-point mixture $\Dc_N$ such
that, with probability at least $0.9$ over $N$ i.i.d.\ samples,
$\sup_t\big|\mathbb{E}_{\Dc_N}\ell_t-\tfrac1N\sum_i\ell_t(I_i)\big|\ge1$. In particular no
distribution-free uniform-convergence bound of the form~\eqref{eq:uc} exists for $\F$.
\end{lemma}

\begin{theorem}[Unavoidable spread dependence]\label{thm:spreadlb}
There are a scorer, an absolute constant $c>0$, and, for every $R\ge3$ and
$\varepsilon\in(0,\frac1{36}]$, two distributions $\Dc^\pm$ over Set Cover instances of cost
spread at most $R$ (and maximum frequency $f=2$) such that any learning rule that, on each of
$\Dc^\pm$, outputs from $N$ i.i.d.\ samples a threshold $\hat t$ with
$\mathbb{E}[\ell^+_{\hat t}]\le\min_t\mathbb{E}[\ell^+_t]+\varepsilon$ with probability at least
$\frac34$ must use $N\ge c\,R/\varepsilon^2$ samples. On the same distributions every CASP
threshold $\tau\in(0,\frac12]$ attains $\ell_\tau\equiv1$ on every realization: the certified
class needs no samples where the commitment class needs $\Omega(R/\varepsilon^2)$, and its rate
$\tilde O(\varepsilon^{-2}\log K)$~\eqref{eq:Nc} is $R$-independent in general. Whether the
$O(R^2/\varepsilon^2)$ upper bound~\eqref{eq:uc} for the commitment class is tight remains open.
\end{theorem}

The two classes have identical combinatorial capacity, $\Pdim=O(\log K)$. The contrast is not
expressiveness but boundedness, which the verifier and fallback supply; the loss range is a
problem constant on one side and an unbounded, input-dependent quantity on the other. It is the
quantitative companion to Theorem~\ref{thm:sepA}. Both statements are, however, statements about
the interface of Definition~\ref{def:pos}, and we now show they do not survive a fair
adversary.

\subsection{A fair adversary collapses both contrasts}\label{sec:collapse}
Theorems~\ref{thm:sepA} and~\ref{thm:sepB} restrict the positive side to consuming a
membership-vector $\hat S$ through a commit-and-complete rule (Definition~\ref{def:pos}). That
restriction accounts for both results. Grant the positive side two standard and entirely legal
powers, the min-combiner of Definition~\ref{def:comb} and the right to run in polynomial time
whatever computation CASP runs, verifier included, and both contrasts vanish. The boundary
itself, rather than a claimed gap, is the result.

\begin{theorem}[Combiner collapse of the boundedness contrast]\label{thm:collapse}
For every instance $I$ and every prediction,
\begin{equation}\label{eq:comb}
  c\big(A^+_{\mathrm{comb}}(I,\hat S)\big)\le\min\big(c(S_{\mathrm{pred}}),\,\alpha\,\OPT(I)\big),
  \qquad\text{hence}\qquad \ell_{\mathrm{comb}}(I)\in[1,\alpha].
\end{equation} Thus the combiner class is uniformly bounded by the problem
constant $\alpha$ (matching CASP's $\max(f,\alpha)$); is \emph{consistent} (a perfect
$\hat S=S^\star$ gives $\ell_{\mathrm{comb}}=1$); and has $\Pdim=O(\log K)$, so by \eqref{eq:uc} its
sample complexity is $\tilde O(\varepsilon^{-2}\log K)$, equal to CASP's. The $O(R^2)$ range gap of
Theorem~\ref{thm:sepB} disappears.
\end{theorem}

\begin{corollary}[No fair capability separation]\label{cor:noeff}
Grant the positive side (a) arbitrary polynomial preprocessing, (b) the min-combiner of
Definition~\ref{def:comb}, and (c) the right to run CASP's own prune--verify--solve pipeline. Then it
matches the negative side on cost, loss range, consistency, sample complexity, \emph{and}
certified-exact rate. In particular a positive algorithm handed a perfect prediction may simply
execute CASP's pipeline and emit the same checkable optimality proof, so even the certified-optimality
advantage of Theorem~\ref{thm:sepA} is an artifact of the commitment interface, not of the sign of the
signal.
\end{corollary}

\begin{remark}[What survives, and why we still build a verifier]
Corollary~\ref{cor:noeff} is not purely negative; it isolates the non-definitional content of
CASP. First, a single OPT-preserving, polynomially verifiable certificate object unifies NT
persistency, complementary slackness, facility-integrality, and reduced-cost fixing as one
construction. Second, self-certification is a real engineering guarantee on
$\mathcal{K}_{\log}$ even though a polynomial positive algorithm could in principle reproduce
it. Third, the verifier makes correctness prediction-independent (Theorems~\ref{thm:robust}
and~\ref{thm:noisy}) without assuming the positive side adopts a combiner. The reframing from
``what to do'' to ``what may be verifiably ignored'' is thus a unification and a design principle
rather than a complexity separation.
\end{remark}

\section{Experiments}\label{sec:exp}
The experiments are confirmatory rather than competitive. Each is tied to one proven statement
and asks whether its prediction holds and whether its bound is tight. We report fourteen experiment
groups (E1--E14, with primed variants E4$'$/E4$''$, E6$'$/E6$''$, E10$'$, and E12$'$) spanning the
five problems, the interface contrasts, the quantitative filtering theory, and the learnability
theory. Table~\ref{tab:map} maps each group to the statement it tests and its headline outcome.
We first verify the framework itself (correctness, quality sources, learnability,
robustness, speedup: E1--E6), then the two heterogeneous problems (E9, E11), then the interface
contrasts and their collapse (E7, E8, E10, E10$'$), and finally the filtering theory and the
learned predictors (E12, E12$'$, E14, then E13). All numbers come from one pipeline---SCIP for exact
and LP solving,\footnote{Audited build: SCIP 10.0 via PySCIPOpt 6.2.1. Solver version and presolve
settings are part of the reproducibility protocol: every speedup comparison is reported against
the solver with its default presolve enabled, since presolve implements some of the same
reductions (Remark after Theorem~\ref{thm:exact}); the artifact release pins the exact solver
build.} time limit
\(3600\)\,s unless stated otherwise, bootstrap \(95\%\) confidence intervals, and an independent reimplementation of every verifier,
the \(\Cl\) checker of Theorem~\ref{thm:sepA}, so that zero-mismatch claims are audited rather
than assumed. Data assets: synthetic Set Cover (\(600\)), Vertex Cover (\(70\)), FL-hard (\(65\)), and
Pisinger Knapsack (\(180\)); \(97\) real DIMACS/SNAP graphs, OR-Library SC (\(40\)),
ORLIB-cap / UFLLib / TSPLIB Facility Location, SteinLib B-class (\(18\)), and \(62\) adversarial
instances.

\begin{table}[t]\centering\small
\caption{The experiment groups, the statement each tests, and its headline outcome (one unified
pipeline). E7--E8 illustrate the \emph{interface contrasts} of Section~\ref{sec:sep}; E12 tests
the capability that survives every fair adversary
(Theorems~\ref{thm:margin},~\ref{thm:lponly}).}\label{tab:map}
\resizebox{\columnwidth}{!}{\begin{tabular}{cllc}\toprule
\# & Experiment / theorem & Headline outcome & Figure/Table\\\midrule
E1 & OPT-preserving exactness (Thm~\ref{thm:nt},\ref{thm:flint}) & 0 mismatch; real FL 37/37 exact & Fig.~\ref{fig:e1nt}\\
E2 & Pruning-rate profile (descriptive; Rem.~\ref{rem:rate}) & rates rise with \(f\); floor vacuous & Fig.~\ref{fig:e2full}\\
E3 & Quality-source ablation (Thm~\ref{thm:lpsafe}) & prune gain \(\to0\) as \(f\) grows & Tab.~\ref{tab:e3}\\
E4 & Sample complexity (Thm~\ref{thm:pac}) & flat landscape; \(0\) excess loss at \(N{=}5\) & Fig.~\ref{fig:learn}\\
E4$'$/E4$''$ & Multi-parameter PAC (Thm~\ref{thm:multipac}) & gap grows with \(p\), vanishes with \(N\) & Fig.~\ref{fig:learn}\\
E5 & Robustness (Thm~\ref{thm:robust},\ref{thm:noisy}) & safety independent of \(\eta\) & Fig.~\ref{fig:e5}\\
E6/E6$'$/E6$''$ & Net speedup (Prop~\ref{prop:overhead}) & \(0\) loss (OPT-pres.); hard-regime net \(13.7\times\) & Fig.~\ref{fig:e6full}, Tab.~\ref{tab:e6}\\
E9 & Heterogeneous Knapsack (Thm~\ref{thm:knap}) & 0 mismatch; core collapse & Fig.~\ref{fig:e9core}\\
E11 & Steiner Tree reductions (\S\ref{sec:steiner}) & 15/18 proven optimal, 0 mismatch & Tab.~\ref{tab:steiner}\\
E7 & Certified-optimality contrast (Thm~\ref{thm:sepA}) & CASP 30/30; bare interface 0/30 & Fig.~\ref{fig:sep}(a)\\
E8 & Boundedness contrast (Thm~\ref{thm:sepB}) & \(1.0\) vs.\ \(172.8\times\) (no fallback) & Fig.~\ref{fig:sep}(b)\\
E10 & Head-to-head vs.\ \citet{antoniadis} & CASP flat; bare commitment degrades & Fig.~\ref{fig:sep}(c)\\
E10$'$ & Fair three-arm rerun (Thm~\ref{thm:collapse}) & combined flat at fallback; bare $+0.65$ & Fig.~\ref{fig:e10fair}\\
E12 & Confidence-filter margin (Thms~\ref{thm:cfilter},\ref{thm:margin}) & margin grows to \(0.29\) & Fig.~\ref{fig:cf}\\
E12$'$ & Pred.-free baselines; CF$^+$ (Thm~\ref{thm:lponly}, Prop~\ref{prop:cfplus}) & adv.\ concentrates on degenerate inst. & Tab.~\ref{tab:e12p}\\
E14 & Exact validation (Thms~\ref{thm:margin},\ref{thm:lponly},\ref{thm:genmargin}) & closed forms recomputed, $O(1/\sqrt n)$ & Figs.~\ref{fig:margin},\ref{fig:lponlyfig}\\
E13 & Learned predictors, verified deployment & unv.\ 26\% loss OOD; verified 0 & Figs.~\ref{fig:verified},\ref{fig:learned}\\
\bottomrule\end{tabular}}
\end{table}

\subsection{E1, E2: correctness invariants}
\begin{figure}[t]\centering
\includegraphics[width=\linewidth]{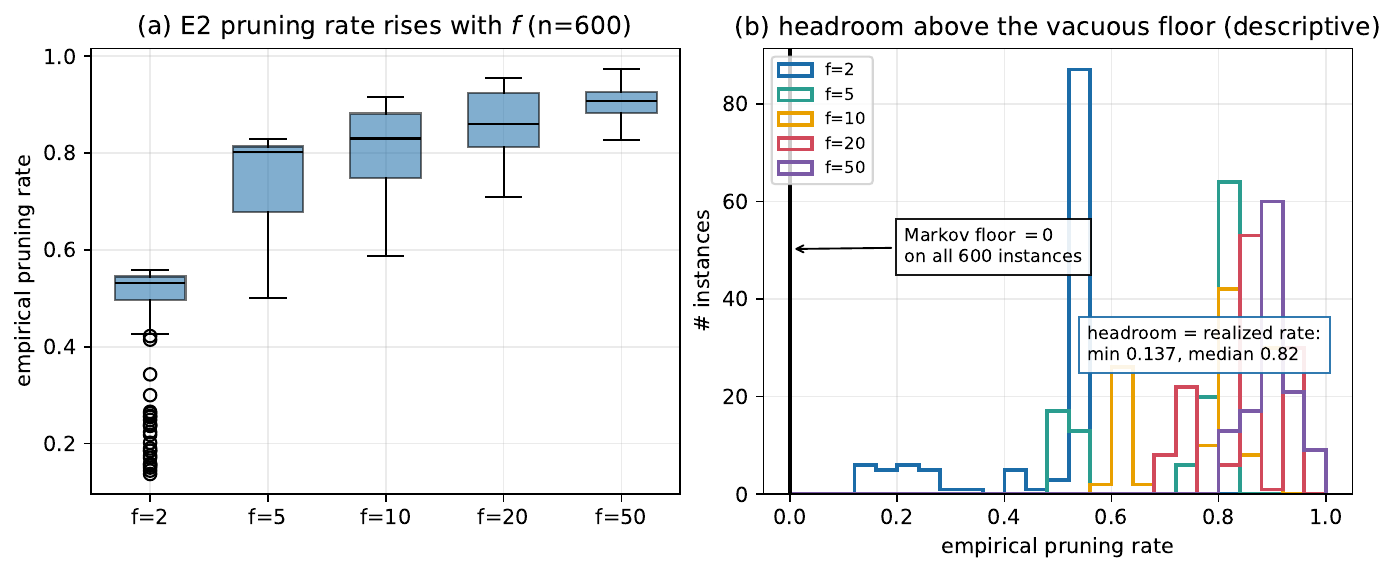}
\caption{\textbf{E2} (all \(600\) Set Cover instances; descriptive). \textbf{(a)} Empirical pruning rate
increases with frequency \(f\). \textbf{(b)} Distribution of realized rates against the Markov
floor of Remark~\ref{rem:rate}, which equals \(0\) on all \(600\) instances; the panel documents
the headroom (min \(0.137\), median \(0.82\)) rather than validating a vacuous
bound.}\label{fig:e2full}
\end{figure}

\begin{figure}[t]\centering
\includegraphics[width=.56\linewidth]{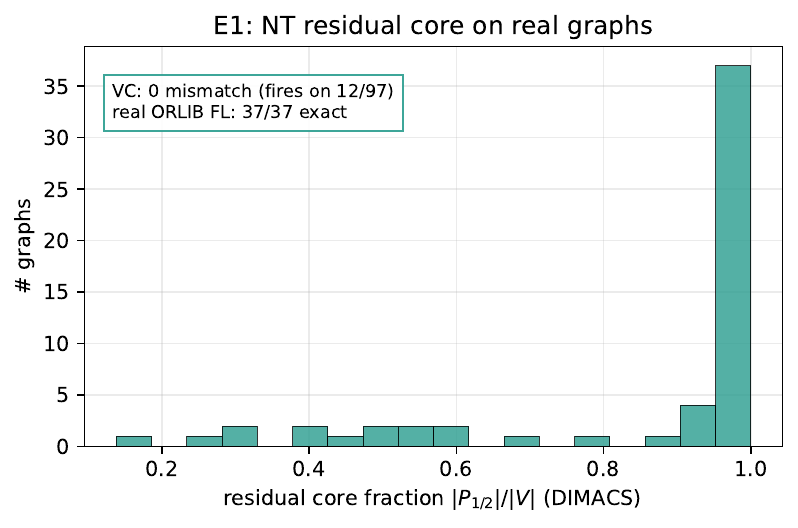}
\caption{\textbf{E1}: distribution of the NT residual-core fraction on the real DIMACS graphs.
The certificate prunes only where structure exists; on real ORLIB-cap Facility Location the
facility-integral trigger fires on all $37$ instances with zero mismatch.}\label{fig:e1nt}
\end{figure}

\textbf{E1 (exactness, Fig.~\ref{fig:e1nt}).} The OPT-preserving certificates incur
\emph{zero} optimality mismatches everywhere they fire. On Vertex Cover, the certificate fires on
\(12/97\) of the real DIMACS/SNAP graphs (the NT core is small enough to prune) and every one
matches the exact optimum; on the remaining real
DIMACS graphs the half-integral core fills the graph (\(|P_{1/2}|\approx|V|\), the histogram in
Fig.~\ref{fig:e1nt}) and the certificate \emph{produces no pruning at all} rather than
fabricating speedup---the behavior a sound certificate must exhibit. On real ORLIB-cap Facility
Location the facility-integral certificate (Theorem~\ref{thm:flint}: the LP sandwich in its proof forces exactness) triggers on \(37/37\)
instances with \(0\) mismatch, turning the historically weakest branch into a real-data
certified-exact result. \textbf{E2 (pruning-rate profile, Fig.~\ref{fig:e2full}).} E2 is descriptive rather than a
bound validation. The Markov floor of Remark~\ref{rem:rate} is vacuous on these benchmarks,
non-trivial on \(0\) of \(600\) instances and still vacuous at \(|\mathcal{S}|{=}5000\), so
plotting violations against it would carry no information. What E2 does establish is the
empirical shape. Across all \(600\) synthetic instances the pruning rate rises monotonically with
the frequency \(f\) (Fig.~\ref{fig:e2full}(a)), consistent with larger \(f\) letting the LP
spread mass over more redundant sets, and Fig.~\ref{fig:e2full}(b) documents how far above the
vacuous floor the realized rates sit. A non-trivial, instance-adaptive rate floor remains open.

\subsection{E3: sources of solution quality}
The four-arm ablation (greedy \(A\), exact \(B\), prune+greedy \(C\), prune+exact \(D\)) isolates
the effect of pruning from the effect of the solver. Table~\ref{tab:e3} stratifies by frequency
\(f\), which resolves a subtlety a blended average would hide. Two invariants hold throughout.
The solver upgrade always helps, \(A{-}B>0\), and \(|D{-}B|=0\) at low \(f\) confirms
OPT-preservation.
The \emph{pruning} contribution to greedy, however, depends sharply on \(f\): at \(f{=}2\),
\(\tau{=}1/f{=}0.5\) prunes so aggressively that the residual is trivialized and greedy already
reaches the optimum (C reaches OPT on \(9/10\), so prune gain \(\approx\) solver gain); as \(f\)
grows the safe threshold $1/f$ of Theorem~\ref{thm:lpsafe} shrinks, the residual stays
genuinely hard, and the pruning gain collapses (ratio
\(0.99\to0.86\to0.45\to0.16\to0.14\)). The claim that pruning's value lies in making exact
solving feasible rather than in improving quality thus holds in the regime with a non-trivial
residual, at moderate and high \(f\); the low-\(f\) windfall is a bonus rather than a
contradiction. The stratification also tells a practitioner when to expect pruning alone to
suffice.

\begin{table}[t]\centering\small
\caption{\textbf{E3}: four-arm ablation stratified by frequency \(f\).
\(A\)=greedy, \(B\)=exact, \(C\)=prune+greedy, \(D\)=prune+exact. Strata sizes are
$10/10/3/1/1$ instances for $f=2/5/10/20/50$; the $f\ge10$ rows are qualitative.}\label{tab:e3}
\begin{tabular}{cccccc}\toprule
\(f\) & prune gain \(A{-}C\) & solver gain \(A{-}B\) & C reaches OPT & \(|D{-}B|\) & prune rate\\\midrule
2 & 526.2 & 528.7 & 9/10 & 0.0 & 0.55\\
5 & 370.4 & 431.7 & 5/10 & 5.6 & 0.79\\
10 & 102.3 & 229.3 & 0/3 & 10.3 & 0.82\\
20 & 10.0 & 64.0 & 0/1 & 1.0 & 0.87\\
50 & 3.0 & 21.0 & 0/1 & 1.0 & 0.92\\
\bottomrule\end{tabular}
\end{table}

\subsection{E4: learning the thresholds}
\begin{figure}[t]\centering
\includegraphics[width=\linewidth]{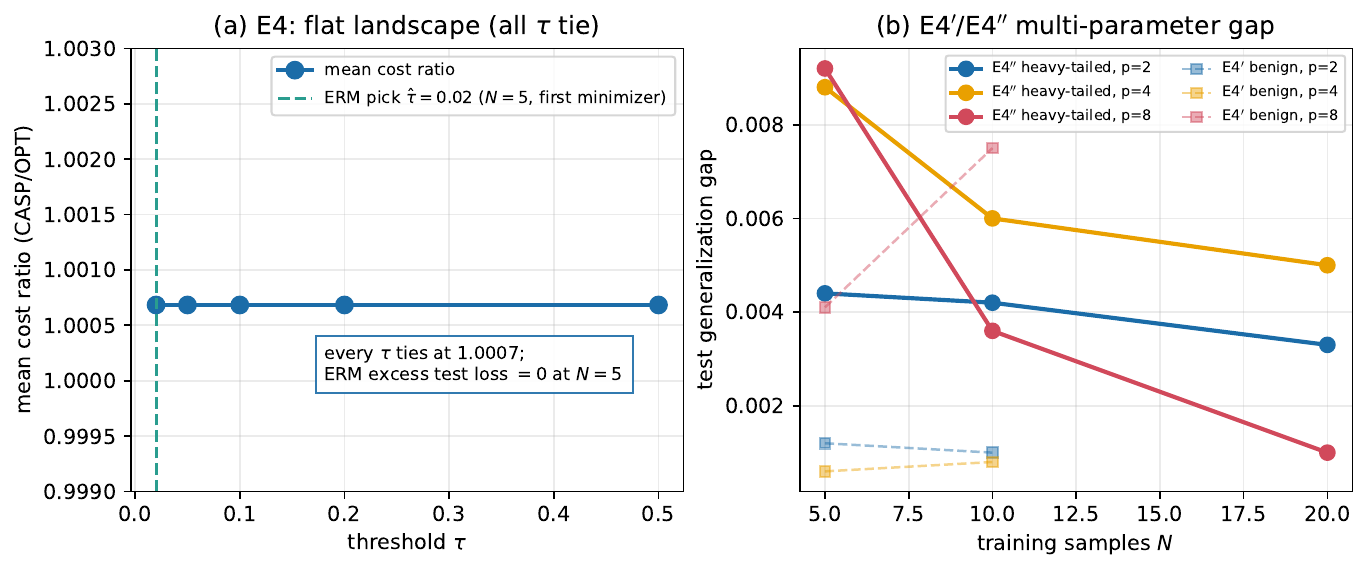}
\caption{\textbf{E4}: learnability. \textbf{(a)} cost-ratio landscape over the single threshold
\(\tau\): on this benign distribution the landscape is \emph{flat} (every \(\tau\) in the safe
grid ties at \(1.0007\)), so ERM attains zero excess test error from \(N{=}5\) trivially; the
discriminating version is E4$''$. \textbf{(b)} multi-parameter generalization gap vs.\ \(N\): on
the heavy-tailed distribution of E4$''$ (solid) the gap is ordered by \(p\) and vanishes as \(N\)
grows, the uniform-convergence shape of Theorem~\ref{thm:multipac}, while the benign-E4$'$
gaps (faint dashed) are an order of magnitude smaller, resolving only the \(p{=}8\) ordering at
\(N{=}5\). Losses stay bounded throughout (\(\le1.14\) benign, \(\le3.02\) heavy-tailed against
the certified ceiling \(f{=}5\)).}\label{fig:learn}
\end{figure}

\textbf{E4 (single threshold, Fig.~\ref{fig:learn}(a)).} On the benign training distribution the
cost-ratio landscape over \(\tau\) is \emph{flat}: every threshold in the safe grid ties at mean
ratio \(1.0007\), so ERM returns the first minimizer (\(\hat\tau{=}0.02\)) and attains zero
excess test error from \(N{=}5\) instances. Zero excess error is the prediction, near-trivial
here, of the logarithmic pseudo-dimension (Proposition~\ref{prop:pdim1}; the
bound~\eqref{eq:pacN} gives the rate~\eqref{eq:Nc}). A landscape with an actual minimum, and the
discriminating learning signal, appear on the heavy-tailed distribution of E4$''$ below. \textbf{E4$'$ (multi-parameter, Fig.~\ref{fig:learn}(b), dashed).} This measures per-bucket threshold vectors \(\theta\in\R^p\). The loss is uniformly bounded
by \(1.14\) for all \(p\in\{2,4,8\}\), the property behind Theorem~\ref{thm:multipac}, and
the generalization gap, while small, grows with \(p\); at \(N{=}5\), \(p{=}8\) exceeds
\(p{=}2,4\), matching \(\Pdim=O(p\log K)\). On this benign distribution (best achievable loss \(\approx1.02\)) the absolute gaps are tiny and
the validating signal is the ordering in \(p\). \textbf{E4$''$ (heavy-tailed costs, Fig.~\ref{fig:learn}(b), solid).}
Re-running the same protocol on \(52\) instances with Pareto(\(1.1\)), size-correlated costs
removes that caveat. The best achievable loss rises to \(1.086\); the observed losses stress
the certified ceiling, with maximum \(3.02\) against the \(f{=}5\) safety factor entering
Theorem~\ref{thm:robust}; and the generalization gap acquires magnitude while keeping both
predicted monotonicities, increasing in \(p\) at fixed \(N\) (\(0.0044/0.0088/0.0092\) at
\(N{=}5\) for \(p{=}2/4/8\)) and vanishing in \(N\) at fixed \(p\) (\(0.0092\to0.0010\)
for \(p{=}8\) as \(N{:}5\to20\)). This is the uniform-convergence shape of
Theorem~\ref{thm:multipac}.

\subsection{E5: robustness}
\begin{figure}[t]\centering
\includegraphics[width=.62\linewidth]{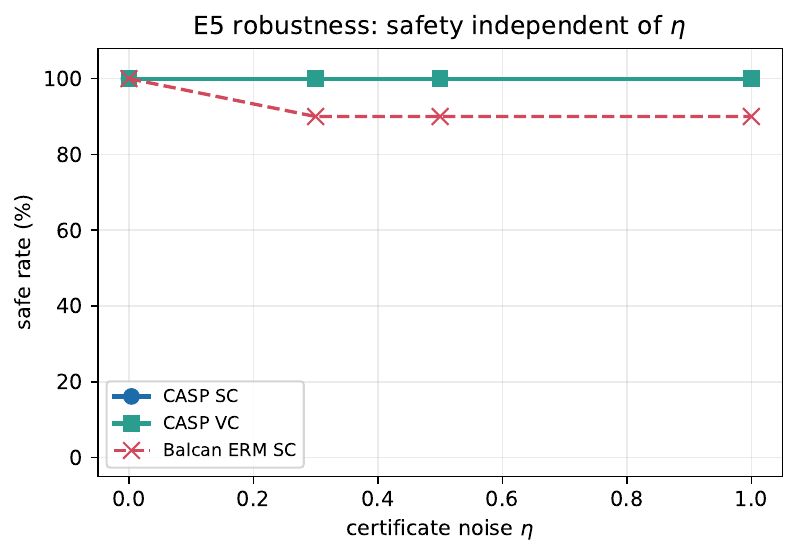}
\caption{\textbf{E5}: safe rate vs.\ certificate noise \(\eta\). CASP (SC and VC) stays at \(100\%\) for all
\(\eta\); safety is independent of the noise level because the verifier and fallback, not the
parameter, provide it (Theorem~\ref{thm:noisy}).}\label{fig:e5}
\end{figure}

Figure~\ref{fig:e5} injects certificate noise at rate \(\eta\) and measures the safe rate (output
feasible and within the worst-case bound). CASP is safe on \(100\%\) of trials at every \(\eta\)
for both Set Cover and Vertex Cover: the two-layer net---verifier rejection of unsound certificates
plus the feasibility recheck with classical fallback---makes the worst-case bound~\eqref{eq:robust}, and hence correctness, \emph{independent of}
\(\eta\) (Theorem~\ref{thm:noisy}). The contrast with a Balcan-style data-driven configuration that applies the learned threshold
without a verifier is instructive. It tracks CASP on clean Set Cover but has no guarantee on
adversarial inputs. The experiment isolates the verifier, not the particular threshold value, as
the source of the robustness.

\subsection{E6: exact-solving speedup}
\begin{figure}[t]\centering
\includegraphics[width=\linewidth]{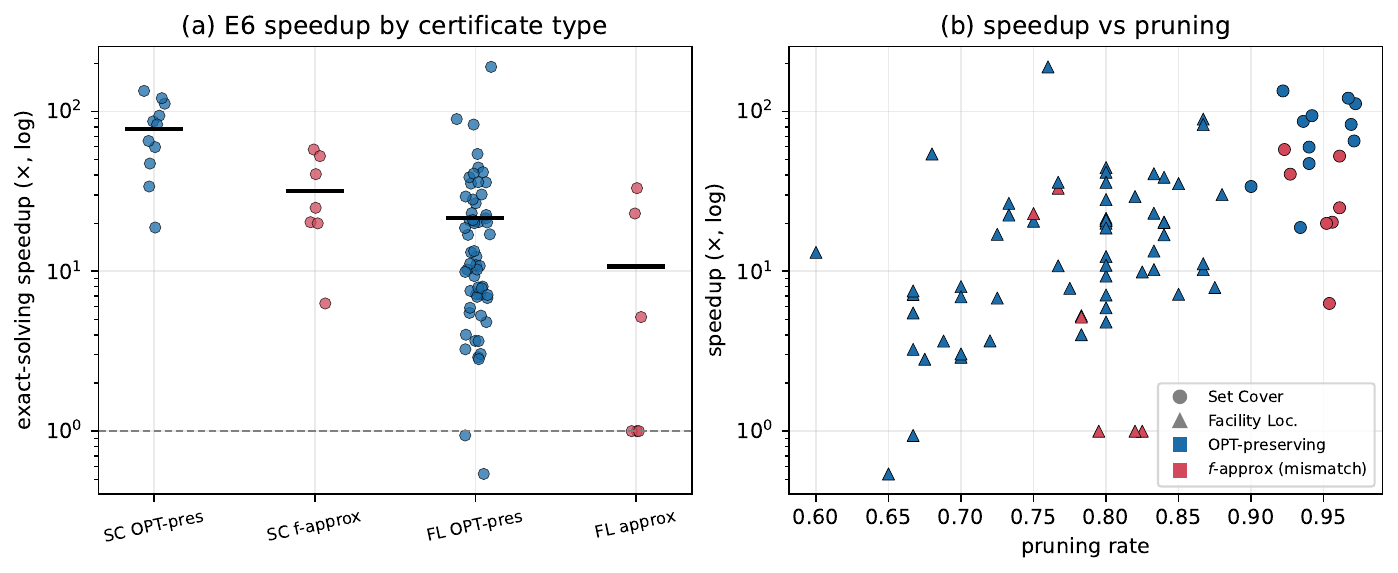}
\caption{\textbf{E6}: net speedup, full data. \textbf{(a)} Distribution by certificate type and problem
(log scale, bars = means); OPT-preserving certificates dominate. \textbf{(b)} Speedup rises with
pruning rate; color = OPT-preserving vs.\ \(f\)-approx, marker = problem.}\label{fig:e6full}
\end{figure}

Table~\ref{tab:e6} and Fig.~\ref{fig:e6full} report speedups split by certificate type. The
split is determined \emph{post hoc} against exact optima for the conditional Set Cover certificate
(Remark~\ref{rem:sup}); a deployment without ground truth can claim the split only for the
certificates whose safety level is itself verifiable (Table~\ref{tab:verif}). E6 covers three
regimes under one protocol (SCIP~10.0, default presolve; see the note opening
Section~\ref{sec:exp}): classical benchmarks (E6), genuinely hard synthetic instances (E6$'$),
and the rail-scale regime (E6$''$).

\textbf{E6 (benchmarks).} The subset of the
$40$ OR-Library instances reported here ($11+7$) is the subset on which the exact solver closes
both the full and the reduced instance within the time limit; the full manifest ships with the
artifact. On real OR-Library Set Cover the OPT-preserving certificates incur \emph{zero}
quality loss, and the \(f\)-approximation-safe cases realize \(0.18\text{--}2.48\%\) cost gaps,
two orders of magnitude below the \(f{=}30\) worst-case bound and direct evidence that the bound
is loose on real instances. Under the artifact's original timing protocol these prunings
measured as mean \(77.8\times\) (max \(134\times\)) and \(31.7\times\) exact-solving speedups,
and synthetic hard Facility Location as \(21.5\times\) (OPT-preserving, \(59\) instances) and
\(10.7\times\) (approx, \(6\)), mean \(20.5\times\), max \(189.9\times\). The speedup
distributions are skewed (medians $82.9\times/24.9\times$ for the two Set Cover arms and
$12.4\times/3.1\times$ for FL; Table~\ref{tab:e6}), and the six approximate FL cases have
realized gaps of $0.06$--$49\%$, three essentially free and three paying the coarse
$\Delta$-safe budget, so the split rather than the blend is the informative summary.
Figure~\ref{fig:e6full}(b) shows speedup rising with the pruning rate across both problems, as
Proposition~\ref{prop:overhead} predicts; the speedup is largest when the solver bottleneck is
the number of variables (FL with many facilities, SC with many sets). A protocol audit, however,
\emph{supersedes the OR-Library wall-clock figures}. A $2\times2$ audit, solver presolve on or
off crossed with CASP pruning on or off, run on the scp4x--7x families, shows every such
instance solving within seconds under SCIP~10.0 with default presolve, so no wall-clock speedup
is measurable there under the audited protocol; the quality figures cross-validate exactly, with
mean $f$-safe gap $0.35\%$, maximum $2.48\%$, and post-hoc OPT-preservation rate $0.60$,
matching the $11/18$ split. Table~\ref{tab:e6} therefore documents the audited \emph{quality}
split, its Set Cover timing columns predate the audit and are retained for provenance, and the
timing claims of this paper rest on the regimes where exact solving is genuinely hard under the
audited protocol, measured next.

\textbf{E6$'$ (hard synthetic).} On $40$ large synthetic Set Cover
instances ($f{=}10$, $m{=}2000$; $300$\,s limit per solve) the full instance times out on
$31/40$. On the $9$
instances both arms solve, the \emph{net} speedup (LP and pruning time charged to CASP) is mean
$13.7\times$, median $12.9\times$, max $21.3\times$, with $f$-safe gaps $\le0.35\%$ (OPT-preserved
post hoc on $1/9$). Two further instances escape the time limit only after pruning, solving in $18$--$148$\,s where
the full instance exceeds $300$\,s, the qualitative regime of Proposition~\ref{prop:overhead},
and on $23$ of the $29$ double-timeouts the pruned run holds the better incumbent at the limit.
Pruning therefore buys nothing where modern presolve already solves in seconds and a consistent
order of magnitude where exact solving is genuinely hard.

\textbf{E6$''$ (rail scale).} The largest OR-Library instances, the rail family parsed with a
dedicated loader for its transposed format, probe the large-$f$ extreme. On rail507, with $507$
rows, $63{,}009$ columns, and $f{=}7753$, the LP solves in $11$\,s; the $\tau{=}1/f$ threshold,
despite its nominally useless $f$-safe factor, prunes $\mathbf{99.5\%}$ of the columns because
the LP support is tiny; and the pruned instance solves to its optimum, $176$, in $7.7$\,s versus
$530$\,s for the full instance, which SCIP~10.0 proves at the known best $174$: a realized gap
of $1.15\%$ and, with the $11$\,s LP charged to CASP, a net $28\times$ end-to-end gain. rail516
behaves the same way ($19.1\,\mathrm{s}\to0.6$\,s, $184$ vs $182$, $1.10\%$). Rail is an extreme case of the loose-bound phenomenon; the certified
$f$-safe factor of roughly $7800$ is vacuous while the realized loss is one percent. Variable pruning works even here; what is missing is a certificate
class whose \emph{guarantee}, not merely its behavior, survives large $f$.

\begin{table}[t]\centering\small
\caption{\textbf{E6}: exact-solving speedup under the artifact's original timing protocol, split
by certificate type (split determined post hoc for the conditional SC certificate;
Remark~\ref{rem:sup}). Medians and bootstrap $95\%$ CIs are recomputed from the released
per-instance artifact ($2\times10^4$ resamples, seed $42$): the four means $77.8/31.7/21.5/10.7$
carry CIs $[57,98]/[20,45]/[15,30]/[2,21]$. The Set Cover timing columns predate the presolve
audit and are retained for provenance (see E6 text); the quality columns are audited.}
\label{tab:e6}
\begin{tabular}{llccccc}\toprule
Problem & certificate & \(n\) & mean & median & max & quality gap\\\midrule
Set Cover & OPT-preserving & 11 & \(\mathbf{77.8\times}\) & \(82.9\times\) & \(134.3\times\) & \(0\%\)\\
Set Cover & \(f\)-approx-safe & 7 & \(31.7\times\) & \(24.9\times\) & \(57.8\times\) & \(0.18\text{--}2.48\%\)\\
Facility Loc. & OPT-preserving & 59 & \(21.5\times\) & \(12.4\times\) & \(189.9\times\) & \(0\%\)\\
Facility Loc. & approx & 6 & \(10.7\times\) & \(3.1\times\) & \(33.1\times\) & \(0.06\text{--}49\%\)\\
\bottomrule\end{tabular}
\end{table}

\subsection{E9: Knapsack}
\begin{figure}[t]\centering
\includegraphics[width=.62\linewidth]{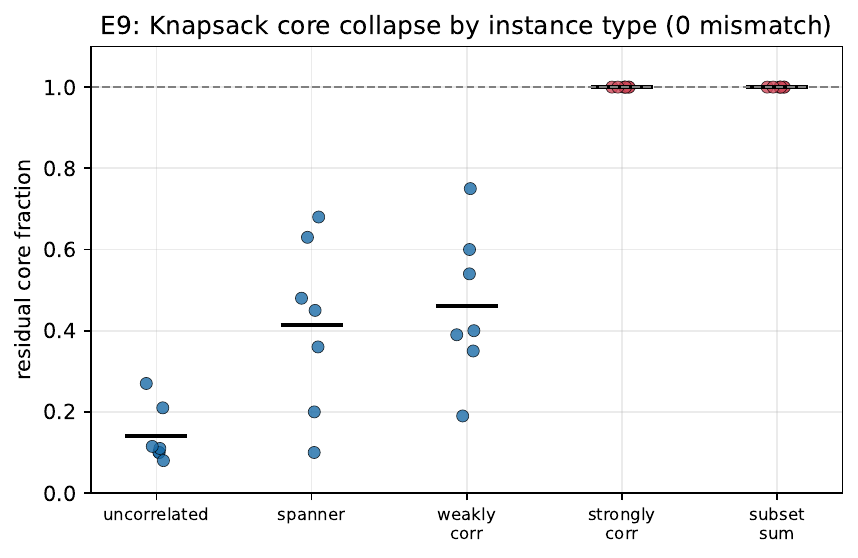}
\caption{\textbf{E9} (Theorem~\ref{thm:knap}): residual core fraction per Knapsack instance type;
points are instances, bars are means. The core collapses on structured types and stays full on the
hard correlated types, with zero mismatch throughout.}\label{fig:e9core}
\end{figure}

Figure~\ref{fig:e9core} demonstrates the framework on 0/1 Knapsack, a packing, maximization,
single-constraint problem disjoint from the covering LPs of SC/VC/FL. The reduced-bound certificate
(Theorem~\ref{thm:knap}) is correct on all \(35\) instances (seven per Pisinger structural type;
the E9/E13 subset manifests ship with the artifact; \(0\) mismatch, audited) and its
behavior is structure-dependent in the way the theory predicts. On uncorrelated and spanner
instances it collapses the residual core to \(0.14\) and \(0.41\), the emergent-exactness
regime; on strongly correlated and subset-sum instances, where the LP gap is tiny and every item
sits in the core, it fixes nothing and the core fraction stays at \(1.0\). This mirrors NT on
hard graphs and answers the objection that the framework is confined to one LP skeleton.

\subsection{E11: Steiner Tree}\label{sec:e11}
E11 evaluates the Steiner certificates of Section~\ref{sec:steiner} (R1/R2 applied iteratively,
re-verified after each deletion) on the SteinLib benchmark.
We solve both the original and the reduced instance with an exact multi-commodity-flow ILP (a
\(3\)-hour budget). Of the \(18\) B-class instances, \textbf{\(15\) are solved to proven
optimality}; on \emph{every} one of them the certificate is OPT-preserving with \textbf{\(0\)
mismatch} (\(\mathrm{opt}(G)=\mathrm{opt}(G_{\text{pruned}})\)), and the value matches the
\emph{published} SteinLib optimum exactly (Table~\ref{tab:steiner}). The certificate prunes a mean
of \(15.3\%\) of edges (max \(28.6\%\)). The remaining three instances reflect the limits of the generic ILP rather than of the
certificate. On B15 the exact solver finds the optimum \(318\) on both graphs yet cannot close
the duality gap, and the other incumbents correctly lie above the true optimum. A specialized
Steiner solver closes the B-class in milliseconds; the point here is only that the certificate is
sound and OPT-preserving wherever exactness can be checked. A second, independent exact solver based on integer row generation
closes \(12\) of the instances within budget. It agrees with the multi-commodity-flow solver and
with the published SteinLib value on every one of them, which independently confirms
OPT-preservation. Steiner Tree thus shows that the framework extends beyond linear-programming
structure to a genuinely different combinatorial problem.

\begin{table}[t]\centering\small
\caption{\textbf{E11}: Steiner Tree on SteinLib B-class (representative + aggregate). On every instance solved
to proven optimality, CASP's optimum on the pruned graph equals the optimum on the full graph and
the published SteinLib value (\(0\) mismatch). On unproven rows (---) the OPT columns show the
solver's incumbents, which correctly lie above the published optimum.}\label{tab:steiner}
\resizebox{\columnwidth}{!}{\begin{tabular}{lcccccccc}\toprule
inst & \(|V|\) & \(|E|\) & \(|T|\) & prune \% & full OPT & pruned OPT & SteinLib & proven\\\midrule
B01 & 50 & 63 & 9 & 28.6 & 82 & 82 & 82 & \checkmark\\
B07 & 75 & 94 & 13 & 24.5 & 111 & 111 & 111 & \checkmark\\
B09 & 75 & 94 & 38 & 16.0 & 220 & 220 & 220 & \checkmark\\
B13 & 100 & 125 & 17 & 27.2 & 165 & 165 & 165 & \checkmark\\
B14 & 100 & 125 & 25 & 21.6 & 235 & 235 & 235 & \checkmark\\
B16 & 100 & 200 & 17 & 5.5 & 127 & 127 & 127 & \checkmark\\
B17 & 100 & 200 & 25 & 10.5 & 131 & 131 & 131 & \checkmark\\
B18 & 100 & 200 & 50 & 5.5 & 219 & 221 & 218 & --- (ILP gap)\\\midrule
\multicolumn{9}{l}{\emph{Aggregate}: \textbf{15/18} proven optimal, \textbf{0 mismatch}, all match
published SteinLib; mean prune \(15.3\%\); 3 unproven (generic ILP gap, not certificate error).}\\
\bottomrule\end{tabular}}
\end{table}

\subsection{E7, E8, E10: interface contrasts}
\begin{figure}[t]\centering
\includegraphics[width=\textwidth]{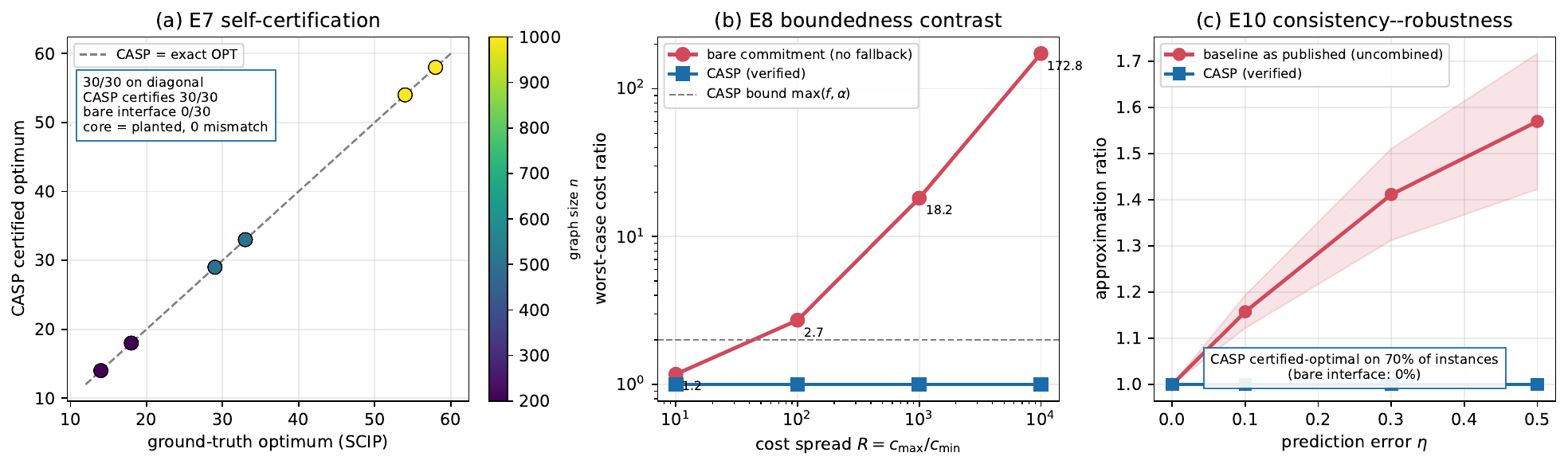}
\caption{The two interface contrasts (Section~\ref{sec:sep}). \textbf{(a)} E7: on \(30\) planted instances, each with a fractional
whole-graph LP optimum, CASP's
certified optimum equals the SCIP ground truth (diagonal, \(0\) mismatch) and is accompanied by a
polynomial-time-checkable proof; the bare-interface baseline emits none by construction
(Proposition~\ref{prop:planted}(iii)). \textbf{(b)} E8: CASP loss stays bounded by
\(\max(f,\alpha)\) while the bare commitment (no fallback) grows as \(\Theta(R)\) to
\(172.8\times\) (Theorem~\ref{thm:sepB}). \textbf{(c)} E10: the ratio of the uncombined baseline
degrades with the noise rate \(\eta\) (band = \(\pm1\) s.d.); CASP is flat and certifies \(70\%\)
of instances.}\label{fig:sep}
\end{figure}

Figure~\ref{fig:sep} measures the interface contrasts of Section~\ref{sec:sep}; the capability that
survives fair adversaries is measured in Section~\ref{sec:cfexp}. \textbf{E7 (certified optimality,
Fig.~\ref{fig:sep}(a)).} On the planted family \(G(n,k,H)\) of
Proposition~\ref{prop:planted}, the realized NT core equals the planted hard core on all \(30/30\)
instances, and CASP returns an optimum with a checker-validated proof on all \(30\); every CASP
optimum coincides with the SCIP ground truth, all points on the diagonal with zero mismatch. The
whole-graph LP optimum is fractional on every instance because the planted core sits at
$\frac12$, so no LP-integrality shortcut applies; the emergence comes from the certificate
trigger rather than from LP-integral easy inputs. The bare-interface baseline of
\citet{antoniadis} certifies optimality on \(0/30\) even when its output happens to be optimal ---
as it must, since its interface returns a solution with no proof object
(Proposition~\ref{prop:planted}(iii)); by Corollary~\ref{cor:noeff} the same baseline equipped with
our checker would certify all \(30\). E7 therefore isolates the value of the checking layer rather than any computational gap between
paradigms.

\textbf{E8 (boundedness, Fig.~\ref{fig:sep}(b)).} On the cost-spread family, CASP's worst-case cost
ratio is pinned at \(1.0\), strictly under the problem constant \(\max(f,\alpha)\), for every
spread \(R\); the structurally identical bare commitment (deliberately run \emph{without} a
fallback, per Definition~\ref{def:pos}) grows linearly,
\(1.17\to2.72\to18.2\to172.8\) as \(R{:}10\to10^4\). Because the two policy classes have
\emph{identical} pseudo-dimension \(O(\log K)\), this gap is not about expressiveness. It is
the verifier rendering the loss uniformly bounded, which controls the \(B^2\) factor of the
uniform-convergence rate~\eqref{eq:uc} behind the bound~\eqref{eq:Nc}. A min-combiner would
flatten the curve at \(\alpha\) (Theorem~\ref{thm:collapse}); boundedness is supplied by the
checking-and-fallback layer, and E8 shows what its absence costs. The empirical \(\Theta(R)\)
growth of the unbounded class matches the \(O(R^2)\) uniform-convergence range gap of
Theorem~\ref{thm:sepB} and the \(\Omega(R/\varepsilon^2)\) sample-complexity floor of
Theorem~\ref{thm:spreadlb}. Note that \(\max(f,\alpha)\) is a problem-family constant rather
than a universal one. On families where \(f\) grows, CASP's own bound grows with it, and the
contrast would have to be renormalized.

\textbf{E10 (head-to-head, Fig.~\ref{fig:sep}(c)).} Sweeping the flip-noise rate \(\eta\) on real
Vertex Cover graphs, the ratio of \citet{antoniadis} \emph{as published} (no combiner) rises
monotonically from \(1.00\) to \(1.57\) (with widening variance), while CASP stays at \(1.0\): its
correctness is decoupled from prediction quality (Theorem~\ref{thm:robust}). The fairness caveat
is explicit. A min-combined variant of the baseline flattens near the fallback ratio
(Theorem~\ref{thm:collapse}), and E10$'$ verifies this point for point on a three-arm rerun
(Section~\ref{sec:e10p}), so E10 measures the decoupling rather than superiority over combined
baselines; the comparison against the combiner itself is E12, where the filter wins by the margin
of Theorem~\ref{thm:margin}. On the same instances CASP remains certified-optimal on \(70\%\)
of this real-VC family, a set distinct from the 97-graph DIMACS/SNAP collection of E1, which
explains the different rate; the instance manifest is released with the artifact. The bare
baseline certifies none, per its interface. Together the three panels show the
checking-and-fallback layer buying decoupled correctness, certified optimality, and bounded-loss
learnability; E12 then shows a finer use of the same verifier that no combiner matches.

\subsection{E10$'$: a fair three-arm rerun}\label{sec:e10p}
\begin{figure}[t]\centering
\includegraphics[width=.62\linewidth]{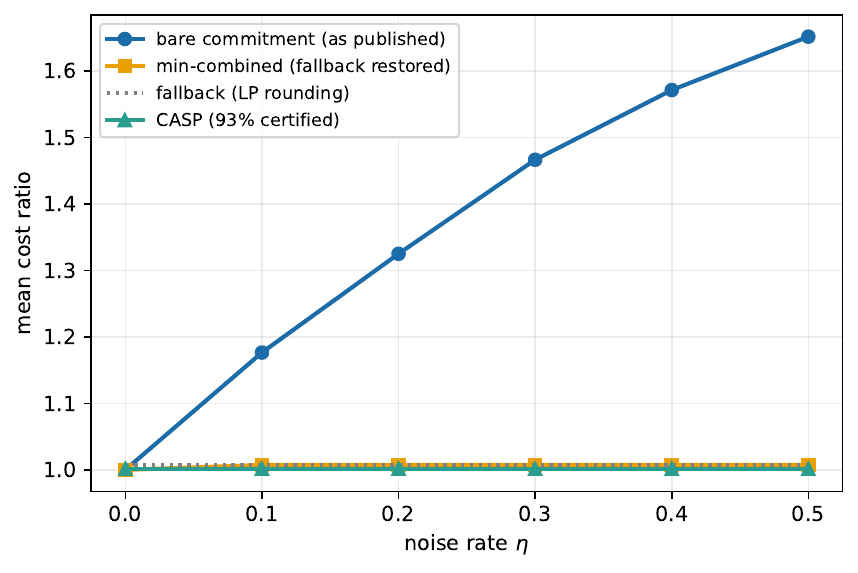}
\caption{\textbf{E10$'$ (fair rerun).} Three arms on the same instances: the bare commitment
degrades with noise, its min-combined variant is flat at the fallback ratio
(Theorem~\ref{thm:collapse}), CASP is flat at $1.002$ and certifies $93\%$.}\label{fig:e10fair}
\end{figure}

We apply the fairness standard of Section~\ref{sec:collapse} to E10 itself. On $30$ synthetic VC
instances and $\eta\in\{0,\dots,0.5\}$, the baseline of \citet{antoniadis} as published
degrades from $1.000$ to $1.651$, while the same output min-combined with the LP-rounding
fallback is flat at $1.007$, matching the collapse of Theorem~\ref{thm:collapse} point for point;
CASP sits at $1.002$ with a checker-validated optimality proof on $93\%$ of instances
(Figure~\ref{fig:e10fair}). E10 and E10$'$ together measure the bare interface's degradation and
the fact that a fallback rather than the sign of the signal removes it. What the combiner still
lacks is the certificate, and what neither single-threshold policy supplies is the
degeneracy-aware filtering of E12$'$.

\subsection{E12: confidence filtering}\label{sec:cfexp}
\begin{figure}[t]\centering
\includegraphics[width=\textwidth]{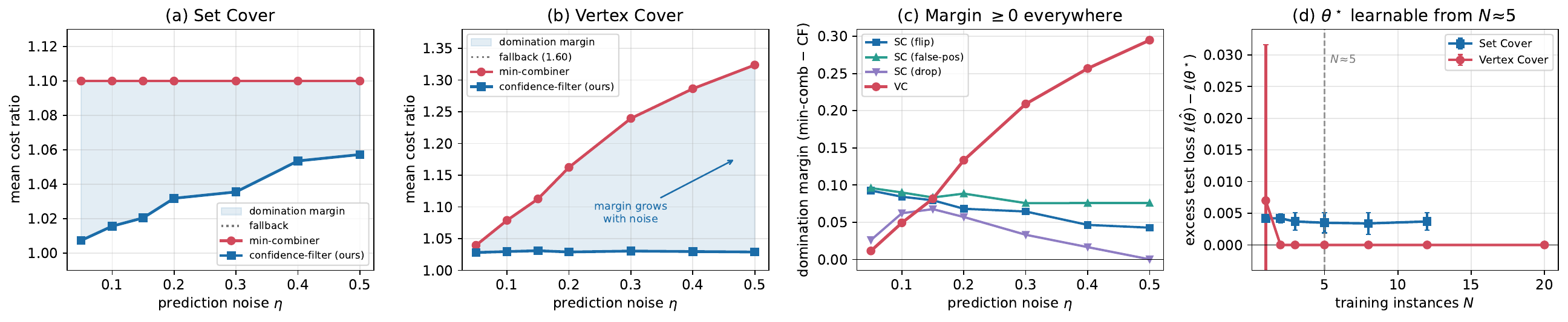}
\caption{\textbf{E12: verifiable confidence filtering dominates the min-combiner.} \textbf{(a)} On Set
Cover the min-combiner is pinned at the fallback ($1.10$ at every noise level $\eta$) while the
confidence-filter stays near-optimal; the shaded band is the domination margin. \textbf{(b)} On Vertex
Cover the confidence-filter is nearly noise-invariant ($\approx1.03$) while the min-combiner degrades
toward the fallback, so the margin \emph{grows} with $\eta$ (to $0.29$ at $\eta{=}0.5$). \textbf{(c)}
The domination margin is non-negative at every noise level across two problems and three Set Cover
noise models (flip, false-positive, drop), the weak domination of Theorem~\ref{thm:cfilter}(ii),
strict wherever the verifiable confidence is informative. \textbf{(d)} The excess test loss of the ERM threshold $\hat\theta$ over the in-hindsight optimum $\theta^\star$ vanishes by $N\!\approx\!5$ training instances on both problems, matching Theorem~\ref{thm:cfilter}(iv): $\theta^\star$ is a single scalar of pseudo-dimension $O(\log K)$. The wide interval at $N{=}1$ on
Vertex Cover reflects that a single training instance occasionally selects a suboptimal $\theta$;
from $N{=}2$ the optimum is recovered exactly.}\label{fig:cf}
\end{figure}

We test Theorems~\ref{thm:cfilter} and~\ref{thm:margin} on Set Cover and Vertex Cover under noisy solution predictions
(membership perturbed at rate $\eta$; confidence $\sigma_i$ is the LP value the verifier already
computes). We compare the \emph{min-combiner} (commit-all, then min with fallback) against the
\emph{confidence-filter} with a single ERM-learned threshold $\theta^\star$ (Figure~\ref{fig:cf} and Table~\ref{tab:cf}).
On Set Cover the min-combiner never improves on the fallback, staying at $1.10$ for every $\eta$
because committing the noisy prediction wholesale is always worse, while the confidence-filter
holds $1.01$--$1.06$. The domination margin of $0.04$--$0.09$ is positive at every noise level
and under all three noise models, flip, false-positive-heavy, and drop-heavy, which is the weak
domination of Theorem~\ref{thm:cfilter}(ii). On Vertex Cover the effect is larger and grows with
noise. The confidence-filter is nearly noise-invariant at about $1.03$ for all $\eta$ while the
min-combiner degrades from $1.04$ to $1.32$, so the margin rises to $\mathbf{0.29}$ at
$\eta=0.5$. This is the profile computed on $D_{\beta,C}$ in Theorem~\ref{thm:margin}, a
near-invariant filter, a degrading combiner, and a margin growing toward the fallback gap; at
$\beta=1$ and $C=2$ the closed form gives $0.2917$ at $\eta=0.5$. The learned threshold is a
single scalar recovered from $N\!\approx\!5$ instances with excess loss below $0.002$. The
learning is deliberately cheap; what is being tested is the strict domination of the min-combiner
by a verifiable filter.

\begin{table}[t]\centering\small
\caption{\textbf{E12}: mean cost ratio under noisy predictions. The confidence-filter (learned scalar
$\theta^\star$ on the verifiable LP value) \emph{dominates} the min-combiner at every noise level
$\eta$; on Vertex Cover the margin grows with $\eta$ (Theorem~\ref{thm:cfilter}).}\label{tab:cf}
\begin{tabular}{clccc}\toprule
Problem & \(\eta\) & min-combiner & confidence-filter & margin\\\midrule
Set Cover & 0.05 & 1.10 & 1.01 & 0.09\\
Set Cover & 0.20 & 1.10 & 1.03 & 0.07\\
Set Cover & 0.50 & 1.10 & 1.06 & 0.04\\\midrule
Vertex Cover & 0.05 & 1.04 & 1.03 & 0.01\\
Vertex Cover & 0.20 & 1.16 & 1.03 & 0.13\\
Vertex Cover & 0.50 & 1.32 & 1.03 & \textbf{0.29}\\
\bottomrule\end{tabular}
\end{table}

\subsection{E12$'$: prediction-free baselines and CF$^+$}\label{sec:e12p}
E12 compared prediction-consuming policies. E12$'$ adds the missing control and asks whether the
filter's advantage comes from the prediction or from the LP the verifier already solves. On the same
distributions and noise models as E12 we add a prediction-free arm $\mathrm{LP}\theta$ (commit
$\{\sigma\ge\theta'\}$ with $\theta'$ learned by ERM over a grid containing $\theta'{=}1$, whose
fixed special case is the $A=\{1\}$ LP-commit policy of Definition~\ref{def:symlp}), and the
two-threshold filter $A^{\mathrm{cf+}}$ of Proposition~\ref{prop:cfplus} with jointly learned
$(\theta_1,\theta_2)$.

\begin{figure}[t]\centering
\includegraphics[width=\textwidth]{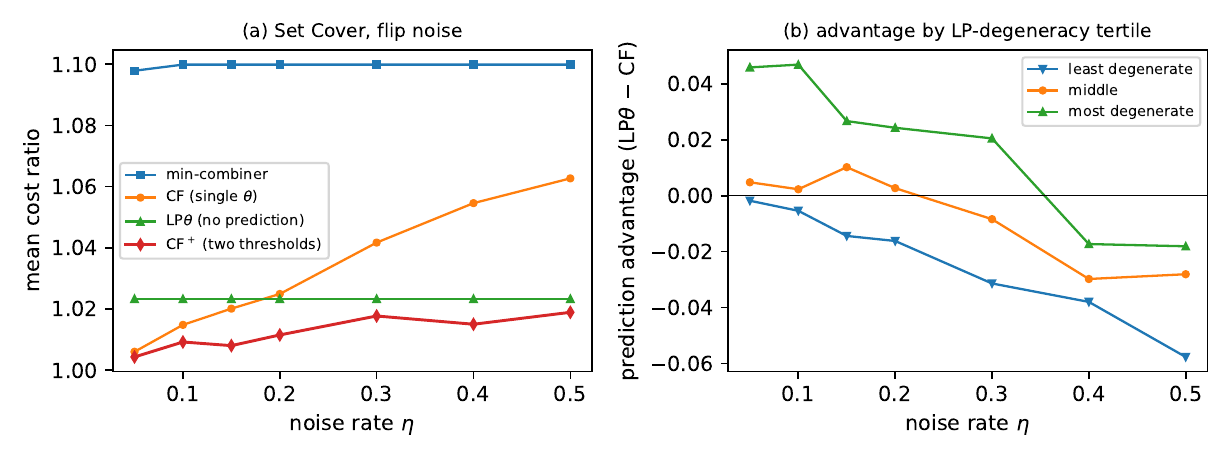}
\caption{\textbf{E12$'$.} \textbf{(a)} Set Cover, flip noise: the single-threshold filter (CF)
dominates the min-combiner everywhere but crosses below the prediction-free LP-threshold arm at
high noise; the two-threshold filter CF$^+$ (Proposition~\ref{prop:cfplus}) tracks the best of
both and improves strictly on 18 of 21 (model,$\eta$) points. \textbf{(b)} The prediction's added
value by LP-degeneracy tertile: it concentrates on degenerate instances, the empirical counterpart
of Theorem~\ref{thm:lponly} and Proposition~\ref{prop:degen}.}\label{fig:cfplus}
\end{figure}

Three findings (Table~\ref{tab:e12p}, Figure~\ref{fig:cfplus}). \emph{First}, the domination of the
min-combiner (Theorems~\ref{thm:cfilter},~\ref{thm:margin}) is untouched: $\mathrm{CF}\le$
min-combiner at every point. \emph{Second}, the prediction's added value over the best
prediction-free arm has a regime structure. It is positive at low noise, $+0.013$ at
$\eta{=}0.05$ for Set Cover under flip noise, and negative at high noise, where the
single-threshold filter cannot express discarding the prediction while still committing high-LP
variables. Stratifying test instances by LP degeneracy, the fraction of fractional $x^\star$
values, the advantage concentrates where Theorem~\ref{thm:lponly} predicts, on the most
degenerate tertile: $+0.046$ against $-0.002$ on the least degenerate at $\eta{=}0.05$, and
$+0.021$ against $-0.031$ at $\eta{=}0.30$. On Vertex Cover, $G(45,0.09)$ with low degeneracy
throughout, filter and LP-commit tie at about $1.03$ for every $\eta$, the empirical face of
Proposition~\ref{prop:degen}. \emph{Third}, the two-threshold filter repairs the crossover.
With a jointly ERM-learned pair it beats the best single-threshold arm strictly at $18$ of the
$21$ model-noise points, ties at two, and trails by at most $0.0013$ at the remaining one, drop
noise at $\eta{=}0.5$, within ERM generalization slack. The improvement is a union effect rather
than a maximum; under flip noise at $\eta{=}0.30$ within the CF$^+$ run, the min-combiner scores
$1.100$, CF $1.042$, LP$\theta$ $1.023$, and CF$^+$ $\mathbf{1.018}$.

\begin{table}[t]\centering\small
\caption{E12$'$ (Set Cover, flip noise; $30$ instances, train/test split): mean cost ratio per arm
and the prediction's added value over the best prediction-free arm
(adv $=$ LP$\theta$ $-$ CF, computed within one run before rounding), overall and by LP-degeneracy
tertile of the test instances. CF$^+$ is the two-threshold filter of
Proposition~\ref{prop:cfplus}, learned in a separately seeded run of the same protocol (its own
train/test split); all other columns come from a single run.}\label{tab:e12p}
\begin{tabular}{cccccccc}\toprule
$\eta$ & min-comb & CF & LP$\theta$ & CF$^+$ & adv (all) & adv (least deg.) & adv (most deg.)\\\midrule
0.05 & 1.098 & 1.010 & 1.023 & 1.004 & $+0.013$ & $-0.002$ & $+0.046$\\
0.15 & 1.100 & 1.019 & 1.023 & 1.008 & $+0.005$ & $-0.014$ & $+0.027$\\
0.30 & 1.100 & 1.033 & 1.023 & 1.018 & $-0.010$ & $-0.031$ & $+0.021$\\
0.50 & 1.100 & 1.061 & 1.023 & 1.019 & $-0.037$ & $-0.058$ & $-0.018$\\
\bottomrule\end{tabular}
\end{table}

\subsection{E14: recomputing the closed forms}\label{sec:e14}
\begin{figure}[t]\centering
\includegraphics[width=\textwidth]{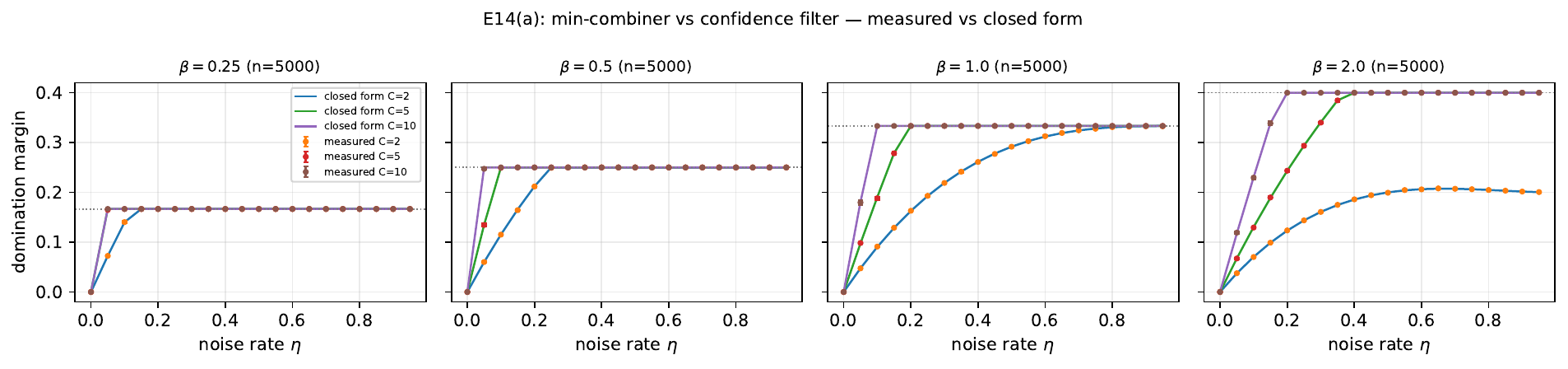}
\caption{\textbf{E14(a): Theorem~\ref{thm:margin} recomputed by Monte Carlo.} Measured
min-combiner-vs-filter margins (points, $\pm3$ s.e.) against the closed form
$\min\{\eta[(C-\eta)+\beta(1-\eta)^2],\beta\}/(1+2\beta)$ (curves), per $\beta$ panel at
$n{=}5000$; the dotted line is the fallback-gap cap
$\beta/(1+2\beta)$ (an upper bound on the margin, attained only where the unsaturated branch
reaches $\beta$).}\label{fig:margin}
\end{figure}

\begin{figure}[t]\centering
\includegraphics[width=\textwidth]{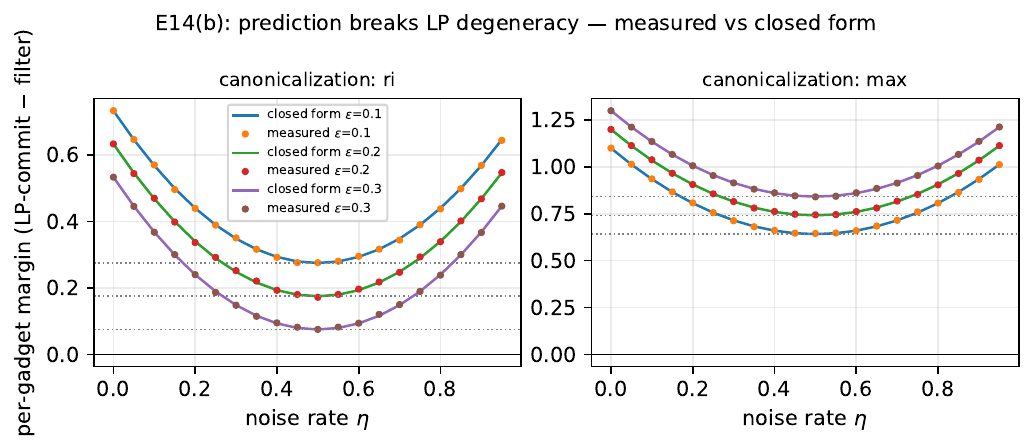}
\caption{\textbf{E14(b): Theorem~\ref{thm:lponly} recomputed by Monte Carlo.} Measured per-gadget
margin of the filter over the LP-commit policy (points) against the closed form (curves) under
both $\sigma$-canonicalizations; dotted lines are the proven floors, never
crossed.}\label{fig:lponlyfig}
\end{figure}

Theorems~\ref{thm:margin} and~\ref{thm:lponly} make numerical predictions, which admits a
strong form of validation, namely Monte-Carlo re-derivation of the closed forms on their own
families, grid point by grid point (Figures~\ref{fig:margin} and~\ref{fig:lponlyfig}).
\textbf{Margin (Theorem~\ref{thm:margin}).} Across the full grid
($n\in\{200,1000,5000\}$, $\beta\in\{0.25,0.5,1,2\}$, $C\in\{2,5,10\}$, twenty noise levels,
$50$ seeds each, $720$ configurations in all), the filter is surely optimal in every
configuration, a per-realization assertion rather than an average, and the measured min-combiner
margin converges to the closed form at the predicted $O(1/\sqrt n)$ rate, with maximum absolute
deviation $0.033$ at $n{=}200$, $0.0065$ at $n{=}1000$, and $0.0027$ at $n{=}5000$. In the
saturated regime, $\beta{=}1$ with $C\in\{5,10\}$ at large $\eta$, the measurement is
$0.33333$ against the theoretical $1/3$.
\textbf{Degeneracy (Theorem~\ref{thm:lponly}).} Across the corresponding grid on
$H_{m,n}(\varepsilon)$ ($720$ rows, both $\sigma$-canonicalizations of
Remark~\ref{rem:canon}), the measured per-gadget margin matches the closed form to at most
$0.018$, is positive at every noise level, and never crosses the proven floors
$\frac38-\varepsilon$ (relative-interior) and $1+\varepsilon-\frac{11}{24}$ (max), with zero
violations. E14 is the most direct check in the paper; rather than qualitatively matching the
theorems, the experiments recompute their closed forms.

\subsection{E13: learned predictors}\label{sec:e13}
All predictions so far were synthetic. E13 trains real predictors and deploys them through the
framework, testing the two central claims that the verifier makes imperfect learned predictions
safe and that their added value concentrates where Theorem~\ref{thm:lponly} says it must.

\begin{figure}[t]\centering
\includegraphics[width=\textwidth]{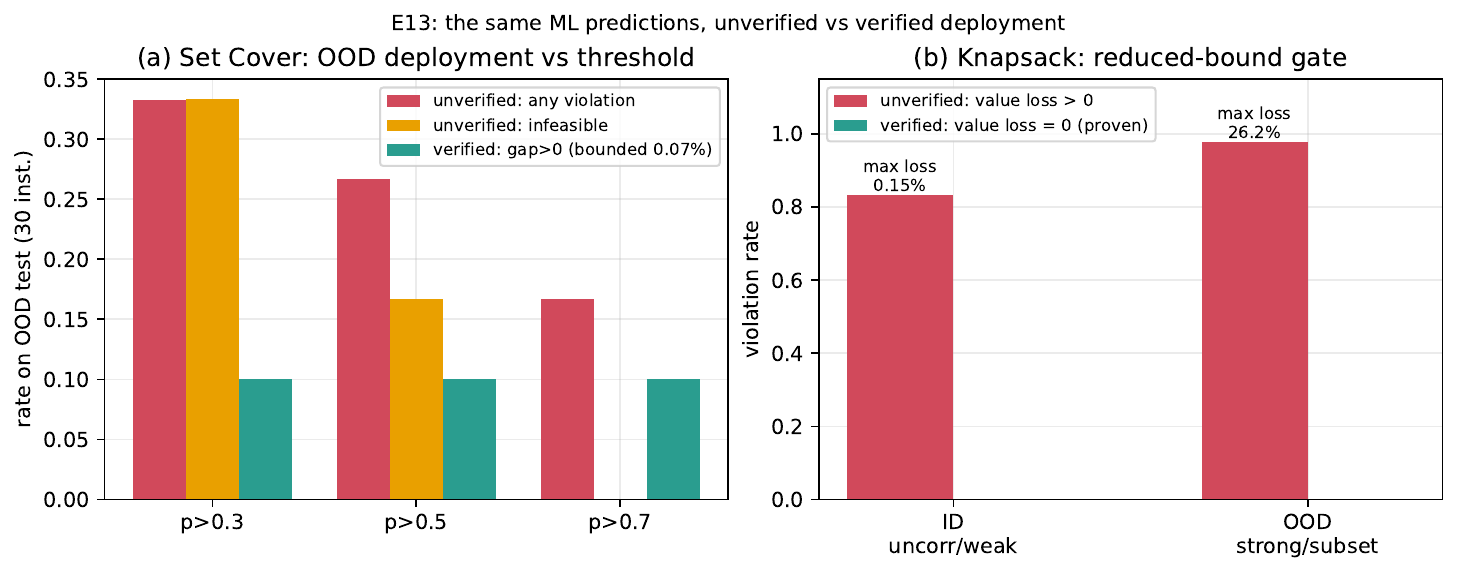}
\caption{\textbf{E13(a): verified vs unverified deployment of the same trained predictor under
distribution shift.} \textbf{(a)} Set Cover, OOD ($f{\to}10,20$): unverified deletion is
infeasible on up to $10/30$ instances depending on the prediction threshold; the verified arm is
never infeasible and its gap stays within the certified $f$-safe budget. \textbf{(b)} Knapsack:
unverified exclusion forfeits optimality frequently already in distribution ($83\%$ of instances,
but by ${\le}0.15\%$) and, OOD on correlated types, loses up to $26.2\%$ of the optimal value on
$97.9\%$ of instances; the reduced-bound gate loses nothing in either regime
(Theorem~\ref{thm:knap}), accepting almost no exclusions OOD.}\label{fig:verified}
\end{figure}

\paragraph{Verified vs.\ unverified ML pruning (Figure~\ref{fig:verified}).}
A gradient-boosted classifier predicts, per variable, ``appears in no optimum'' (six combinatorial
and LP features; trained on Set Cover $f{=}5$ instances, respectively on
uncorrelated/weakly-correlated Knapsack). The same predictions are deployed in two ways. The unverified deployment deletes every predicted
variable, following the ML-problem-reduction template \citep{sunernst,lauridutta}. The verified
deployment deletes only the predictions accepted by the CASP verifier, through the $f$-safe
LP-threshold gate for Set Cover and the OPT-preserving reduced-bound gate of
Theorem~\ref{thm:knap} for Knapsack. In distribution the two deployments are
indistinguishable on Set Cover (zero violations either way), while on Knapsack unverified
exclusion already forfeits optimality on $83\%$ of instances, though only by ${\le}0.15\%$.
Under distribution shift the magnitudes separate sharply. On Set Cover with $f{\to}10,20$,
unverified pruning produces hard infeasibility on $10$, $5$, and $0$ of $30$ instances at
prediction thresholds $0.3$, $0.5$, and $0.7$, with violation rates, output infeasible or
suboptimal, of $33\%$, $27\%$, and $17\%$. The verified arm has zero infeasibility at every
threshold, a quality gap of at most $0.071\%$ inside the $f$-safe budget, and essentially all
the pruning ($0.931$ against $0.932$). On Knapsack, shifting from uncorrelated to strongly
correlated and subset-sum types, unverified exclusion loses optimality on $\mathbf{97.9\%}$ of
instances with value losses up to $\mathbf{26.2\%}$. The verified arm has zero loss on all $48$
instances, the OPT-preservation guarantee of Theorem~\ref{thm:knap} observed without exception, while its
acceptance rate collapses to $0.038$ and $0.000$ on the correlated types, the same certificate
silence as in E9. Verification thus costs coverage, and skipping it costs up to $26\%$ of the
optimal value.

\begin{figure}[t]\centering
\includegraphics[width=\textwidth]{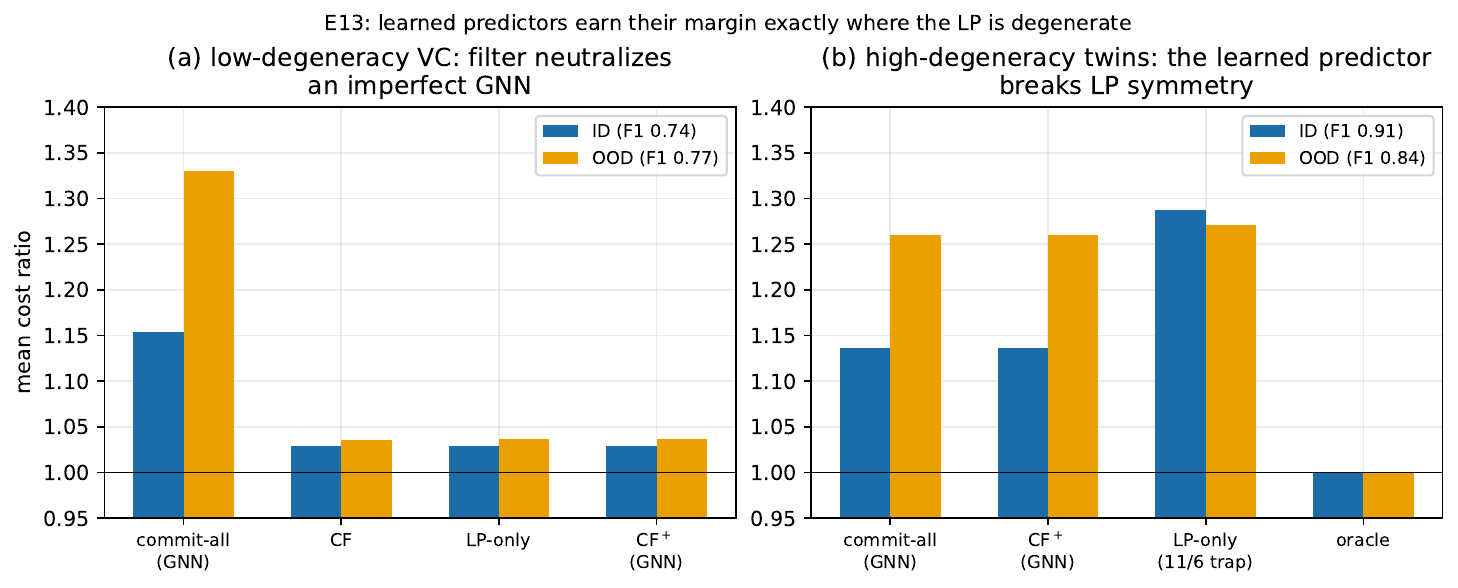}
\caption{\textbf{E13(b): learned predictors through the filter.} \textbf{(a)} Low-degeneracy VC:
an F1 $\approx0.75$ GNN committed wholesale costs $+15$--$33\%$; filtered, it ties LP-commit ---
the filter learns to ignore it (Proposition~\ref{prop:degen}). \textbf{(b)} High-degeneracy tagged
twins: the GNN reads side information the LP cannot express and CF$^+$ beats every LP-commit
policy by $+0.151$ (Theorem~\ref{thm:lponly}, learning-side).}\label{fig:learned}
\end{figure}

\paragraph{A GNN through the filter (Figure~\ref{fig:learned}).}
A hand-rolled message-passing GNN predicts VC membership at F1 about $0.74$--$0.77$ from purely
combinatorial features, degrees, neighbor statistics, and triangle counts, deliberately excluding
the LP values. Committed wholesale, this imperfect predictor costs $1.154$ in distribution and
$1.331$ out of distribution on $G(80)$; through CF and CF$^+$ it is pinned back to $1.029$ and
$1.037$, tying the LP-commit policy. On this low-degeneracy distribution the learned threshold
settles at $\theta{=}1.0$, so the filter learns to trust only the LP, the behavior
Proposition~\ref{prop:degen} predicts.
On a high-degeneracy family the picture inverts. On tagged twin gadgets, the $H$-family of
Theorem~\ref{thm:lponly} in which the generator marks the historically chosen twin with an
observable tag the LP cannot see, a bipartite GNN reaches F1 $0.911$, and CF$^+$ attains $1.137$
against the LP-commit policy's $1.288$, the $11/6$ harmonic trap; the learned advantage is
$+0.151$ per unit OPT, and $+0.011$ under $30\%$ tag noise. Together the two panels show a
learned predictor earning its margin where the LP is degenerate and the verifier making it
deployable everywhere else.

\subsection{Discussion and limitations}
The experiments track the theory closely. The interface contrasts appear as predicted and for
the predicted, interface-bound reason (E7, E8); prediction-independent safety is confirmed end to
end (E5, E10); and the confidence-filter margin has the shape computed in
Theorem~\ref{thm:margin}, zero at zero noise, growing, and capped at the fallback gap (E12). The OPT-preserving
certificates are \emph{never} observed to violate optimality (E1, E3, E9), with verification done
by an independent checker. Three limitations temper the speedup results. First, the certificates are silent on inputs
lacking the relevant structure: NT on dense random and real graphs, firing on \(12\) of
\(97\); the Knapsack certificate on correlated instances; the Steiner certificate beyond the
ILP-solvable B-class; and the FL threshold on the original CAP library. This is the price of
soundness rather than a defect.
Second, the headline speedups are largest when the solver bottleneck is the number of variables; at rail scale the certificate still prunes $99.5\%$ and yields a net $28\times$ end-to-end gain at a ${\sim}1\%$ realized gap (E6$''$), but its certified $f$-safe factor is vacuous there; the open problem is not pruning power but a certificate whose guarantee remains meaningful at large $f$. Third, the proven pruning-rate floor (E2) and the worst-case approximation bounds
(E6) are loose on real instances, so the empirical gains exceed what the bounds guarantee; this
is good in practice but a sign that the analysis can be tightened. On the planted E7 instances SCIP's presolve already solves quickly, so the value of CASP there is the checkable optimality proof rather than wall-clock speedup. Speedup is claimed only where exact solving is genuinely hard; the presolve audit and the hard-regime measurements inside E6/E6$'$ delimit exactly where that is. On validity, results use one solver, SCIP 10.0 via PySCIPOpt 6.2.1, and the E4$'$ distribution
is benign, so there we present only the ordering in \(p\); E4$''$ removes this caveat on a
heavy-tailed distribution. A further limitation follows from Proposition~\ref{prop:degen} and is confirmed by E12$'$. The
confidence filter earns its margin only where the LP confidence is dispersed; on the
low-degeneracy VC distribution the prediction adds nothing over LP-commit, and at high noise the
single-threshold filter loses to a prediction-free arm unless upgraded to the two-threshold
filter of Proposition~\ref{prop:cfplus}. This motivates predictor-supplied, verifier-checked
confidence as future work. None of these caveats touch the boundedness theory, the
collapse analysis (Section~\ref{sec:collapse}), or the quantitative filtering guarantees
(Theorems~\ref{thm:margin},~\ref{thm:lponly}), which are the paper's core contribution and rest on
exact computations on explicit constructions rather than on favorable instances.

\section{Conclusion}
CASP inverts the information flow of prediction-augmented algorithms, asking the predictor for
\emph{verifiable certificates of what may be ignored} rather than \emph{what to do}, and thereby
brings predictions to offline NP-hard approximation with prediction-independent safety. The
central results are quantitative and belong to the checking layer. The verifier makes the
learnable loss class uniformly bounded where the bare interface provably admits no
distribution-free rate (Theorem~\ref{thm:sepB}, Lemma~\ref{lem:heavytail}); a verifiable
confidence filter dominates the min-combiner with an exactly computed margin, zero at zero noise
and growing with small noise (Theorem~\ref{thm:margin}); and the prediction is provably not
redundant given the LP, since its role is to break the tie on a degenerate optimal face, which no
symmetric LP policy can do (Theorem~\ref{thm:lponly}). We have been equally explicit about what
checking does not buy. Certified optimality and the boundedness gap are interface contrasts that
a fair min-combiner adversary matches (Section~\ref{sec:collapse}), and the filter degenerates
on LP-opaque cores (Proposition~\ref{prop:degen}). Empirically the prediction's advantage
concentrates on LP-degenerate instances as Theorem~\ref{thm:lponly} predicts, and the
two-threshold filter of Proposition~\ref{prop:cfplus} carries it across the noise range
(E12$'$).

With trained predictors the same pattern appears end to end. An imperfect GNN is neutralized by
the filter on low-degeneracy inputs and earns $+0.15$ on degenerate ones, and unverified ML
pruning loses up to $26\%$ under distribution shift where the verified deployment loses nothing
(E13). We instantiated the framework on five structurally diverse problems by recasting classical
reductions as verifiable certificates, made emergent exact solvability a triggerable algorithm on
a named class and a planted distribution, and generalized the PAC theory to multi-parameter
certificate classes. Several directions remain open: certificate classes whose guarantees survive
large frequencies, as in the rail regime of E6$''$ where behavior is excellent but the $f$-safe
bound is vacuous; OPT-preserving certificates for non-integral Set Cover; predictor-supplied,
verifier-checked confidence for LP-opaque cores; and learned, adversarially robust certificate
generators evaluated against unverified ML problem reduction \citep{sunernst,lauridutta}.

\paragraph{Code and data availability.}
All experiment code, the synthetic-instance generators together with the generated instances
(Set Cover, Vertex Cover, Facility Location, Knapsack, and the adversarial set), per-instance
manifests and checksums for every benchmark used (OR-Library Set Cover and rail, DIMACS/SNAP
graphs, ORLIB-cap / UFLLib / TSPLIB Facility Location, SteinLib B-class, Pisinger Knapsack),
the reimplemented positive-signal baseline of \citet{antoniadis} used in E7/E10/E10$'$, solver
configurations (SCIP~10.0 via PySCIPOpt~6.2.1, seeds included), the independent verifier
reimplementation, and the $\Cl$ checker of Theorem~\ref{thm:sepA} are packaged as a versioned
artifact, available at \url{https://github.com/llfuture/casp}. The public benchmarks remain available from their
cited sources; the artifact mirrors instance files only where the source license permits and
otherwise ships download manifests with checksums.


\bibliographystyle{plainnat}
\bibliography{casp}

\appendix
\section{Proofs}\label{app:proofs}
The statements appear in the main text; each proof is given below in its own subsection.

\subsection{Proof of Lemma~\ref{lem:mono}}
Let $S$ be optimal for $I_\phi$. By Definition~\ref{def:legal}, $S\cup F$ is feasible in $I$ with
cost
\[
  c(S)+c_{\mathrm{fix}}=\OPT(I_\phi)+c_{\mathrm{fix}}\ge\OPT(I).
\]
Block pruning has
$F=\varnothing$, $c_{\mathrm{fix}}=0$, and $S$ itself is feasible in $I$.
\hfill\ensuremath{\square}

\subsection{Proof of Theorem~\ref{thm:optpres}}
$\ge$ by Lemma~\ref{lem:mono}. $\le$, block case: an optimum $S^\star\subseteq\G(I_\phi)$ is
feasible in $I_\phi$ (it covers $U(I_\phi)=U(I)$) at cost $\OPT(I)$, so $\OPT(I_\phi)\le\OPT(I)$.

$\le$, fix--reduce case: Definition~\ref{def:safe}(1) supplies an optimum $S^\star\supseteq F$
with $S^\star\setminus F\subseteq\G(I_\phi)$ and $S^\star\setminus F$ feasible in $I_\phi$, at
cost $c(S^\star)-c_{\mathrm{fix}}=\OPT(I)-c_{\mathrm{fix}}$; hence
$\OPT(I_\phi)+c_{\mathrm{fix}}\le\OPT(I)$, and Lemma~\ref{lem:mono} makes it an equality. (For
the NT certificate the feasibility clause is exactly the structure lemma inside the proof of
Theorem~\ref{thm:nt}: every edge of the core has both ends in $P_{1/2}$, so it is covered by
$S^\star\setminus P_1$.)
\hfill\ensuremath{\square}

\subsection{Proof of Theorem~\ref{thm:compose}}
We have $\OPT(I_{\phi_1})\le\rho_1\OPT(I)$, hence
\[
  \OPT(I_{\phi_1,\phi_2})\le\rho_2\,\OPT(I_{\phi_1})\le\rho_1\rho_2\,\OPT(I);
\]
the chain
$\G(I_{\phi_1,\phi_2})\subseteq\G(I_{\phi_1})\subseteq\G(I)$ transfers feasibility. For
fix--reduce prunings the same chain holds with the fixed costs carried along:
$\OPT(I_{\phi_1,\phi_2})+c^{(2)}_{\mathrm{fix}}\le\rho_2\,\OPT(I_{\phi_1})$, hence, using
$\rho_2\ge1$,
$\OPT(I_{\phi_1,\phi_2})+c^{(1)}_{\mathrm{fix}}+c^{(2)}_{\mathrm{fix}}
\le\rho_2\big(\OPT(I_{\phi_1})+c^{(1)}_{\mathrm{fix}}\big)\le\rho_1\rho_2\,\OPT(I)$.

Sequential OPT-preservation (block pruning): Theorem~\ref{thm:optpres} applied twice gives
$\OPT(I_{\phi_1,\phi_2})=\OPT(I_{\phi_1})=\OPT(I)$; any optimum of $I_{\phi_1,\phi_2}$ is
feasible in $I$ (block prunings leave $U$ unchanged) at cost $\OPT(I)$, hence is an optimum of
$I$ contained in $\G(I_{\phi_1,\phi_2})$, which is OPT-preservation of the composite.

Simultaneous application preserving the same $S^\star$: then
$S^\star\subseteq\G(I_{\phi_1,\phi_2})$ and Theorem~\ref{thm:optpres} applies.
\hfill\ensuremath{\square}

\subsection{Proof of Theorem~\ref{thm:robust}}
Branch one (valid certificate, reduced instance feasible, $\mathrm{Solve}$ succeeds): let $S_0$ be
the reduced solution and $S=S_0\cup F$ the returned solution ($F=\varnothing$,
$c_{\mathrm{fix}}=0$ for block pruning). Then $c(S_0)\le\alpha_{\mathrm{red}}\,\OPT(I_\Phi)$ and,
using $\alpha_{\mathrm{red}}\ge1$ and $\rho$-safety in the form
$\OPT(I_\Phi)+c_{\mathrm{fix}}\le\rho\,\OPT(I)$ (Definition~\ref{def:safe}),
\[
  c(S)=c(S_0)+c_{\mathrm{fix}}\le\alpha_{\mathrm{red}}\,\OPT(I_\Phi)+c_{\mathrm{fix}}
  \le\alpha_{\mathrm{red}}\big(\OPT(I_\Phi)+c_{\mathrm{fix}}\big)\le\rho\,\alpha_{\mathrm{red}}\,\OPT(I);
\]
$S$ is feasible in $I$ by legality (Definition~\ref{def:legal}).

Branch two (no valid certificate, infeasible after pruning, or timeout): return
$\mathcal{A}(I)$, $c(S)\le\alpha\OPT(I)$. Take the max.
\hfill\ensuremath{\square}

\subsection{Proof of Theorem~\ref{thm:exact}}
Joint OPT-preservation means some optimum of $I$ survives the composite pruning, so
Theorem~\ref{thm:optpres}, applied to the composite, gives $\OPT(I_\Phi)=\OPT(I)$ (block case;
the fix--reduce identity carries $c_{\mathrm{fix}}$); the hypothesis is discharged either by the
sequential branch of Theorem~\ref{thm:compose} or because each certificate preserves every
optimum. The number of feasible solutions of $I_\Phi$ is at most
\[
  2^{|\G(I_\Phi)|}=2^{O(\log|\G(I)|)}=\mathrm{poly}(|\G(I)|);
\]
brute force (or exact IP) finds the
optimal $S^\star$ of $I_\Phi$ in polynomial time (each subset is checked by the
polynomial-time feasibility predicate of Section~\ref{sec:framework}). $S^\star$ is feasible in
$I$ at cost $\OPT(I_\Phi)=\OPT(I)$; for fix--reduce certificates return $S^\star\cup F$, whose
cost is $\OPT(I_\Phi)+c_{\mathrm{fix}}=\OPT(I)$ by Theorem~\ref{thm:optpres}.
\hfill\ensuremath{\square}

\subsection{Proof of Theorem~\ref{thm:noisy}}
Wrong certificates fail the verifier; infeasible pruning triggers fallback ($\le\alpha\OPT$); valid
certificates fall under Theorem~\ref{thm:robust} branch one ($\le\rho\alpha_{\mathrm{red}}\OPT$).
Take the max; the bound is independent of $\eta$.
\hfill\ensuremath{\square}

\subsection{Proof of Lemma~\ref{lem:scfeas}}
For each element $e$,
\[
  \sum_{S\ni e}x^\star_S\ge1 ,
\]
and at most $f$ sets contain $e$, so by averaging some
$S\ni e$ has $x^\star_S\ge1/f\ge\tau$ and survives; $e$ remains covered.
\hfill\ensuremath{\square}

\subsection{Proof of Theorem~\ref{thm:lpsafe}}
Classical LP rounding. Let $\hat S=\{S:x^\star_S\ge1/f\}$; by Lemma~\ref{lem:scfeas} it covers $U$.
Its cost satisfies
\begin{equation}\label{eq:fsafe}
  c(\hat S)=\!\!\sum_{S:\,x^\star_S\ge1/f}\!\!c_S
  \ \le\ f\!\!\sum_{S:\,x^\star_S\ge1/f}\!\!c_S x^\star_S
  \ \le\ f\sum_S c_S x^\star_S=f\,\LP(I)\le f\,\OPT(I).
\end{equation}
Every $S\in\hat S$ has $x^\star_S\ge1/f\ge\tau$, so it
survives in $I_\tau$; hence $\OPT(I_\tau)\le c(\hat S)\le f\OPT(I)$.
\hfill\ensuremath{\square}

\subsection{Proof of Theorem~\ref{thm:cscs}}
Complementary slackness gives
\[
  x^\star_S>0\ \Longrightarrow\ \sum_{e\in S}y^\star_e=c_S ,
\]
whose contrapositive gives
$\phi_{CS}(S)\Rightarrow x^\star_S=0$, i.e.\ the pruned sets lie outside $\supp(x^\star)$. Under
\Sup{} the integer optimum $S^\star\subseteq\supp(x^\star)$ uses no pruned set, so
$S^\star\subseteq\G(I_{\phi_{CS}})$. (``Unique LP optimum'' alone is insufficient: a unique but
fractional optimum only preserves the \emph{LP} value, not an integer optimum; \Sup{} is
required.)
\hfill\ensuremath{\square}

\subsection{Proof of Theorem~\ref{thm:nt}}
\emph{(i) Half-integrality} \citep{nt} yields the partition.

\emph{(ii) Structure lemma}: no edge
has both ends in $P_0$, and none has one end in $P_0$ and the other in $P_{1/2}$; else
$x^\star_u+x^\star_v\le0+\frac12<1$. So every neighbor of a $P_0$ vertex lies in $P_1$.

\emph{(iii) Persistency by perturbation--exchange}: let $S$ be any minimum integer cover,
$S'=(S\setminus P_0)\cup P_1$. \underline{$S'$ covers}: for edge $uv$, if an end is in $P_1$ it is
covered; else both ends are in $P_{1/2}$ (by (ii) $P_0$ is impossible), and $S$ covers $uv$ using a
$P_{1/2}$ vertex, retained in $S'$. \underline{$c(S')\le c(S)$}: we show $c(P_1\setminus S)\le
c(S\cap P_0)$. For small $\varepsilon\in(0,\frac12]$ perturb
$x_v=x^\star_v+\varepsilon$ on $S\cap P_0$ (value $\to\varepsilon$),
$x_v=x^\star_v-\varepsilon$ on $P_1\setminus S$ (value $\to1-\varepsilon$), else $x^\star_v$.
This $x$ is LP-feasible: for $v\in P_1\setminus S$ (value $1-\varepsilon$), $v\notin S$ so all its
neighbors $u\in S$; if $u\in P_1$ then $x_u\ge1-\varepsilon$, sum $\ge2-2\varepsilon\ge1$; if $u\in
P_{1/2}\cap S$, sum $=\frac12+1-\varepsilon\ge1$; if $u\in P_0\cap S$, $x_u=\varepsilon$, sum $=1$.
For $v\in S\cap P_0$ (value $\varepsilon$), by (ii) its neighbors are in $P_1$: if in $S$, $x_u=1$;
if in $P_1\setminus S$, $x_u=1-\varepsilon$, sum $=1$. Other edges are unaffected. By optimality of
$x^\star$, $c\cdot x\ge c\cdot x^\star$, i.e.
\[
  \varepsilon\big(c(S\cap P_0)-c(P_1\setminus S)\big)\ge0
  \ \Longrightarrow\ c(P_1\setminus S)\le c(S\cap P_0).
\]
Hence
\[
  c(S')=c(S)-c(S\cap P_0)+c(P_1\setminus S)\le c(S),
\] so $S'$ is a minimum cover with
$P_1\subseteq S'\subseteq P_1\cup P_{1/2}$.

\emph{(iv)} The residual is $G[P_{1/2}]$ and
$\OPT(G)=c(P_1)+\OPT(G[P_{1/2}])$; also $\LP(G)\ge\frac12 c_{\min}|P_{1/2}|$, so for unit weights
$|P_{1/2}|\le2\LP(G)\le2\OPT(G)$.
\hfill\ensuremath{\square}

\subsection{Proof of Theorem~\ref{thm:Cdet}}
Theorem~\ref{thm:nt} gives $\OPT(G)=c(P_1)+\OPT(G[P_{1/2}])$; brute-force the core over
\[
  2^{|P_{1/2}|}\le n^{c_0}\ \text{subsets, each checked in polynomial time.}
\]
The bound $|P_{1/2}|\le2\LP(G)$ is in the proof of
Theorem~\ref{thm:nt}.
\hfill\ensuremath{\square}

\subsection{Proof of Theorem~\ref{thm:flsafe}}
Since the survivor set grows as $\tau$ shrinks and $\OPT(I_\tau)$ is monotone in the survivor
set, it suffices to prove the bound at $\tau=1/\Delta$.

\emph{Feasibility}: for client $j$, $\sum_i x^\star_{ij}\ge1$ over its $\delta_j\le\Delta$ support
facilities and $x^\star_{ij}\le y^\star_i$; if every support facility had $y^\star_i<\tau$ then
\[
  1\le\sum_{i:\,x^\star_{ij}>0} x^\star_{ij}\le\sum_{i:\,x^\star_{ij}>0} y^\star_i<\delta_j\,\tau\le1,
\]
a contradiction; each client keeps a serving facility.

\emph{Cost}: WLOG $\sum_i x^\star_{ij}=1$ for every $j$ (excess mass can be removed without
increasing cost or breaking optimality). Because the assignment constraints decouple across
clients, $x^\star_{\cdot j}$ minimizes $\sum_i d_{ij}x_{ij}$ subject to $\sum_i x_{ij}\ge1$,
$0\le x_{ij}\le y^\star_i$; hence if $x^\star_{ij}>0$ and $d_{i'j}<d_{ij}$ then
$x^\star_{i'j}=y^\star_{i'}$, else moving mass from $i$ to $i'$ would improve the LP. Let $i_j$ be
the nearest surviving support facility of $j$. Every support facility strictly closer than $i_j$
is pruned and therefore carries mass $x^\star_{i'j}=y^\star_{i'}<\tau=1/\Delta$, and there are at
most $\delta_j-1\le\Delta-1$ of them; so the mass at distance $\ge d_{i_jj}$ exceeds
$1-(\Delta-1)/\Delta=1/\Delta$, giving
\[
  C_j:=\sum_i d_{ij}x^\star_{ij}\ \ge\ \frac{d_{i_jj}}{\Delta},
  \qquad\text{i.e.}\qquad d(j,\hat{F})\le d_{i_jj}\le\Delta\,C_j ,
  \qquad \hat F:=\{i:y^\star_i\ge1/\Delta\}.
\]
Open all of $\hat F$ and assign each client to its nearest open facility: the open cost is
$\sum_{i\in\hat F}f_i\le\Delta\sum_i f_iy^\star_i$ and the connection cost is at most
$\Delta\sum_jC_j$, so this integer solution, which uses only survivors, costs at most
$\Delta\,\LP(I)\le\Delta\,\OPT(I)$; hence $\OPT(I_\tau)\le\Delta\,\OPT(I)$. No triangle
inequality was used.
\hfill\ensuremath{\square}

\subsection{Proof of Theorem~\ref{thm:flint}}
With $y^\star$ fixed integral, the assignment subproblem $\min\sum_{ij}d_{ij}x_{ij}$ s.t.\
$\sum_i x_{ij}\ge1$, $0\le x_{ij}\le y^\star_i$ is solved by sending each client entirely to
$\arg\min_{i\in F^\star}d_{ij}$, value $C^\star=\sum_j\min_{i\in F^\star}d_{ij}$. Since $x^\star$ is
feasible for the fixed $y^\star$, $C^\star\le\sum_{ij}d_{ij}x^\star_{ij}$, so the integer solution
$(F^\star,\text{nearest})$ has cost
\begin{equation}\label{eq:flint}
  \sum_{i\in F^\star}f_i+C^\star\ \le\ \sum_i f_i y^\star_i+\sum_{ij}d_{ij}x^\star_{ij}=\LP(I).
\end{equation}
Being feasible it also costs $\ge\OPT(I)\ge\LP(I)$; the two inequalities force
$\sum_{i\in F^\star}f_i+C^\star=\OPT(I)=\LP(I)$. Thus it is an integer optimum using only open facilities, so closing
$y^\star_i=0$ facilities preserves an optimum. Verification: solve the LP, check facility-integral
and optimality, output nearest assignment --- all polynomial.
\hfill\ensuremath{\square}

\subsection{Proof of Theorem~\ref{thm:knap}}
Any feasible solution with $x_i=1$ has value $\le U_i$ (LP relaxation with the same forcing
dominates). If
\[
  U_i<z_{\mathrm{low}}\le\OPT ,
\]
then every $x_i=1$ solution has value $<\OPT$, hence is non-optimal; so every optimum has $x_i=0$ and excluding $i$ keeps all optima. The inclusion case is
symmetric. Verification recomputes the one-constraint LP bound in $O(n)$ and checks that the
exhibited packing realizes $z_{\mathrm{low}}$.
\hfill\ensuremath{\square}

\subsection{Proof of Proposition~\ref{prop:pdim1}}
Sort $\{x^\star_S\}$; as $\tau$ crosses a value exactly one block flips, so $\ell_\tau(I)$ has
$\le K$ breakpoints. If $\F$ pseudo-shatters $m$ instances with witnesses $r_i$, the sign vector
$(\mathrm{sign}(\ell_\tau(I_i)-r_i))_i$ has $\le mK+1$ constant pieces, so
\[
  2^m\le mK+1\ \Longrightarrow\ m=O(\log K).
\]
\hfill\ensuremath{\square}

\subsection{Proof of Theorem~\ref{thm:pac}}
$\ell_\theta\in[1,B]$, pseudo-dimension $d$; the Pollard-type uniform-convergence theorem for
bounded real-valued families gives the bound, and ERM optimality gives the $2\varepsilon$ slack.
\hfill\ensuremath{\square}

\subsection{Proof of Theorem~\ref{thm:multipac}}
By the Milnor--Thom/Warren sign-pattern bound, $\Lambda$ degree-$\Delta$ polynomials in $p$
variables realize $\le(8e\Delta\Lambda/p)^p$ sign vectors, partitioning $\R^p$ into that many cells
on which the pruned set --- hence $\ell_\theta(I)$ --- is constant. Pseudo-shattering $m$ instances
overlays $\le m\Lambda$ polynomials, giving $\le(8e\Delta m\Lambda/p)^p$ cells, so
\[
  2^m\le(8e\Delta m\Lambda/p)^p
  \ \Longrightarrow\ m=O\big(p\log(\Delta\Lambda)\big)=O(p\log K).
\] Boundedness is
Theorem~\ref{thm:robust} for each still-$\rho$-safe accepted certificate, independent of $\theta$.
\hfill\ensuremath{\square}

\subsection{Proof of Theorem~\ref{thm:cfilter}}
(i) At $\theta=0$ every predicted element clears the threshold, so the committed set is $\hat S$ and
the policy \emph{is} the min-combiner (same committed set \emph{and} the same realized fallback), so
$\ell^{\mathrm{cf}}_0\equiv\ell_{\mathrm{mc}}$ exactly; the $\min$ with $c(\mathrm{fb})\le\alpha\,\OPT$ bounds every
$\ell^{\mathrm{cf}}_\theta$ by $\alpha$; a perfect $\hat S=S^\star$ with $\theta\le\min_{i\in S^\star}\sigma_i$ commits all of
$S^\star$, completing to $S^\star$ at cost $\OPT$.

(ii) The family $\{\ell^{\mathrm{cf}}_\theta\}$ contains $\ell^{\mathrm{cf}}_0=\ell_{\mathrm{mc}}$, so
\begin{equation}\label{eq:cfdom}
  \min_{\theta}\ \mathbb{E}_{\Dc}[\ell^{\mathrm{cf}}_\theta]\ \le\ \mathbb{E}_{\Dc}[\ell^{\mathrm{cf}}_0]=\mathbb{E}_{\Dc}[\ell_{\mathrm{mc}}].
\end{equation}

(iii) Instantiated by
Theorems~\ref{thm:margin} and~\ref{thm:lponly}, proved below, and delimited by
Proposition~\ref{prop:degen}.

(iv) As $\theta$ increases it crosses each $\sigma_i$ once, flipping one element between
committed and filtered, so $\ell^{\mathrm{cf}}_\theta$ has $\le|\hat S|$ breakpoints; the pseudo-dimension and sample
bound follow as in Theorem~\ref{thm:sepB}.
\hfill\ensuremath{\square}

\subsection{Proof of Theorem~\ref{thm:margin}}
Throughout, components (edges, triangles) are vertex-disjoint, so greedy completion decomposes
across components, and the noise coins are independent across elements.

\emph{LP uniqueness and $\sigma$.} On a weighted edge $\{a,b\}$ with $c(a)=1<C=c(b)$, the LP
$\min x_a+Cx_b$ s.t.\ $x_a+x_b\ge1$, $x\ge0$ has the unique optimum $(1,0)$ (any feasible point with
$x_b=t>0$ costs $\ge(1-t)+Ct=1+(C-1)t>1$). On a unit triangle the LP
$\min\sum x_v$ s.t.\ pairwise sums $\ge1$ has value $\frac32$, attained only at
$(\frac12,\frac12,\frac12)$ (any vertex at $0$ forces its two neighbors to $1$, cost $\ge2$).
Hence $\sigma$ is as stated in Definition~\ref{def:Dfam}.

\emph{(i) The filter is surely optimal.} For $\theta\in(\frac12,1]$ the committed set is
$\{a_i:a_i\in\hat S\}$: junk $b_i$ has $\sigma=0$ and every triangle vertex has
$\sigma=\frac12<\theta$. Completion: an uncovered pair edge offers $a_i$ (cost $1$, covers $1$ edge)
versus $b_i$ (cost $C$); greedy takes $a_i$ under any tie-breaking since $C>1$. An untouched
triangle: the first pick covers $2$ of its $3$ unit-cost edges (any vertex; all have ratio
$\frac12$), the second pick covers the remaining edge at cost $1$; total $2=\OPT(K_3)$ for
\emph{every} tie-breaking. Hence the filtered branch costs exactly $n+2g=\OPT$ on every noise
realization; the $\min$ with the fallback cannot increase it, so $\ell^{\mathrm{cf}}_\theta=1$ surely, and in
particular $\mathbb{E}[\ell^{\mathrm{cf}}_\theta]=1$.

\emph{(ii) The min-combiner.} Let $X$ be the commit-all-then-complete cost. Per pair,
$X_i=\mathbf{1}\{a_i\in\hat S\}+C\,\mathbf{1}\{b_i\in\hat S\}
+\mathbf{1}\{a_i\notin\hat S,\,b_i\notin\hat S\}$ (the last term is greedy completion, which takes
$a_i$), so
$\mathbb{E}[X_i]=(1-\eta)+\eta C+\eta(1-\eta)=1+\eta(C-\eta)$.
Per triangle with designated $\{x,y\}$ and non-member $z$: the committed set $T$ contains $x,y$
independently w.p.\ $1-\eta$ and $z$ w.p.\ $\eta$; any two vertices cover $K_3$; if $|T|=1$
completion adds $1$ (one endpoint of the single remaining uncovered edge), and if
$T=\varnothing$ it adds $2$. Thus
\begin{align*}
\mathbb{E}[X_j]&=\mathbb{E}|T|+1\cdot\Pr[|T|=1]+2\cdot\Pr[T=\varnothing]\\
&=(2-\eta)+\big[2\eta(1-\eta)^2+\eta^3\big]+2\eta^2(1-\eta)=2+\eta(1-\eta)^2 .
\end{align*}
Summing, $\mathbb{E}[X]=n\big(1+\eta(C-\eta)\big)+g\big(2+\eta(1-\eta)^2\big)$. The fallback cost is
$n+3g$ surely. The components of $X$ are independent and bounded by $\max(1+C,3)$, so
$\mathrm{Var}(X)\le(n+g)\max(1+C,3)^2$ and, since $|\min(x,c)-\min(y,c)|\le|x-y|$,
\[
\big|\mathbb{E}[\min(X,n+3g)]-\min(\mathbb{E}X,\,n+3g)\big|\le\mathbb{E}|X-\mathbb{E}X|
\le\sqrt{\mathrm{Var}(X)}=O(\sqrt n),
\]
which vanishes after dividing by $\OPT=n+2g=\Theta(n)$. Finally
$\min(\mathbb{E}X,n+3g)-\OPT=\min\{n\eta(C-\eta)+g\eta(1-\eta)^2,\ g\}$; dividing by $n+2g$ and
letting $n\to\infty$ with $g/n\to\beta$ gives the displayed limit.

\emph{(iii)} Subtract (i) from (ii). Monotonicity near $0$: the unsaturated branch
$\eta[(C-\eta)+\beta(1-\eta)^2]$ has derivative $C+\beta>0$ at $\eta=0$; the saturated branch is the
constant $\beta$, attained once $\eta[(C-\eta)+\beta(1-\eta)^2]\ge\beta$. For $C>1+\beta$ the
unsaturated branch is continuous with value $C-1>\beta$ at $\eta=1$, so it crosses $\beta$ at some
$\eta<1$ and the cap is attained; for $C\le1+\beta$ it may stay below $\beta$ on all of $[0,1)$
(e.g.\ $\beta=2$, $C=2$, where its supremum is ${\approx}1.04$), and the margin then never reaches
the cap.
\hfill\ensuremath{\square}

\subsection{Proof of Theorem~\ref{thm:lponly}}
\emph{The optimal LP face of a gadget.} Coverage of $p_j,q_j,r_j$ requires
$t+x_{s^k_j}\ge1$ for each $k$, where $t:=x_{T_j}+x_{T'_j}$. The cost is
$(1+\varepsilon)t+\sum_k c_k x_{s^k_j}\ge(1+\varepsilon)t+\tfrac{11}6(1-t)
=\tfrac{11}6-t(\tfrac56-\varepsilon)$, decreasing in $t$ since $\varepsilon<\tfrac56$; so the LP
optimum is $t=1$, $x_s=0$, value $1+\varepsilon$, and the optimal face is the segment
$\{x_{T_j}+x_{T'_j}=1,\,x_s=0\}$, whose relative interior is
$x_{T_j}=x_{T'_j}=\frac12$ by symmetry. Pairs have unique optimum $x_{A_i}=1,x_{B_i}=0$ as in
Theorem~\ref{thm:margin}. Hence $\bar\sigma$ is as stated.

\emph{Symmetric policies.} Fix $A\subseteq[0,1]$ and either canonicalization. On every gadget the
two twins carry equal confidence ($\frac12$ under the analytic center, $1$ under the max), so a
symmetric policy commits both twins or neither. Committing both pays $2(1+\varepsilon)$ on the
gadget. Committing neither leaves greedy completion, which is uniquely determined: ratios are
$\frac13$ ($s^3$) vs $\frac{1+\varepsilon}3$ (twins), so $s^3_j$; then $\frac12$ ($s^2$) vs
$\frac{1+\varepsilon}2$, so $s^2_j$; then $1$ ($s^1$) vs $1+\varepsilon$, so $s^1_j$: cost
$\frac{11}6$, with no tie ever occurring (committing singletons, possible when $0\in A$, only adds
cost). Since $2(1+\varepsilon)>\frac{11}6$ for every $\varepsilon>0$, every symmetric policy pays
at least $\frac{11}6$ per gadget, surely. On pairs the cost is at least $1$ (commit $A_i$, or
completion picks the $a$-side; committing junk $B_i$, possible when $\sigma_{B_i}=0\in A$, only
adds $C$). Total: at least $n+\frac{11}6m$ surely, attained by the policy $A=\{1\}$ under the
analytic-center canonicalization, which commits $\{A_i\}$ and nothing on gadgets. The optional
min with a symmetric fallback does not lower this: a deterministic symmetric procedure outputs a
cover invariant under the twin swap, hence containing
both twins or neither on each gadget; the neither-twin covers of a gadget are exactly the three
singletons (cost $\frac{11}6$) and the both-twin covers cost at least $2(1+\varepsilon)$, so
\emph{every} symmetric branch --- commit, completion, or fallback --- pays at least
$\min\{\frac{11}6,2(1+\varepsilon)\}=\frac{11}6$ per gadget and at least $1$ per pair, and the
min over such branches stays at least $n+\frac{11}6m$.

\emph{Filter at $\theta=\frac12$.} It commits predicted elements with
$\bar\sigma\ge\frac12$: the $A_i$'s present in $\hat S$ and any predicted twin; junk $B_i$ and
singletons have $\bar\sigma=0$ and are filtered. Pairs cost $1$ each surely (committed $A_i$, or
completion picks the $a$-side as before). On gadget $j$: $T_j\in\hat S$ w.p.\ $1-\eta$ (it is in
$S^\star$) and $T'_j\in\hat S$ w.p.\ $\eta$, independently; each committed twin costs $1+\varepsilon$
and covers the gadget; if neither is committed (probability $\eta(1-\eta)$), greedy completion pays
$\frac{11}6$ as above. Since $\mathbb{E}[\#\text{twins committed}]=(1-\eta)+\eta=1$,
\[
\mathbb{E}[\text{gadget cost}]=(1+\varepsilon)\cdot1+\tfrac{11}6\,\eta(1-\eta).
\]
Subtracting from $\frac{11}6$ and using $\eta(1-\eta)\le\frac14$:
\[
\tfrac{11}6\big(1-\eta(1-\eta)\big)-(1+\varepsilon)\ \ge\ \tfrac{11}6\cdot\tfrac34-1-\varepsilon
=\tfrac38-\varepsilon\ >\ 0 .
\]
(The min with the realized fallback can only enlarge the gap, since the filter branch is never
above the LP-commit cost in expectation and the same symmetric fallback is available to both.) For
Remark~\ref{rem:canon}: under $\sigma_i=\max\{x_i:x\ \text{optimal}\}$ both twins have $\sigma=1$,
LP-commit takes both ($2(1+\varepsilon)$ per gadget, covering it), and the filter's expected
advantage per gadget is $2(1+\varepsilon)-(1+\varepsilon)-\tfrac{11}6\eta(1-\eta)
\ge1+\varepsilon-\tfrac{11}{24}>0$.
\hfill\ensuremath{\square}

\subsection{Proof of Proposition~\ref{prop:degen}}
If $\sigma_i=s$ for all $i\in\hat S$, then $\{i\in\hat S:\sigma_i\ge\theta\}$ equals $\hat S$ for
$\theta\le s$ and $\varnothing$ for $\theta>s$. The first case is the min-combiner
(Theorem~\ref{thm:cfilter}(i)); the second commits nothing and returns
$\min(\text{completion-from-scratch},\,\mathrm{fb})$. No other committed set is realizable, so the
family contains exactly these two policies.
\hfill\ensuremath{\square}

\subsection{Proof of Proposition~\ref{prop:cfplus}}
\emph{(i)} Each named policy is realized by the indicated parameter setting (for
``$\theta$ above every $\sigma_i$'' take $\theta=1+\epsilon$, permitted since thresholds range over
an interval containing $[0,1]$): the committed set reduces to $\hat S$, to
$\hat S\cap\{\sigma\ge\theta_1\}$, to $\{i:\sigma_i\ge\theta_2\}$, or to $\varnothing$
respectively, and the completion and fallback are shared. Containment gives
$\min_{\theta_1,\theta_2}\mathbb{E}[\ell^{\mathrm{cf}}_{\theta_1,\theta_2}]\le\mathbb{E}[\ell_P]$ for each
contained policy $P$.

\emph{(ii)} As $\theta_1$ increases it crosses each of the $\le|\hat S|$ values
$\{\sigma_i:i\in\hat S\}$ once, and $\theta_2$ crosses each of the $\le K$ values
$\{\sigma_i\}$ once; the committed set --- hence the loss --- is constant on each of the
$\le(|\hat S|{+}1)(K{+}1)$ open rectangles of the induced grid. This is the
$p{=}2$, $\Delta{=}1$, $\Lambda\le2K$ case of Theorem~\ref{thm:multipac}, giving
$\Pdim=O(2\log K)=O(\log K)$ and the stated sample complexity; boundedness is
Theorem~\ref{thm:robust} via the shared fallback branch.
\hfill\ensuremath{\square}

\subsection{Proof of Theorem~\ref{thm:genmargin}}
Surely-optimality: by (A1) the filter's branch cost is $\sum_c n_c\OPT_c=\OPT(I_n)$ on every
realization; the min with the fallback cannot increase it. For the min-combiner let $X$ be the
commit-all cost; by (A2) and linearity $\mathbb{E}X=\sum_c n_c(\OPT_c+\varphi_c(\eta))$, and by
(A3) the fallback cost is $F=\sum_c n_c(\OPT_c+\delta_c)$ surely. Components are independent with
costs bounded by a constant $b$, so $\mathrm{Var}(X)\le b^2n$ and, since
$|\min(x,F)-\min(y,F)|\le|x-y|$,
$|\mathbb{E}\min(X,F)-\min(\mathbb{E}X,F)|\le\mathbb{E}|X-\mathbb{E}X|\le b\sqrt n$. Divide by
$\OPT(I_n)=\Theta(n)$ and pass to the limit $n_c/n\to w_c$.
\hfill\ensuremath{\square}

\subsection{Proof of Theorem~\ref{thm:gendeg}}
Fix a core copy $g$. Conditioning on whether a cost-$\OPT_g$ cover is committed (probability
$q_g$), the filter's branch cost on $g$ is at most $\OPT_g$ plus surplus in the first case and at
most $\OPT_g(1+\delta_g)$ in the second (B3); taking expectations and using (B2),
\[
  \mathbb{E}[\mathrm{filter}_g]\ \le\ \OPT_g+\rho_g(\eta)+(1-q_g(\eta))\,\delta_g\,\OPT_g .
\]
By (B1), $\mathbb{E}[\mathrm{LP\mbox{-}commit}_g]\ge\OPT_g(1+\delta_g)$; subtracting gives
$q_g\delta_g\OPT_g-\rho_g$ per copy. Summing over core copies and canceling the non-core
components by (B0) yields the display. For $H_{m,n}$: surplus occurs exactly when both twins are
committed, $\rho_g=(1+\varepsilon)\eta(1-\eta)$, and (B3) holds with equality at $\frac{11}6$, so
$\mathbb{E}[\mathrm{filter}_g]=(1+\varepsilon)+\tfrac{11}6\eta(1-\eta)$ exactly, matching
Theorem~\ref{thm:lponly}.
\hfill\ensuremath{\square}

\subsection{Proof of Lemma~\ref{lem:barrier}}
Suppose $\Cl$ certifies optimality on all instances. To decide $(I,k)\in\mathrm{OptVer}$, an NP
machine guesses an optimal $S$ and a proof $\pi$, verifies $\Cl(I,S,\pi)=1$ in polynomial time,
and accepts iff $c(S)\ge k$. Soundness ($c(S)=\OPT(I)$) makes this correct, so
$\mathrm{OptVer}\in\mathrm{NP}$; the coNP-hardness of $\mathrm{OptVer}$ then gives $\mathrm{coNP}\subseteq\mathrm{NP}$.
\hfill\ensuremath{\square}

\subsection{Proof of Theorem~\ref{thm:sepA}}
\emph{(i)} Let $\pi=(x^\star,(P_0,P_{1/2},P_1),T)$ where $T$ is the brute-force table of the core
$G[P_{1/2}]$ listing each of the $2^{|P_{1/2}|}$ subsets, whether it covers $G[P_{1/2}]$ and its
cost, with $S_{\mathrm{core}}$ a minimum one. Define $\Cl(G,S,\pi)=1$ iff: (a) recomputing the VC
LP confirms $x^\star$ is feasible, optimal (value matches), half-integral, and induces
$(P_0,P_{1/2},P_1)$ (polynomial); (b) $|P_{1/2}|\le c_0\log|W|$, the table $T$ recomputes
(polynomial, since $2^{|P_{1/2}|}\le|W|^{c_0}$), and $S_{\mathrm{core}}$ is a minimum cover of the
core; (c) $S=P_1\cup S_{\mathrm{core}}$ and $c(S)=c(P_1)+c(S_{\mathrm{core}})$. \emph{Soundness}:
upon acceptance, the NT persistency theorem (Theorem~\ref{thm:nt}, hard-coded in the checker)
gives $\OPT(G)=c(P_1)+\OPT(G[P_{1/2}])$; (b) gives $\OPT(G[P_{1/2}])=c(S_{\mathrm{core}})$; the NT
structure lemma guarantees $S=P_1\cup S_{\mathrm{core}}$ is a cover; hence $c(S)=\OPT(G)$. CASP
produces an accepted $(S,\pi)$ on every $G\in \mathcal{K}_{\log}$ in time
$\mathrm{poly}(|W|)+|W|^{c_0}$; the LP solve is the trigger.

\emph{(ii)} ``No polynomial-time algorithm on every instance'' is Lemma~\ref{lem:barrier} applied
to VC, and it binds CASP and $P$-algorithms alike. A perfect prediction does not lift the barrier:
a $P$-algorithm handed $\hat S=S^\star$ may output $S^\star$, but certifying it on \emph{all}
instances would exhibit an optimality-proof system on all instances, contradicting
Lemma~\ref{lem:barrier}. The interface statement is definitional: the output of
Definition~\ref{def:pos} is a solution with no proof slot, so no certificate is emitted on any
instance; and the prediction $\hat S$ itself carries no checkable lower-bound witness --- the
generic poly-checkable witness available from LP/dual feasibility is tight only on LP-integral
instances. This does \emph{not} bound what an unrestricted polynomial algorithm consuming
$\hat S$ can do: it may ignore the advice and run the pipeline of (i), certifying all of
$\mathcal{K}_{\log}$ (Corollary~\ref{cor:noeff}) --- which is why we present this contrast as
interface-bound. $\mathcal{K}_{\log}$ contains instances whose whole-graph LP optimum is fractional yet whose core is
logarithmic (Proposition~\ref{prop:planted}).
\hfill\ensuremath{\square}

\subsection{Proof of Proposition~\ref{prop:planted}}
(i) Take $x^\star=1$ on centers, $0$ on leaves and outer vertices, and the canonical $\frac12$
optimum on $H$. Feasibility: center--leaf and center--outer edges are covered by the center;
center--center edges have both ends $1$; edges inside $H$ sum $\frac12+\frac12=1$. Lower bound: for
each center its two leaf constraints give gadget cost
\[
  x_{c}+x_{l_1}+x_{l_2}\ge2-x_c\ge1;
\]
summing
$k$ disjoint gadgets gives $\ge k$, and $H$ contributes its odd-cycle bound, both attained by
$x^\star$, so $x^\star$ is LP-optimal (and it is the unique optimum: equality forces $x_c=1$,
leaves and outer at $0$, and the all-$\frac12$ point is the unique optimum on $H$). The induced
partition has $P_1\supseteq\{\text{centers}\}$,
$P_0\supseteq\{\text{leaves, outer}\}$, $P_{1/2}=V(H)$; persistency (Theorem~\ref{thm:nt})
certifies an optimum containing all centers and no leaf/outer vertex, residual exactly $H$.

(ii)
$s=O(\log n)\Rightarrow2^s=\mathrm{poly}(n)$, and Theorem~\ref{thm:sepA}(i) supplies the proof; the
whole-graph LP is fractional because $H$ sits at $\frac12$.

(iii) The first clause is definitional
(the interface of Definition~\ref{def:pos} outputs no proof object); the second follows from
Corollary~\ref{cor:noeff} because every $G(n,k,H)$ with $s=O(\log n)$ lies in $\mathcal{K}_{\log}$ by (i).
\hfill\ensuremath{\square}

\subsection{Proof of Theorem~\ref{thm:sepB}}
\emph{(1)} Boundedness is Theorem~\ref{thm:robust}: every accepted certificate is $f$-safe and the
fallback is $\alpha$-approximate, so the branch maximum is $\max(f,\alpha)$ on every instance and
for every $\tau$ --- a problem constant (the verifier rejects $\tau>1/f$,
Corollary~\ref{cor:tradeoff}, so that branch too returns the fallback). For the pseudo-dimension, $\ell_\tau(I)$ is
piecewise-constant in $\tau$ with $\le|\G(I)|\le K$ breakpoints (each crossed LP value flips one
set between pruned/survived); pseudo-shattering $m$ instances needs $2^m\le mK+1$, so
$m=O(\log K)$. Substitute into \eqref{eq:uc}.

\emph{(2)} \emph{Same pseudo-dimension}: $\ell^+_t(I)$ is piecewise-constant in $t$ with $\le K$
breakpoints (the item scores), so the identical dual-counting gives $\Pdim(\F^+)=O(\log K)$.
\emph{Unbounded range}: take $U=\{e\}$ with two sets $S_1=\{e\}$ ($c=1$) and $S_2=\{e\}$ ($c=M$),
so $\OPT=1$; a popularity-biased scorer ranks $S_2$ above $S_1$, $s(S_2)=1>s(S_1)=0$. For any $t\in(0,1]$ the policy commits $S_2$ (already feasible) at cost $M$, so
\[
  \ell^+_t(I)=M\longrightarrow\infty .
\]
There is no verifier to reject the commitment. CASP on the same instance solves the LP
($x^\star_{S_1}=1,x^\star_{S_2}=0$); for $\tau\in(0,\frac12]$ ($f{=}2$) it prunes $S_2$ and gets
$\ell_\tau=1$, and for $\tau>\frac12$ the verifier rejects the threshold
(Corollary~\ref{cor:tradeoff}) and the greedy fallback returns $S_1$, so $\ell_\tau=1$ for every
$\tau$.
Capping $c_{\max}/c_{\min}\le R$ caps the range at $\Theta(R)$; the $B^2$ factor of \eqref{eq:uc}
then exceeds CASP's by $O(R^2)$, and Lemma~\ref{lem:heavytail} rules out any distribution-free
rate for the uncapped class.
\hfill\ensuremath{\square}

\subsection{Proof of Lemma~\ref{lem:heavytail}}
Fix $N$. Choose $M:=10N(B_0+2)$ and let $\Dc_N=(1-p)\delta_{I_0}+p\,\delta_{I_M}$ with
$p:=\frac1{10N}$. With probability $(1-p)^N\ge1-Np=0.9$ the sample contains no copy of $I_M$; on
that event, for $t^\circ\in\arg\sup_t\ell_t(I_M)$ the empirical mean is
$\frac1N\sum_i\ell_{t^\circ}(I_i)=\ell_{t^\circ}(I_0)\le B_0$, while
$\mathbb{E}_{\Dc_N}[\ell_{t^\circ}]\ge p\,M=B_0+2$ (losses are nonnegative). Hence
$\sup_t|\mathbb{E}\ell_t-\hat{\mathbb{E}}\ell_t|\ge2>1$ with probability at least $0.9$. Since
$N$ was arbitrary, no distribution-free uniform-convergence rate exists.
\hfill\ensuremath{\square}

\subsection{Proof of Theorem~\ref{thm:spreadlb}}
Fix $R\ge3$. Two instances over disjoint universes; the scorer is fixed with the instances.

\emph{Gadget instance} $I_G$: elements $\{e_1,e_2\}$; sets $T=\{e_1,e_2\}$ (cost $\frac65$, score
$0.7$), $X_1=\{e_1\}$ (cost $\frac12$, score $0$), $X_2=\{e_2\}$ (cost $\frac9{10}$, score $0$);
$\OPT=\frac65$. For $t\in(0,0.7]$ the policy commits exactly $T$: $\ell^+_t=1$. For $t>0.7$
nothing is committed and greedy completes by cost per uncovered element: $X_1$
($\frac12<\frac35$), then $X_2$ ($\frac9{10}<\frac65$), total $\frac75$: $\ell^+_t=\frac76$. For
$t\le0$ every set is committed: $\ell^+_t=\frac{13}6$.
\emph{Junk instance} $I_J$: element $\{e\}$; $S_0=\{e\}$ (cost $1$, score $0$) and $S_J=\{e\}$
(cost $R$, score $0.7$); $\OPT=1$. For $t\in(0,0.7]$: $\ell^+_t=R$; for $t>0.7$: greedy picks
$S_0$ (ratio $1<R$), $\ell^+_t=1$; for $t\le0$: $\ell^+_t=R+1$. Every instance has spread at most
$\max(R,\frac{12}5)=R$ and frequency $f=2$.

\emph{Two arms.} Under $\Dc_p=(1-p)\delta_{I_G}+p\,\delta_{I_J}$ every threshold realizes one of
three profiles: the \emph{low arm} $t\in(0,0.7]$ with mean $(1-p)+pR$; the \emph{high arm}
$t>0.7$ with mean $(1-p)\frac76+p$; and $t\le0$, pointwise dominated by the low arm. Their
difference
\[
  \Delta(p)=\big[(1-p)+pR\big]-\big[(1-p)\tfrac76+p\big]=p(R-1)-\tfrac{1-p}6
\]
is affine in $p$ with slope $R-\frac56$ and root $p_0=\frac1{6R-5}\le\frac1{13}$.

\emph{Two distributions.} Let $\delta:=\frac{3\varepsilon}{R-5/6}$ and
$\Dc^\pm:=\Dc_{p_0\pm\delta}$; then $\Delta(p_0\pm\delta)=\pm3\varepsilon$: under $\Dc^+$ the high
arm is better by $3\varepsilon$, under $\Dc^-$ the low arm is. Any $\hat t$ with
$\mathbb{E}[\ell^+_{\hat t}]\le\min_t\mathbb{E}[\ell^+_t]+\varepsilon$ therefore lies in the high
arm under $\Dc^+$ and in the low arm under $\Dc^-$, so the rule ``report $+$ iff $\hat t>0.7$''
distinguishes $(\Dc^+)^{\otimes N}$ from $(\Dc^-)^{\otimes N}$ with error at most $\frac14$ on
each, forcing $\mathrm{TV}\big((\Dc^+)^{\otimes N},(\Dc^-)^{\otimes N}\big)\ge\frac12$.

\emph{Le Cam.} Since $R-\frac56=\frac{6R-5}6$, the condition $\varepsilon\le\frac1{36}$ is
exactly $\delta\le p_0/2$. A sample reveals only which of $I_G,I_J$ was drawn, so it is a
$\mathrm{Bern}(p)$ observation, and by
$\mathrm{KL}(\mathrm{Bern}(p)\Vert\mathrm{Bern}(q))\le(p-q)^2/(q(1-q))$ with
$q=p_0-\delta\ge p_0/2$ and $1-q\ge\frac12$,
\[
  \mathrm{KL}\big(\Dc^+\Vert\Dc^-\big)\le\frac{(2\delta)^2}{(p_0/2)(1/2)}=\frac{16\delta^2}{p_0}.
\]
Pinsker gives $\frac14\le\mathrm{TV}^2\le\frac12N\,\mathrm{KL}\le\frac{8N\delta^2}{p_0}$, i.e.
\[
  N\ \ge\ \frac{p_0}{32\delta^2}
  \ =\ \frac{(R-5/6)^2}{288\,(6R-5)\,\varepsilon^2}
  \ =\ \frac{R-5/6}{1728\,\varepsilon^2}
  \ \ge\ \frac{R}{3456\,\varepsilon^2}\, .
\]

\emph{The CASP side.} On $I_G$ the covering LP has the unique optimum $x^\star_T=1$,
$x^\star_{X_1}=x^\star_{X_2}=0$ (every feasible $x$ costs at least $\frac75-\frac15x_T\ge\frac65$,
with equality only at $x_T=1$); on $I_J$ it is $x^\star_{S_0}=1$, $x^\star_{S_J}=0$. For every
$\tau\in(0,\frac12]$ ($f=2$) the pruned instance retains exactly the optimal sets, so the reduced
solve returns them: $\ell_\tau=1$ on every realization.
\hfill\ensuremath{\square}

\subsection{Proof of Theorem~\ref{thm:collapse}}
$c(S_{\mathrm{fb}})\le\alpha\OPT(I)$, and the combiner returns the cheaper of
$S_{\mathrm{pred}},S_{\mathrm{fb}}$, giving the displayed bound and range $[1,\alpha]$ for every
prediction and instance; a perfect prediction makes the min equal $\OPT(I)$. The loss is
piecewise-constant in the threshold with $\le K$ breakpoints intersected with a
prediction-independent baseline, so $\Pdim=O(\log K)$ is unchanged and \eqref{eq:uc} applies with
range $\alpha$. Theorem~\ref{thm:sepB}(2) obtained an unbounded range only by \emph{forbidding}
$S_{\mathrm{fb}}$; restoring it removes the gap.
\hfill\ensuremath{\square}

\subsection{Proof of Corollary~\ref{cor:noeff}}
Cost, range, consistency, and sample complexity are Theorem~\ref{thm:collapse}. For the
certified-exact rate: the pipeline of Theorem~\ref{thm:sepA}(i) is a fixed polynomial-time
computation, so the combiner --- permitted (c) --- runs it and reproduces $(S,\pi)$ with
$\Cl(I,S,\pi)=1$ on every $I\in \mathcal{K}_{\log}$; the coNP barrier of Lemma~\ref{lem:barrier} bounds
\emph{both} paradigms identically off $\mathcal{K}_{\log}$. No axis remains on which signing direction, rather
than the modeling restriction, produces a gap.
\hfill\ensuremath{\square}

\end{document}